%% file: acl_latex.tex
\documentclass[11pt]{article}
\usepackage[most]{tcolorbox}

\usepackage[preprint]{acl}
\usepackage{titletoc}
\usepackage{times}
\usepackage{latexsym}
\usepackage{pifont}

\usepackage[T1]{fontenc}
\usepackage{colortbl}
\usepackage{makecell}
\usepackage{multirow}
\usepackage[utf8]{inputenc}
\usepackage{booktabs}
\usepackage{microtype}

\usepackage{inconsolata}

\usepackage{pifont,xcolor,colortbl,adjustbox}

\usepackage{graphicx}
\usepackage{algorithm}
\usepackage{algpseudocode}
\usepackage{amsmath}
\usepackage{amssymb}
\usepackage{kotex}
\usepackage[most]{tcolorbox}
\definecolor{dropred}{HTML}{E00000}
\definecolor{groupbg}{HTML}{E3ECF7}
\definecolor{basebg}{gray}{0.93}
\definecolor{oursbg}{HTML}{FDF1E3}
\usepackage{tabularx}
\newcolumntype{Y}{>{\centering\arraybackslash}X}
\newcolumntype{R}{>{\centering\arraybackslash\leavevmode\color{dropred}}X}
\title{Binding Paraphrased Instructions to Encountered Objects \\ in sLLM Agents under Partial Observability}

\author{
  Chanyoung Kim$^{1*}$, Harim Song$^{2*}$, Byoungchan Ko$^{2}$, Keonyoung Lee$^{2}$, Myungwon Lee$^{2}$, Dahuin Jung$^{2\dagger}$ \\
  \\
  $^{1}$School of Computer Science and Engineering, Soongsil University, Seoul, Korea \\
  $^{2}$Department of Artificial Intelligence, Chung-Ang University, Seoul, Korea \\
  \texttt{dahuinjung@cau.ac.kr} \\
}

\begin{document}
\maketitle

\let\thefootnote\relax
\footnotetext{$^{*}$Equal contribution. $^{\dagger}$Corresponding author.
\vspace{-21.5pt}}
\addtocounter{footnote}{0}

\input{sections/00_abstract}
\input{sections/01_introduction}

\input{sections/02_relatedwork}
\input{sections/03_alfworldpara}
\input{sections/04_analysis}
\input{sections/05_method}

\input{sections/06_experiment}

\input{sections/07_conclusion}

\input{sections/08_limitation}

\bibliography{custom}
\input{sections/50_appendix_face}

\input{sections/93_appen_alfworldpara}
\input{sections/94_appen_analysis}

\input{sections/95_appen_method}

\input{sections/96_appen_experiment}
\input{sections/97_appen_scipara}
\input{sections/98_ai_assistants}

\end{document}

%% file: sections/00_abstract.tex
\begin{abstract}
Reliable deployment of LLM agents requires grounding diverse user expressions in their environments. Although RL-finetuned smaller large language model (sLLM) agents rival larger prompted LLMs on embodied tasks, their robustness to instruction paraphrases remains unclear. We introduce ALFWorld-Para to evaluate this robustness under controlled paraphrases, revealing 21--52\% declines in success rates (SRs), predominantly driven by object synonyms. Despite retaining knowledge of synonym referents, agents often fail to take the corresponding objects even after observing them, exposing a gap between referential knowledge and object binding. Partial observability further complicates binding, as referents may remain unseen or have already been encountered under different labels. To address this challenge, we propose BINDIntent, a training-free framework that probabilistically binds referring expressions to observed labels using semantic proximity and estimated persistence against unseen candidates. Without additional training or LLM calls, BINDIntent improves synonym-paraphrase success of RL-finetuned sLLMs by 17.5--49.1 percentage points (pp) with only 0.4\,ms of CPU overhead per step. It also raises GiGPO-7B's overall SR from 61.8\% to 77.4\%, surpassing a prompted 24B LLM. Code and benchmark are available at \url{https://github.com/cau-hai-lab/ALFWorld-Para}.
\end{abstract}

%% file: sections/01_introduction.tex
\section{Introduction}
\input{figures/intro_figure}

Smaller large language models (sLLMs; fewer than 10B parameters)~\cite{belcak2025small} have gained traction as cost-efficient backbones for LLM agents. In partially observable embodied environments, sLLM agents fine-tuned with reinforcement learning (RL) can rival or outperform larger prompted LLMs~\cite{feng2026gigpo, xue2026strata}, positioning them as viable deployment candidates. Yet deployment exposes agents to linguistic variability: as illustrated in Figure~\ref{fig:intro_figure}, users articulate the same intent in different ways and refer to identical objects using different terms~\cite{ervin-tripp1976, furnas1987vocabulary}. Robustness to instruction paraphrases is therefore important for reliable deployment. Despite this need, existing agents are trained and evaluated on instructions drawn from a set of fixed templates~\cite{shridhar2020alfworld}. As summarized in Table~\ref{tab:relatedwork_framing}, prior robustness studies likewise focus predominantly on system-side perturbations, including prompt templates and action interfaces. Consequently, the robustness of sLLM agents, particularly those fine-tuned with RL, to paraphrased user instructions remains largely unexplored.

To evaluate this robustness, we introduce ALFWorld-Para, which varies only goal instruction phrasing while preserving the underlying task. It independently paraphrases action and object expressions to isolate their contributions to task failure, following LIBERO-Para~\cite{liberopara}. RL-finetuned sLLM agents and prompting-based LLMs exhibit 21--52\% declines in SR on ALFWorld-Para, largely eroding RL gains. This degradation stems predominantly from object synonym substitutions. Yet 7B models identify the corresponding environment labels with nearly 80\% accuracy, with negligible changes after RL fine-tuning. Despite this knowledge, fine-tuned agents take observed targets in over 90\% of cases under original labels, compared with only 20--56\% under synonyms. These findings implicate failures in binding paraphrased expressions to encountered objects. Under partial observability, agents must determine whether an exact-matching label remains unseen or its referent has already been observed under another label. Exhaustive exploration resolves this ambiguity but incurs substantial step costs and requires deviations from learned behavior.


To address this binding challenge without altering the agent's policy, we propose \textbf{BINDIntent: Binding Instruction Objects in Partially Observable Environments}, a training-free framework that probabilistically commits to binding as observational evidence accumulates. BINDIntent extracts object referring expressions and identifies the closest observed label using word embedding distances (candidate distances). To account for unseen candidates, it models observed distances with a Weibull distribution and estimates persistence: the probability that the best match remains superior across unexplored units (e.g., unopened drawers). As persistence increases, BINDIntent accumulates binding probability and, upon binding, substitutes the expression with the selected label before forwarding it to the frozen agent. Requiring neither additional training nor LLM calls, BINDIntent incurs about 0.4 ms of CPU overhead per step. On RL-finetuned sLLMs, it improves SRs by 17.5--49.1\,pp under synonym paraphrases and 7.0--20.6\,pp overall.

%% file: figures/intro_figure.tex

\begin{figure}[t]
  \centering
  \includegraphics[width=\columnwidth]{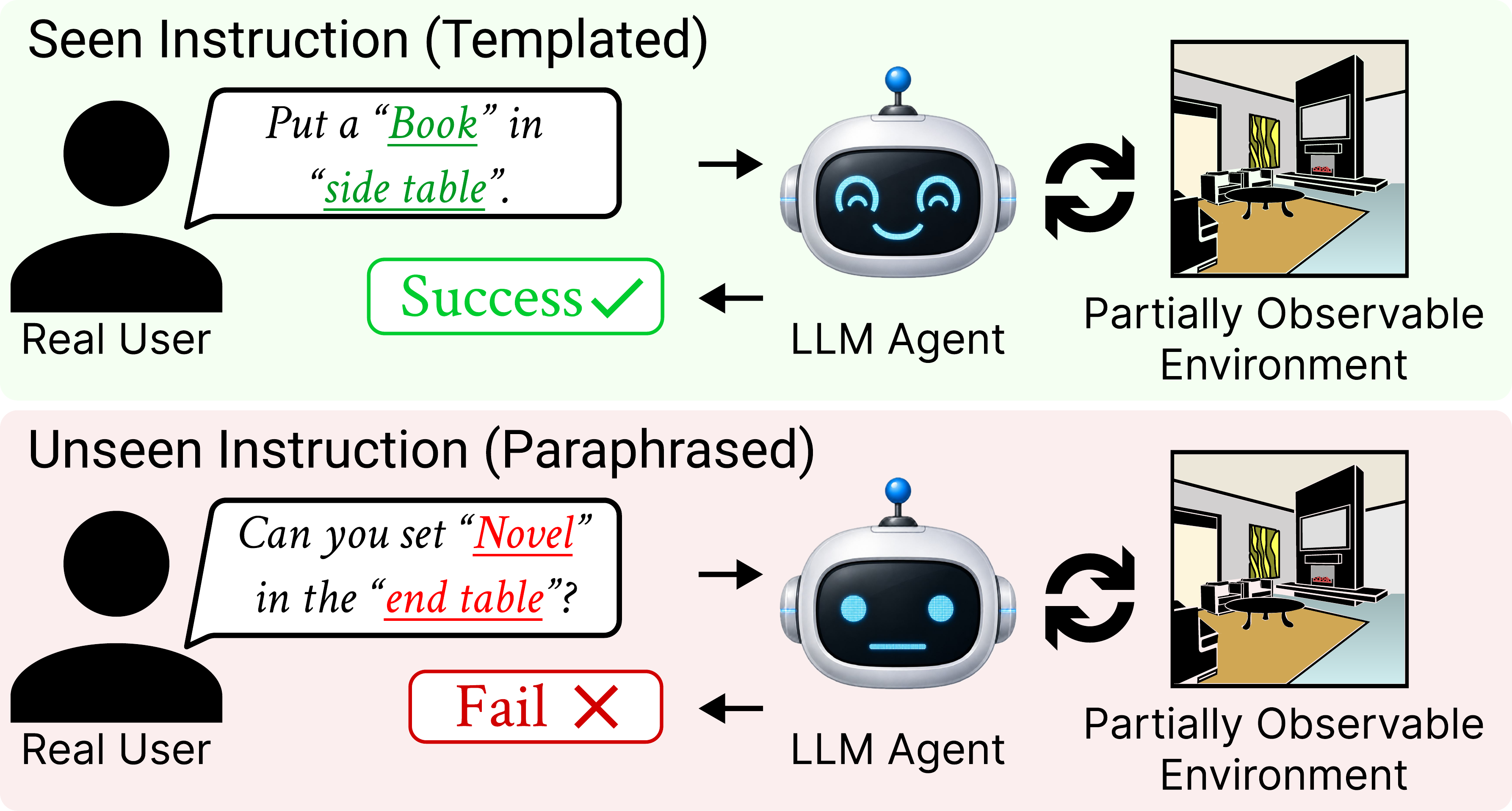}
  \vspace{-19pt}
  \caption{Under partial observability, an sLLM agent that succeeds on a templated instruction fails on its paraphrased counterpart.}
  \vspace{-10pt}
  \label{fig:intro_figure}
\end{figure}

%% file: sections/02_relatedwork.tex
\section{Related Work}

\subsection{From Prompted LLMs to RL-Finetuned sLLM Agents}


LLMs have increasingly been employed as agents in long-horizon environments characterized by partial observability~\cite{kaelbling1998planning} and sparse rewards~\cite{wang2024survey, shridhar2020alfworld, wang2022scienceworld}. To tackle these challenges, agent development has progressed from prompting~\cite{yao2023react, shinn2023reflexion} and supervised fine-tuning (SFT)~\cite{zeng2024agenttuning, chen2023fireact} to reinforcement learning (RL)~\cite{zhang2025landscape}. Within RL, algorithms have evolved from PPO and GRPO~\cite{schulman2017ppo, deepseek-math_grpo} toward finer-grained credit assignment (GiGPO)~\cite{feng2026gigpo} and strategy abstraction (StraTA)~\cite{xue2026strata}. These advances have enabled RL-finetuned sLLM agents to outperform larger prompted LLMs, highlighting their deployment potential~\cite{belcak2025small}. However, the agents are evaluated on the same instruction templates used during training~\cite{shridhar2020alfworld}, leaving their robustness to diverse user expressions~\cite{ervin-tripp1976, micheli2021language} insufficiently examined.

\input{tables/relatedwork_framing}

\subsection{Linguistic Robustness of LLM Agents}

LLMs are sensitive to variations in instruction phrasing~\cite{sclar2024llm_prompt_sensitive, mizrahi2024stateofwhatart}, yet this sensitivity has primarily been studied in single-turn tasks, such as question answering and mathematical reasoning~\cite{sun2024evaluating}. In multi-turn agents, instruction misinterpretations can propagate through subsequent actions, amplifying their impact. Nevertheless, as Table~\ref{tab:relatedwork_framing} shows, existing robustness studies focus predominantly on system-side perturbations, including prompt templates~\cite{aissi2025reinforcement}, action interfaces~\cite{fu2025agentrefine, xia2025agentrm, gu2026agents}, environment configurations~\cite{ye2026look}, and model parameters~\cite{wang2026learning}. Studies on instruction variability have examined either free-form instructions without distinguishing paraphrase types~\cite{yang2024embodied} or systematic action and object paraphrases limited to vision-language-action (VLA) models~\cite{liberopara}. Furthermore, prior studies on RL-finetuned agents each evaluated a single RL algorithm, while proposed mitigation strategies require additional training. To address these gaps, we evaluate 29 paraphrase types across five prompted LLMs and agents independently fine-tuned with four RL algorithms, and introduce a training-free mitigation framework.

\subsection{Grounding Instruction Vocabulary to Environment Entities}

Existing grounding approaches for LLM agents fall into three categories. Some map free-form actions to admissible actions through semantic similarity~\cite{huang2022zeroshot} or select executable skills based on affordances~\cite{ahn2022saycan}, assuming all candidate actions are available upfront. Others replan using observed objects~\cite{song2023llmplanner} or reconcile object names between planning and perception modules~\cite{kim2025multi}, addressing vocabulary discrepancies within the agent. A third line requests assistance under uncertainty~\cite{ren2023knowno, suri2025sage} or learns policies for both acting and querying through RL~\cite{ramrakhya2025asktoact}. These approaches share a common reliance on interactions with responsive users to resolve ambiguous instructions. 
Our work focuses on binding paraphrased object expressions to incrementally revealed entities in partially observable multi-turn environments without access to a global object vocabulary or additional user responses.

\input{tables/alfworldpara_example}

%% file: tables/relatedwork_framing.tex
\providecommand{\hdr}[1]{\begin{tabular}[c]{@{}c@{}}#1\end{tabular}}
\providecolor{oursbg}{gray}{0.92}

\begin{table}[t]
  \centering
  \small
  \renewcommand{\arraystretch}{0.95}
  \resizebox{\columnwidth}{!}{%
  \begin{tabular}{lcccc}
    \toprule
    \hdr{Prior Work}
      & \hdr{Perturbed\\ Component}
      & \hdr{Instruction\\ Para.\ Types (\#)}
      & \hdr{RL Algorithms}
      & \hdr{Mitigation} \\
    \midrule
    \citealp{yang2024embodied}       & Instruction & 1  & --           & Training \\
    \citealp{aissi2025reinforcement} & System      & -- & PPO          & Training \\
    \citealp{fu2025agentrefine}      & System      & -- & --           & Training \\
    \citealp{xia2025agentrm}         & System      & -- & --           & Training \\
    \citealp{gu2026agents}           & System      & -- & --           & -- \\
    \citealp{ye2026look}             & System      & -- & GRPO         & Training \\
    \citealp{wang2026learning}       & System      & -- & GiGPO & Training \\
    \citealp{liberopara}$^\dagger$   & Instruction & 43 & --           & -- \\
    \midrule
    \rowcolor{oursbg}
      &  &  & \textbf{PPO, GRPO,} &  \\
    \rowcolor{oursbg}
    \multirow{-2}{*}{\textbf{Ours}}
      & \multirow{-2}{*}{\textbf{Instruction}}
      & \multirow{-2}{*}{\textbf{29}}
      & \textbf{GiGPO, StraTA}
      & \multirow{-2}{*}{\textbf{Training-free}} \\
    \bottomrule
  \end{tabular}}
\vspace{-5pt}
\caption{Comparison with prior work on LLM agent robustness in terms of perturbed component, instruction paraphrase types, RL algorithms, and whether mitigation requires training. $^\dagger$Evaluated on VLA models.}
  \label{tab:relatedwork_framing}
\vspace{-10pt}
\end{table}

%% file: tables/alfworldpara_example.tex
\definecolor{paraor}{HTML}{B3540E}
\newcommand{\ed}[1]{\textit{\textcolor{paraor}{#1}}}
\begin{table}[t]
\centering\footnotesize
\setlength{\tabcolsep}{4pt}
\renewcommand{\arraystretch}{0.95}
\begin{tabular}{@{}ll@{}}
\toprule
\textbf{Type} & \textbf{Instruction} \\
\midrule
\textit{Original} & put a book in sidetable \\
\midrule
\multicolumn{2}{@{}l}{\scriptsize\textbf{ACTION}\ $\cdot$\ Lexical} \\
Addition       & \ed{directly put a} book \ed{in the} sidetable \\
Same-Polarity  & \ed{set a} book \ed{in the} sidetable \\
\addlinespace[3pt]
\multicolumn{2}{@{}l}{\scriptsize\textbf{ACTION}\ $\cdot$\ Structural} \\
Coordination   & \ed{grasp a} book \ed{and put it in the} sidetable \\
Subordination  & \ed{having found a} book\ed{, put it in the} sidetable \\
\addlinespace[3pt]
\multicolumn{2}{@{}l}{\scriptsize\textbf{ACTION}\ $\cdot$\ Pragmatic} \\
Need        & \ed{I want a} book \ed{inside the} sidetable \\
Embedded    & \ed{could you put a} book \ed{in the} sidetable\ed{?} \\
Permission  & \ed{may I get a} book \ed{moved into the} sidetable\ed{?} \\
Question    & \ed{is the} book \ed{in the} sidetable \ed{yet?} \\
Hint        & \ed{the} sidetable \ed{is currently missing a} book \\
\addlinespace[3pt]
\midrule
\multicolumn{2}{@{}l}{\scriptsize\textbf{OBJECT}\ $\cdot$\ Lexical} \\
Addition       & put a \ed{hardcover book} in the \ed{accent side table} \\
Same-Polarity  & put a \ed{novel} in the \ed{end table} \\
\bottomrule
\end{tabular}
\vspace{-5pt}
\caption{Paraphrase taxonomy in ALFWorld-Para. 
Colored italic text marks the span controlled by each axis; upright text is fixed. 
}
\label{tab:alfworldpara_example}
\end{table}

%% file: sections/03_alfworldpara.tex
\section{ALFWorld-Para}
\vspace{-2pt}
\label{sec:ALFWorld-Para}

\subsection{Benchmark Generation}

Goal instructions in ALFWorld~\cite{shridhar2020alfworld} are instantiated from a limited set of predefined task templates with entity-specific slot fillers (e.g., \texttt{put a [ITEM] in [RECEP]}). Consequently, training and evaluation instructions follow identical linguistic structures, differing only in entities. This template-based structure makes ALFWorld particularly suitable for controlled paraphrase evaluation. Environments with free-form instructions, such as WebShop~\cite{yao2022webshop}, do not permit such evaluation.
We construct ALFWorld-Para by adapting the linguistically grounded paraphrase taxonomy of LIBERO-Para~\cite{liberopara} to ALFWorld and applying it to 226 unique goal instructions from the seen split. This setting isolates instruction phrasing as the sole source of variation relative to training and aligns with the standard evaluation protocol for RL-finetuned agents~\cite{feng2026gigpo, xue2026strata}. As summarized in Table~\ref{tab:alfworldpara_example}, the taxonomy varies action and object expressions along distinct linguistic categories. Because these transformations target separate instruction spans, they can be combined to yield 29 paraphrase types and 6{,}554 paraphrased instructions, as detailed in Appendix~\ref{appen:ALFWorld-Para}. All paraphrases are designed to preserve the original task semantics and uniquely identify their intended referents in the environment. However, under partial observability, the agent knows only the object labels it has encountered, and must therefore locate each referent through exploration. We assume that each instruction specifies the user's intent without requiring subsequent clarification.

\input{tables/alfworldpara_sr_drop}

\subsection{Experimental Results}
\input{figures/analysis_object_drop}
We evaluate sLLMs RL-finetuned with PPO~\cite{schulman2017ppo}, GRPO~\cite{deepseek-math_grpo}, GiGPO~\cite{feng2026gigpo}, and StraTA~\cite{xue2026strata} on shared Qwen2.5-1.5B/7B backbones, on ALFWorld and ALFWorld-Para. For reference, we evaluate pretrained LLMs with ReAct~\cite{yao2023react}: Qwen2.5-7B~\cite{qwen2025qwen25technicalreport}, Llama3.1-8B~\cite{grattafiori2024llama}, Qwen3-14B~\cite{yang2025qwen3}, Phi-4-14B~\cite{abdin2024phi}, and Mistral-Small-3-24B~\cite{mistralsmall3}. Details are in Appendix~\ref{appen:ALFWorld-Para:Setup}. All models degrade substantially on ALFWorld-Para, with relative SR drops of 21--52\% (Table~\ref{tab:alfworldpara_sr_drop}). Although RL-finetuned sLLMs outperform much larger pretrained LLMs on the original ALFWorld, this advantage largely vanishes under paraphrased instructions. For instance, GiGPO-7B surpasses Qwen3-14B by 13.7\,pp on ALFWorld (95.8 vs.\ 82.1), but the margin shrinks to 1.7\,pp on ALFWorld-Para (61.8 vs.\ 60.1). Degradation also varies across RL algorithms, reversing their rankings. StraTA-1.5B has the highest original SR among 1.5B models but suffers a 41.9\% relative drop, falling behind GiGPO-1.5B. Similarly, PPO-7B leads GRPO-7B by 10.2\,pp on ALFWorld but trails it on ALFWorld-Para. Thus, gains on template-based benchmarks do not necessarily transfer to paraphrased instructions. Scale does not prevent this degradation. Even the largest model, Mistral-Small-3-24B, shows a 20.9\% relative drop. Pretrained 7--8B LLMs suffer larger drops of 43--46\%, although their lower original SRs make it hard to disentangle paraphrase sensitivity from baseline instability. Section~\ref{sec:analysis} examines action and object paraphrases separately to identify which transformations drive these losses.

%% file: tables/alfworldpara_sr_drop.tex
\begin{table}[t]
  \centering
  \scriptsize
  \setlength{\tabcolsep}{5.5pt}
  \renewcommand{\arraystretch}{0.95}
  \resizebox{\columnwidth}{!}{%
  \begin{tabular}{lcc>{\color{dropred}}c}
    \toprule
    \multicolumn{1}{c}{Model} & ALFWorld & ALFWorld-Para & SR Drop (\%) \\
    \midrule\noalign{\vskip-\belowrulesep}
    \rowcolor{groupbg}
    \multicolumn{4}{l}{\textit{Pretrained (Prompting method: ReAct)}} \\
    Mistral-S-24B & 80.7 & 63.8 & $-$20.9 \\
    Qwen3-14B     & 82.1 & 60.1 & $-$26.8 \\
    Phi-4-14B     & 63.9 & 49.0 & $-$23.3 \\
    Llama3.1-8B   & 31.4 & 17.8 & $-$43.3 \\
    Qwen2.5-7B    & 54.0 & 29.0 & $-$46.3 \\
    \midrule\noalign{\vskip-\belowrulesep}
    \rowcolor{groupbg}
    \multicolumn{4}{l}{\textit{RL-finetuned (Base model: Qwen2.5)}} \\
    \rowcolor{basebg}
    Base-7B       & 24.2 & 15.3 & $-$36.8 \\
    PPO-7B        & 86.0 & 49.8 & $-$42.1 \\
    GRPO-7B       & 75.8 & 50.9 & $-$32.8 \\
    GiGPO-7B      & 95.8 & 61.8 & $-$35.5 \\
    StraTA-7B     & 92.1 & 62.5 & $-$32.1 \\
    \rowcolor{basebg}
    Base-1.5B     &  5.6 &  3.1 & $-$44.6 \\
    PPO-1.5B      & 56.5 & 27.1 & $-$52.0 \\
    GRPO-1.5B     & 73.4 & 46.7 & $-$36.4 \\
    GiGPO-1.5B    & 84.4 & 53.8 & $-$36.3 \\
    StraTA-1.5B   & 91.6 & 53.2 & $-$41.9 \\
    \bottomrule
  \end{tabular}%
  }
  \vspace{-5pt}
  \caption{Success rate (SR) (\%) on ALFWorld and ALFWorld-Para. SR Drop: relative change.}
  \vspace{-5pt}
  \label{tab:alfworldpara_sr_drop}
\end{table}

%% file: figures/analysis_object_drop.tex

\begin{figure*}[t]
  \centering
  \includegraphics[width=\textwidth]{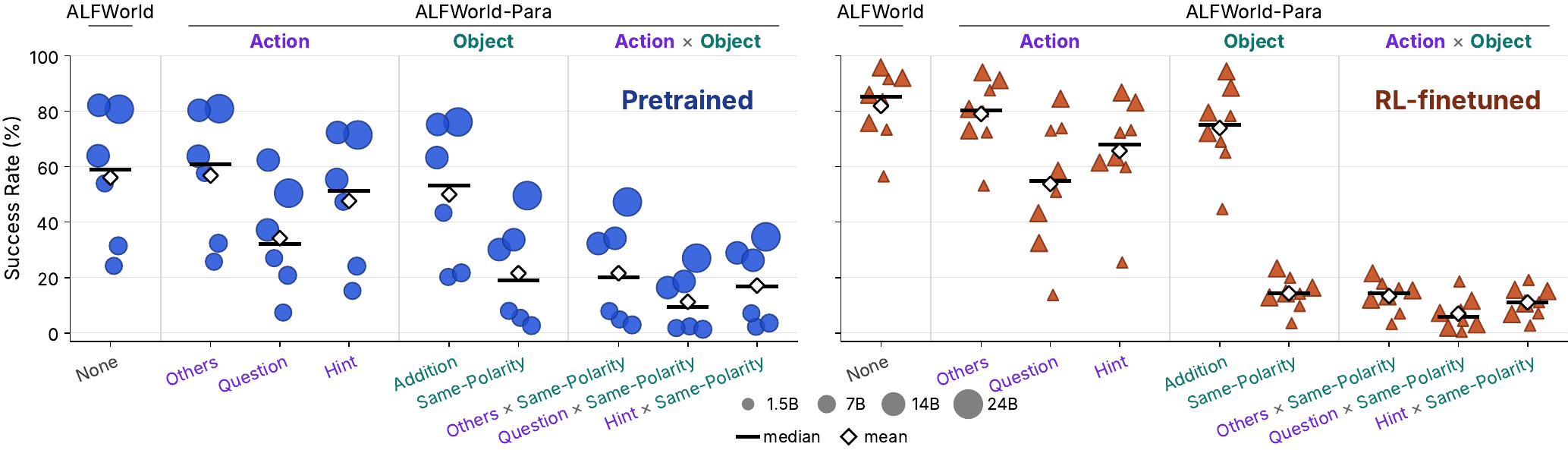}
  \vspace{-19pt}
  \caption{Success rate (\%) of pretrained (left) and RL-finetuned (right) LLMs on original ALFWorld (None) and on ALFWorld-Para, grouped by paraphrase axis and type. Each marker is one model, with marker size indicating parameter count; horizontal bars and diamonds denote the median and mean across models. 
  }
  \vspace{-1pt}
  \label{fig:analysis_object_drop}
\end{figure*}

%% file: sections/04_analysis.tex
\section{Analysis}
\label{sec:analysis}
\vspace{-2pt}

\subsection{Analysis 1: Object Synonym Substitution Is the Dominant Failure Factor}
\label{analysis:analysis1}

Figure~\ref{fig:analysis_object_drop} presents SRs across paraphrase types for pretrained LLMs and RL-finetuned sLLMs. As defined in \S\ref{sec:ALFWorld-Para}, Addition preserves the original object word, whereas Same-Polarity (SP) replaces it with a co-referring name. Pretrained LLMs exhibit moderate, broadly distributed degradation across types. In contrast, RL-finetuned sLLMs maintain high SRs under Others (action axis) and Addition, but collapse sharply in columns involving SP.


Along the Object axis, Addition causes only minor SR drops in both groups, with RL-finetuned sLLMs retaining their advantage over pretrained LLMs. Under SP, however, their median SR falls below that of pretrained LLMs, whose performance varies widely across models, while RL-finetuned sLLMs consistently cluster at low SRs regardless of model size or RL algorithm. This indicates that RL gains diminish when the original object word is replaced, revealing strong lexical dependence.


Along the Action axis, Others, comprising seven types that retain an action verb or explicit request, achieves SRs comparable to None in both groups. Question and Hint depart from imperative phrasing and require intent inference (Table~\ref{tab:alfworldpara_example}), causing the largest median SR declines. RL-finetuned sLLMs retain higher median SRs but exhibit substantially greater inter-model variation than under Others. Non-imperative phrasing leads to varying degrees of degradation across RL-finetuned sLLMs, while object synonym substitution results in consistently low SRs. Combining both axes (Action$\times$Object) compounds these effects, reducing overall SRs to or below the SP level.


Why, then, do RL-finetuned sLLMs consistently fail under SP?
One hypothesis is that fine-tuning on a narrow, template-based distribution erodes their knowledge that paraphrased and original object expressions refer to the same entity. If this holds, RL-finetuned sLLMs should exhibit weaker referential knowledge than their pretrained backbones. 


\subsection{Analysis 2: sLLMs Still Know What the Paraphrased Object Refers to}
\label{analysis:analysis2}
\input{figures/analysis_paraknow}

To test this hypothesis, we conduct a referential knowledge probe that asks each model to match an SP-substituted expression to its original ALFWorld label from the complete label list. We define paraphrase knowledge as the proportion of correct selections (chance level: 1.2\%; Appendix~\ref{appen:analysis:analysis2}). Figure~\ref{fig:analysis_paraknow} shows pretrained LLMs exhibit high paraphrase knowledge, with even Qwen2.5-1.5B substantially exceeding chance. Comparing Qwen2.5-7B and 1.5B with their RL-finetuned counterparts reveals negligible differences across RL algorithms. Thus, paraphrase knowledge is largely preserved despite fine-tuning on narrow templates and fixed action formats, contradicting the hypothesis.




This knowledge, however, diverges from task performance in \S\ref{analysis:analysis1}. RL-finetuned 7B models achieve nearly 80\% paraphrase knowledge but typically below 20\% SR under SP-paraphrased instructions. They match or outperform pretrained LLMs with comparable knowledge on original ALFWorld, yet lose this advantage under SP. Despite substantially higher paraphrase knowledge in 7B than 1.5B models, both exhibit similarly low SRs under SP. These results suggest that the degradation is not attributable to knowledge loss during RL fine-tuning. We therefore examine agent trajectories to understand why failures occur.




\subsection{Analysis 3: Agents Encounter the Target Object but Do Not Take It}
\label{analysis:analysis3}

An agent must observe the target object before taking it and proceeding with the task. Figure~\ref{fig:analysis_seennottake} reports the taken rate: across all episodes of
each object-axis paraphrase type, the fraction of steps with the target observed and
not yet taken, up to and including its take, at which the agent takes it (Appendix~\ref{appen:analysis:analysis3}). Both groups generally take the target under None and Addition, whereas SP markedly reduces taken rates despite target observation. RL fine-tuning consistently improves taken rates over pretrained backbones under None and Addition, but not SP; some algorithms at 7B even fall below the backbone (41.3\%). Under SP, taken rates also vary widely across RL algorithms despite identical backbones, with rankings reversing across model sizes. Thus, agents are less consistent in taking objects referred to by synonyms than those specified by their original labels.

\input{figures/analysis_seennottake}
\input{figures/method_figure}

Agent performance involves three stages: identifying the referent of an instruction expression, observing it during exploration, and binding the expression to the observed object for execution. \S\ref{analysis:analysis1} identifies object synonyms as the dominant failure factor; \S\ref{analysis:analysis2} confirms preserved referential knowledge; and \S\ref{analysis:analysis3} shows agents encounter target objects and readily take them under original labels. These findings point to a difficulty in binding paraphrased expressions to encountered objects. RL fine-tuning preserves referential knowledge and taking behavior for original labels, but its gains do not generalize to this binding process.

%% file: figures/analysis_paraknow.tex

\begin{figure}[t]
  \centering
  \includegraphics[width=\columnwidth]{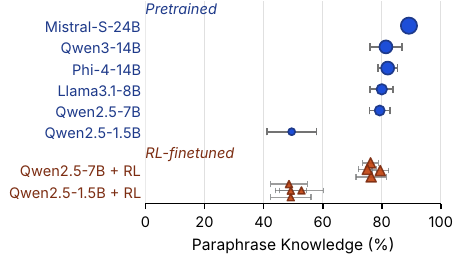}
  \vspace{-20pt}
    \caption{Paraphrase knowledge (\%) of reference pretrained LLMs (top) and RL-finetuned sLLMs (bottom). Each RL row shows four algorithms on one backbone.}
    \vspace{-2pt}
  \label{fig:analysis_paraknow}
\end{figure}

%% file: figures/analysis_seennottake.tex
\begin{figure}[t]
  \centering
  \includegraphics[width=\columnwidth]{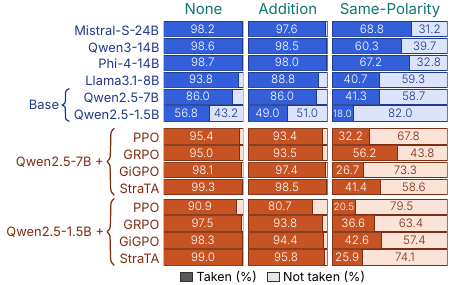}
  \vspace{-19pt}
  \caption{Taken rate (\%): among steps in which the target object is observed, up to and including its first take, the proportion of those at which the agent takes it.}
  \label{fig:analysis_seennottake}
  \vspace{-3pt}
\end{figure}


%% file: figures/method_figure.tex
\begin{figure*}[t]
  \centering
  \includegraphics[width=\textwidth]{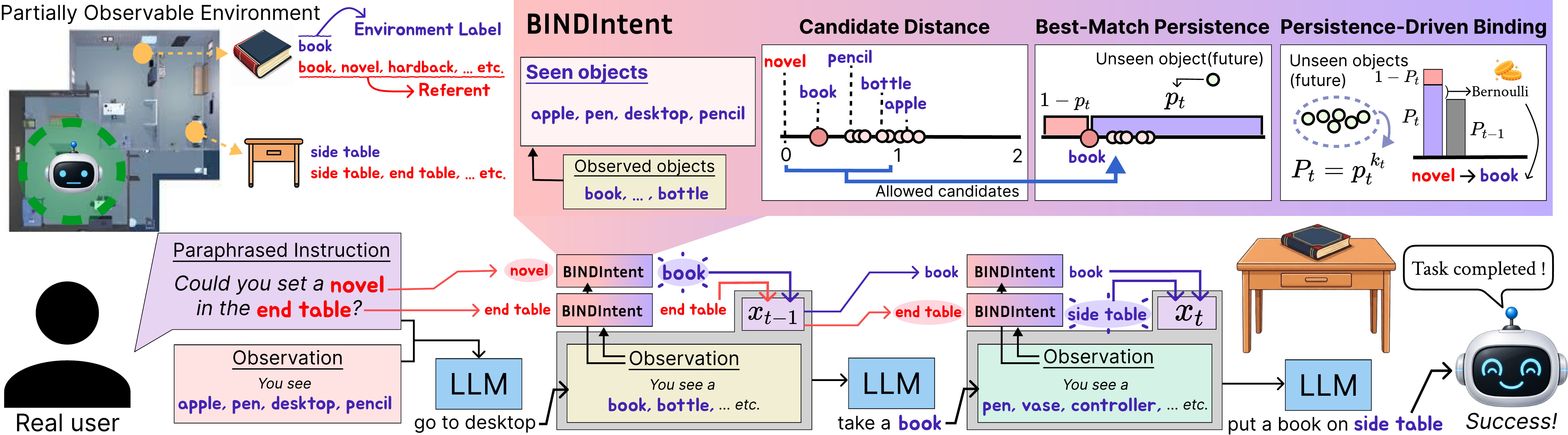}
  \vspace{-14pt}
  \captionsetup{width=\textwidth}
  \caption{
  Overview of BINDIntent. The framework binds paraphrased object expressions to observed environment labels and forwards the rewritten instruction to a frozen agent.
  }
  \vspace{-4pt}
  \label{fig:method}
\end{figure*}

%% file: sections/05_method.tex
\section{Proposed Method}
\label{sec:method}

\vspace{-2pt}

\subsection{Motivating Observation}
\label{method:motivation}
The findings in \S\ref{sec:analysis} suggest that binding failures stem from partial observability rather than a lack of referential knowledge. When all environment object labels are available, models correctly identify the intended referents (\S\ref{analysis:analysis2}). During execution, however, agents encounter objects incrementally and must determine whether a matching label remains unseen or an observed object corresponds to the instruction (\S\ref{analysis:analysis3}). Given our focus on instructions that sufficiently specify the user's intent (\S\ref{sec:ALFWorld-Para}), this uncertainty lies in what the agent has observed rather than in the instruction. Exhaustive exploration could remove the uncertainty but only at substantial interaction cost. Motivated by these observations, we target object synonym substitutions, the dominant source of performance degradation across model sizes and RL algorithms (\S\ref{analysis:analysis1}), and delegate referent binding to an external framework, allowing agents to focus on task execution.

\subsection{BINDIntent: Binding Instruction Objects in Partially Observable Environments}
\label{method:BindIntent}

\subsubsection{Problem Setup and Notation}
\label{method:BindIntent:ProblemSetup}


We extract object referring expressions from instruction $x$ as a set of slots $\mathcal{S}(x) = \{s^{(1)}, \dots, s^{(I)}\}$. For example, ``Could you set a \underline{novel} in the \underline{end table}?'' yields $s^{(1)}=\textit{novel}$ and $s^{(2)}=\textit{end table}$. Let $V_t$ be the object labels observed by step $t$, and define a unit as an object that can contain others (e.g., a drawer or cabinet). Let $k_t$ count unexplored units that may reveal new objects. For each slot, BINDIntent identifies the closest object in $V_t$, termed the best match $v^\star_t$, as the most likely referent of the user's expression. At each step, it decides whether to bind the slot to $v^\star_t$. Once bound, the slot is replaced with $v^\star_t$ before the instruction is passed to the frozen agent. This decision depends on the confidence in $v^\star_t$ given the instruction, accumulated observations, and unexplored portions of the environment. We formalize this process in three stages: candidate distance (\S\ref{sec:scoring}), best-match persistence (\S\ref{sec:persistence}), and persistence-driven binding (\S\ref{sec:binding}), as shown in Figure~\ref{fig:method}. Since slots are processed independently, we omit the slot index $(i)$.

\subsubsection{Candidate Distance}
\label{sec:scoring}


Given a word embedding $E$, we define the cosine distance between two words $w_1$ and $w_2$ as
\begin{equation}
\delta(w_1, w_2) = 1 - \cos\big(E(w_1), E(w_2)\big).
\label{eq:distance}
\end{equation}
We use ConceptNet Numberbatch~\citep{speer2017conceptnet} as the default embedding, selecting the object in $V_t$ with the smallest distance as the best match $v^\star_t$. Before selecting $v^\star_t$, we subtract the mean embedding of $V_t$ to remove scene-specific common components (Appendix~\ref{appen:Method:Candidate_Scoring}). We consider only objects within a lexical allowance $\tau_t$ of the slot as candidate referents, collecting their distances as
\begin{equation}
D_t = \big( \delta(s,v) : v \in V_t,\ \delta(s,v) \le \tau_t \big).
\label{eq:candidates}
\end{equation}
No binding occurs if no candidate referent exists. Early in an episode, an object whose label exactly matches the user's expression may remain in an unexplored unit. We thus initially permit only exact matches ($\tau_0 = 0$), avoiding premature binding to differently named objects. 
The lexical allowance then increases by a fixed step size $\Delta\tau$, $\tau_t = t\,\Delta\tau$, gradually admitting objects with different labels.

\input{tables/experiment_main}

\subsubsection{Best-Match Persistence under Partial Observability}
\label{sec:persistence}

The current best match $v^\star_t$ is optimal among observed objects, yet a closer candidate may reside in an unexplored unit. To quantify this uncertainty, we define the best-match distance as $d^\star_t = \delta(s, v^\star_t)$ and estimate $p_t = \Pr[\, d > d^\star_t \mid D_t \,]$, the probability that an unseen candidate has a greater distance. Since distances are bounded below by zero, the likelihood of encountering a closer candidate is governed by the lower tail of the distance distribution~\cite{coles2001introduction}. We therefore model candidate distances with a parsimonious Weibull distribution~\cite{weibull1951statistical}:
\begin{equation}
\Pr[d>z\mid\beta,\eta] = \exp\left[-(z/\eta)^\beta\right],
\label{eq:weibull}
\end{equation}
where $\eta$ and $\beta$ denote the scale and shape parameters, respectively. Given observed candidate distances $D_t$ and the prior $\pi(\beta,\eta)\propto 1/(\beta\eta)$, we infer the posterior $\pi(\beta,\eta\mid D_t)$. The posterior predictive probability that an unseen candidate has a greater distance than the current best match is
\begin{equation}
p_t = \iint e^{-(d_t^\star/\eta)^\beta}\,
\pi(\beta,\eta\mid D_t)\,d\beta\,d\eta,
\label{eq:predictive}
\end{equation}
where $e^{-(d_t^\star/\eta)^\beta}$ denotes the Weibull survival function at $d_t^\star$. We analytically marginalize $\eta$ and numerically integrate over $\beta$. When fewer than $c$ candidates are available, we adopt an exchangeability-based fallback to ensure stable estimation. Detailed derivations are provided in Appendix~\ref{app:weibull-derivation}.

\subsubsection{Persistence-Driven Binding}
\label{sec:binding}

The probability $p_t$ represents the likelihood that the current best match remains superior to a single unseen candidate, while $1-p_t$ denotes the probability of encountering a closer candidate. For the current best match to persist, it must remain superior to all future candidates. Since the number of unseen objects is unknown, we treat each of the $k_t$ unexplored units as one candidate opportunity, yielding the persistence probability $P_t = p_t^{k_t}$. $P_t$ increases as the best match becomes more distinct or fewer units remain unexplored. BINDIntent progressively commits to binding as this confidence increases. Whenever $P_t$ exceeds its previous maximum, the excess is added to the cumulative binding probability. Thus, letting $t_s$ denote the step at which slot $s$ is bound, the cumulative binding probability equals the running maximum of $P_t$:
\begin{equation}
\Pr[t_s \le t] = \max_{u \le t} P_u.
\label{eq:target}
\end{equation}
Here, $u$ ranges over the steps up to $t$. Binding determines whether a slot is replaced with an observed object label. For each unbound slot, a Bernoulli trial determines whether binding occurs at each step, with success probability given by the increment in cumulative binding probability divided by the remaining unbound probability mass, as detailed in Appendix~\ref{appen:Method-Persistence-Driven-Binding}. Once bound, the slot uses the current best match as its replacement label at each step, as specified in Algorithm~\ref{alg:method}.





%% file: tables/experiment_main.tex
\newcolumntype{Q}{w{c}{16pt}}
  \renewcommand{\arraystretch}{0.95}
\begin{table*}[t]
  \centering
  \scriptsize
  \setlength{\tabcolsep}{1.8pt}
  \renewcommand{\arraystretch}{0.95}
  \begin{tabular}{l!{\color{black}\vrule}QQQQ!{\color{black}\vrule}QQQQ!{\color{black}\vrule}QQQQ!{\color{black}\vrule}QQQQ!{\color{black}\vrule}>{\columncolor{oursbg}}Q>{\columncolor{oursbg}}Q>{\columncolor{oursbg}}Q>{\columncolor{oursbg}}Q}
    \toprule
    \multicolumn{1}{c}{Model}
      & \multicolumn{4}{c}{Baseline}
      & \multicolumn{4}{c}{Greedy}
      & \multicolumn{4}{c}{Self-clarification}
      & \multicolumn{4}{c}{BINDIntent\textsubscript{SBERT}}
      & \multicolumn{4}{c}{\cellcolor{oursbg}BINDIntent (ours)} \\
    \cmidrule(lr){2-5} \cmidrule(lr){6-9} \cmidrule(lr){10-13} \cmidrule(lr){14-17} \cmidrule(lr){18-21}
    \multicolumn{1}{r}{\tiny Split =} & \multicolumn{1}{c}{\tiny None} & \multicolumn{1}{c}{\tiny Add} & \multicolumn{1}{c}{\tiny Same-P} & \multicolumn{1}{c}{\tiny All} & \multicolumn{1}{c}{\tiny None} & \multicolumn{1}{c}{\tiny Add} & \multicolumn{1}{c}{\tiny Same-P} & \multicolumn{1}{c}{\tiny All} & \multicolumn{1}{c}{\tiny None} & \multicolumn{1}{c}{\tiny Add} & \multicolumn{1}{c}{\tiny Same-P} & \multicolumn{1}{c}{\tiny All} & \multicolumn{1}{c}{\tiny None} & \multicolumn{1}{c}{\tiny Add} & \multicolumn{1}{c}{\tiny Same-P} & \multicolumn{1}{c}{\tiny All} & \multicolumn{1}{c}{\cellcolor{oursbg}\tiny None} & \multicolumn{1}{c}{\cellcolor{oursbg}\tiny Add} & \multicolumn{1}{c}{\cellcolor{oursbg}\tiny Same-P} & \multicolumn{1}{c}{\cellcolor{oursbg}\tiny All} \\
    \midrule
    \rowcolor{groupbg}
    \multicolumn{21}{l}{\textit{Pretrained (Prompting method: ReAct)}} \\
    Mistral-S-24B & 76.5 & 72.0 & 44.2 & 63.8 & 38.1 & 38.2 & 28.7 & 34.9 & 70.2 & 68.8 & 55.0 & 64.5 & 71.6 & 70.4 & 44.6 & 61.9 & 73.7 & 70.9 & 56.5 & \textbf{66.8} \\
    Qwen3-14B     & 77.5 & 72.9 & 31.7 & 60.1 & 39.0 & 37.9 & 27.8 & 34.7 & 66.6 & 63.4 & 49.2 & 59.5 & 72.2 & 68.6 & 37.6 & 59.1 & 73.3 & 69.1 & 52.8 & \textbf{64.8} \\
    Phi-4-14B     & 59.9 & 58.0 & 30.1 & 49.0 & 30.5 & 30.1 & 22.0 & 27.4 & 60.3 & 59.1 & 40.8 & \textbf{53.2} & 55.3 & 56.5 & 32.5 & 47.8 & 56.6 & 55.9 & 44.9 & 52.3 \\
    Llama3.1-8B   & 30.2 & 21.6 & \phantom{0}2.8 & 17.8 & 15.2 & 14.5 & 11.7 & 13.7 & 25.1 & 28.0 & 15.8 & \textbf{22.9} & 24.5 & 23.0 & 8.7 & 18.5 & 29.2 & 25.7 & 12.4 & 22.2 \\
    Qwen2.5-7B      & 46.4 & 36.3 & \phantom{0}5.9 & 29.0 & 24.9 & 28.2 & 19.8 & 24.3 & 50.3 & 48.8 & 21.0 & 39.7 & 50.9 & 48.6 & 19.6 & 39.3 & 50.1 & 46.7 & 25.4 & \textbf{40.4} \\
    \midrule
    \rowcolor{groupbg}
    \multicolumn{21}{l}{\textit{RL-finetuned (Base model: Qwen2.5)}} \\
    \rowcolor{basebg}
    Base-7B       & 22.5 & 19.7 & \phantom{0}4.4 & 15.3 & 12.6 & 11.5 & \phantom{0}8.3 & 10.7 & 20.9 & 22.3 & \phantom{0}9.6 & 17.5 & 22.8 & 22.0 & \phantom{0}8.8 & 17.7 & \cellcolor{oursbg}22.0 & \cellcolor{oursbg}21.2 & \cellcolor{oursbg}12.0 & \cellcolor{oursbg}\textbf{18.3} \\
    PPO-7B        & 73.5 & 67.5 & 10.6 & 49.8 & 41.0 & 37.1 & 29.6 & 35.7 & 68.5 & 70.3 & 21.1 & 52.8 & 71.1 & 68.9 & 28.4 & 55.6 & 70.6 & 69.2 & 46.3 & \textbf{61.7} \\
    GRPO-7B       & 68.4 & 66.2 & 19.7 & 50.9 & 41.9 & 40.2 & 28.5 & 36.7 & 63.3 & 65.1 & 31.3 & 52.9 & 67.4 & 67.8 & 30.8 & 54.9 & 67.4 & 67.9 & 43.8 & \textbf{59.4} \\
    GiGPO-7B      & 89.2 & 86.8 & 12.1 & 61.8 & 62.5 & 58.7 & 41.9 & 54.1 & 86.7 & 87.6 & 23.9 & 65.3 & 88.2 & 88.3 & 35.1 & 69.9 & 88.9 & 86.8 & 57.6 & \textbf{77.4} \\
    StraTA-7B     & 89.6 & 85.5 & 15.0 & 62.5 & 67.2 & 63.3 & 51.4 & 60.4 & 85.0 & 84.2 & 41.9 & 69.9 & 90.0 & 87.3 & 29.0 & 68.0 & 89.2 & 86.5 & 44.3 & \textbf{72.8} \\
    \rowcolor{basebg}
    Base-1.5B     & \phantom{0}4.7 & \phantom{0}3.5 & \phantom{0}1.2 & \phantom{0}3.1 & \phantom{0}4.9 & \phantom{0}3.5 & \phantom{0}2.9 & \phantom{0}3.7 & \phantom{0}2.9 & \phantom{0}3.1 & \phantom{0}1.1 & \phantom{0}2.4 & \phantom{0}4.7 & \phantom{0}4.5 & \phantom{0}2.6 & \phantom{0}\textbf{3.9} & \cellcolor{oursbg}\phantom{0}5.3 & \cellcolor{oursbg}\phantom{0}4.1 & \cellcolor{oursbg}\phantom{0}2.0 & \cellcolor{oursbg}\phantom{0}3.8 \\
    PPO-1.5B      & 45.7 & 34.6 & \phantom{0}2.9 & 27.1 & 27.2 & 28.0 & 20.2 & 25.1 & 35.4 & 32.8 & \phantom{0}6.1 & 24.4 & 44.8 & 42.6 & 15.2 & 33.8 & 44.0 & 39.0 & 20.4 & \textbf{34.1} \\
    GRPO-1.5B     & 68.6 & 59.0 & 14.7 & 46.7 & 49.5 & 47.0 & 37.0 & 44.3 & 55.5 & 52.8 & 13.5 & 40.1 & 68.6 & 66.4 & 26.7 & 53.4 & 68.2 & 64.5 & 39.1 & \textbf{56.9} \\
    GiGPO-1.5B    & 78.3 & 67.2 & 18.2 & 53.8 & 68.9 & 65.0 & 54.2 & 62.5 & 63.3 & 58.9 & 17.6 & 46.0 & 77.4 & 76.9 & 33.4 & 62.1 & 78.2 & 73.3 & 49.6 & \textbf{66.6} \\
    StraTA-1.5B   & 84.4 & 71.3 & \phantom{0}7.0 & 53.2 & 81.9 & 76.4 & 63.7 & 73.7 & 75.6 & 67.8 & 11.5 & 50.8 & 85.7 & 85.9 & 33.9 & 67.9 & 85.2 & 81.2 & 56.1 & \textbf{73.8} \\
    \specialrule{\heavyrulewidth}{0pt}{0pt}
  \end{tabular}
  \vspace{-7pt}
  \caption{Success rate (\%) on ALFWorld-Para by object-axis paraphrase type: None (object unchanged), Add (Addition), Same-P (Same-Polarity), and All. 
  Best All score per row in \textbf{bold}. Averaged over three seeds.}
  \label{tab:experiment_main}
   \vspace{-6pt}
\end{table*}

%% file: sections/06_experiment.tex
\section{Experiments}
\label{sec:experiment}

\vspace{-2pt}

We evaluate BINDIntent against three baselines and one embedding variant. All methods share parsed slots and observed objects, differing only in referent selection and instruction modification. Baseline retains the paraphrased instruction, leaving grounding to the LLM. Greedy uses BINDIntent's embedding distance but immediately binds each slot to its closest observed object. Self-clarification adapts query-based disambiguation~\cite{ren2023knowno, ramrakhya2025asktoact} without user interaction. At each step, the agent LLM selects a referent from observed objects or defers, replacing the expression upon selection. BINDIntent\textsubscript{SBERT} replaces relation-aware Numberbatch~\cite{speer2017conceptnet} with general-purpose SBERT embeddings~\cite{reimers2019sentence} to assess the contribution of relation-aware representations. We use the same LLMs as in \S\ref{sec:ALFWorld-Para} (prompts and hyperparameters in Appendix~\ref{appen:Experiment:Setup}). We report SRs by object-axis type (None, Addition, Same-Polarity) and overall (All).



\subsection{Main Results}
\label{sec:MainResults}
Table~\ref{tab:experiment_main} shows that BINDIntent improves SP by 17.5--49.1\,pp over Baseline for RL-finetuned sLLMs, while None varies by at most 3\,pp and Addition is maintained or improved. Overall SR (All) increases by 7.0--20.6\,pp. GiGPO-7B (77.4) and StraTA-1.5B (73.8) surpass Mistral-Small-3-24B with and without BINDIntent (66.8 and 63.8, respectively; Appendix~\ref{appen:Experiment:MainResults}).
\input{figures/experiment_latency}
Base models show limited gains (+3.0, +0.7), attributable to low SR even on original instructions (24.2, 5.6; Table~\ref{tab:alfworldpara_sr_drop}), indicating that task-solving capability remains the primary limitation. Greedy highlights the importance of accounting for unseen objects. Despite achieving higher SP on some models (StraTA-7B, GiGPO-1.5B, StraTA-1.5B), premature substitutions degrade None and Addition (e.g., GiGPO-7B None: 89.2 $\rightarrow$ 62.5).
Consequently, Greedy underperforms BINDIntent in All across RL-finetuned sLLMs and falls below Baseline for every pretrained LLM. The gap is marginal for StraTA-1.5B (73.7 vs.\ 73.8), whose vague initial strategies sustain exploration despite premature substitutions (Appendix~\ref{appen:Experiment:MainResults}). BINDIntent mitigates premature binding by explicitly modeling the likelihood of superior unseen candidates.

Self-clarification depends on the agent's referential knowledge. It improves All for 7B RL-finetuned sLLMs and pretrained LLMs except Qwen3-14B (−0.6 pp), but underperforms Baseline for all 1.5B models, consistent with knowledge probes (\S\ref{analysis:analysis2}: approximately 50\% for 1.5B vs. 78\% for 7B). Incorrect selections degrade even None (e.g., GRPO-1.5B: 68.6 $\rightarrow$ 55.5), while each decision incurs an additional LLM call. BINDIntent\textsubscript{SBERT} improves SP, but achieves roughly half the gains of Numberbatch (e.g., GiGPO-7B: +23.0 vs. +45.5\,pp) and occasionally underperforms Baseline on larger pretrained LLMs. Numberbatch, built on ConceptNet relations, appears better at separating synonyms from merely related words. BINDIntent also improves All (+3.0--11.4\,pp) and SP (+9.6--21.1\,pp) across all reference pretrained LLMs. However, None declines by 2.8--4.2\,pp for models with at least 14B parameters, showing some substitution cost for models already strong on the original words.

All compared methods are executed before agent inference at each step, directly contributing to user-perceived response latency. Figure~\ref{fig:experiment_latency} presents the per-step latency of each method measured independently of agent inference. Greedy, BINDIntent, and BINDIntent\textsubscript{SBERT} require no agent LLM calls, rendering their latency independent of backbone size. BINDIntent incurs only 0.38\,ms per step, comparable to Greedy (0.25\,ms). Substituting SBERT increases latency to 2.9\,ms, primarily attributable to encoder forward passes for newly observed object names. By comparison, Self-clarification requires an agent LLM call per decision, with latency increasing from approximately 30\,ms at 1.5B to 107\,ms at 24B, representing 80--281$\times$ the latency of BINDIntent (details in Appendix~\ref{appen:Experiment:Latency}).  



\subsection{Cross-Benchmark Validation}
To assess generalization beyond ALFWorld, we construct ScienceWorld-Para from ScienceWorld~\citep{wang2022scienceworld} using the same paraphrase taxonomy and evaluate four independently RL-finetuned Qwen2.5-7B agents, with detailed analyses in Appendix~\ref{appen:ScienceWorld-Para}. Consistent with ALFWorld-Para, paraphrasing reduces scores by 10.1--20.3\%, primarily due to object synonym substitutions (\S\ref{analysis:analysis1}). Action-axis paraphrases yield scores of 50.4--52.2, comparable to 50.9 for original instructions, whereas SP drops to 29.9. Referential knowledge is likewise preserved after RL fine-tuning (69.8--73.5\% vs. 71.2\% pretrained; \S\ref{analysis:analysis2}). Similarly, RL fine-tuning improves taken rates under None (19.7\% $\rightarrow$ 40.3--87.7\%) and Addition (16.9\% $\rightarrow$ 31.5--84.5\%), but not consistently under SP (16.4\% $\rightarrow$ 4.2--65.9\%; \S\ref{analysis:analysis3}). Without specific tuning, our BINDIntent improves SP by 7.0--10.2\,pp and All by 2.1--3.4\,pp over Baseline. Notably, its per-step overhead is only 0.78\,ms, two orders of magnitude below Self-clarification, which requires an LLM call per decision.

%% file: figures/experiment_latency.tex
\begin{figure}[t]
  \centering
  \includegraphics[width=\columnwidth]{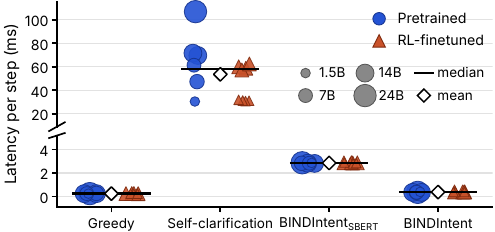}
  \vspace{-10pt}
  \caption{Per-step latency introduced by each method across agent models. Note the discontinuous y-axis.}
  \label{fig:experiment_latency}
  \vspace{-7pt}
\end{figure}

%% file: sections/07_conclusion.tex
\section{Conclusion}
\vspace{-2pt}
This study reveals that paraphrase failures arise from referent binding under partial observability rather than knowledge loss. Such failures evade fully observable probes but emerge during execution, when agents must distinguish unseen referents from previously encountered objects. These observations suggest that template-based benchmarks and knowledge probes alone may not adequately capture this vulnerability. Robustness therefore calls for principled reasoning about unobserved referents without compromising sLLM efficiency, rather than relying solely on fine-tuning or model scaling. BINDIntent substantiates this direction by recovering substantial performance through training-free probabilistic binding.

%% file: sections/08_limitation.tex
\section*{Limitations}
This work primarily focuses on binding failures induced by object synonym substitutions, which exhibit the most pronounced performance degradation (\S\ref{analysis:analysis1}). While BINDIntent addresses the associated binding failures, its applicability to other paraphrase types, particularly action-axis variations such as Question and Hint, remains to be investigated. Furthermore, BINDIntent operates on observed objects without modifying the agent's exploration policy. As exploration behaviors may vary across RL-finetuned agents, their influence on binding performance has not been systematically isolated. Future research could extend BINDIntent by integrating binding with exploration strategies to facilitate the discovery of relevant objects.

%% file: sections/50_appendix_face.tex
\clearpage
\appendix
\onecolumn

{\LARGE\scshape Appendix}\\[0.3em]
\rule{\linewidth}{0.8pt}

\vspace{1.5em}
\startcontents[appendices]
\printcontents[appendices]{}{1}{\setcounter{tocdepth}{3}}

\twocolumn   

\clearpage
\appendix

%% file: sections/93_appen_alfworldpara.tex
\input{appenfigure/Appen_alfworld_template}

\section{ALFWorld-Para}
\label{appen:ALFWorld-Para}
This section describes the paraphrase taxonomy, dataset generation process, dataset statistics, and human evaluation of ALFWorld-Para.

\subsection{Benchmark Generation}
\label{appen:ALFWorld-Para:BenchmarkGeneration}
Figure~\ref{fig:alfworld_example} illustrates an example interaction trajectory in ALFWorld: at each step, the agent receives a textual observation together with the admissible actions and issues a single action toward the goal instruction.
LIBERO-Para~\cite{liberopara} partitions Same-Polarity within the lexical type into two sub-types, contextual and habitual.
Contextual denotes synonyms contingent on the utterance context, whereas habitual denotes synonyms fixed by linguistic convention~\cite{kovatchev-etal-2018-etpc}.
This distinction is pertinent to environments with a rich entity vocabulary, wherein an entity admits varied references depending on context.
The entity vocabulary of ALFWorld, however, consists of simple common nouns such as \texttt{book}, \texttt{mug}, and \texttt{knife}; consequently, the synonym candidates available to the two sub-types largely overlap.
We therefore consolidate the two sub-types into a single Same-Polarity type on the object axis and, for consistency, apply the same consolidation to lexical Same-Polarity on the action axis.

\subsubsection{Paraphrase Dataset Generation}
\label{app:para_generation}
Paraphrases in ALFWorld-Para are produced by factored generation, generating the object axis and the action axis independently before composing them.
We use Gemini 3 Flash~\cite{gemini3} as the generation model.

\input{appenfigure/Appen_alfworld-para_stat}

\textbf{Action Axis.}
ALFWorld goal instructions are built from 12 templates, two for each of the six task types (Table~\ref{tab:alfworld_templates}).
We replace the object and receptacle slots in each template with placeholders and define the remainder as the action frame.
To the action axis prefix (Figure~\ref{fig:prompt-action}), we append a type-specific suffix to generate paraphrased frames (lexical, Figure~\ref{fig:prompt-action-suffixes-lexical}; structural, Figure~\ref{fig:prompt-action-suffixes-structural}; pragmatic, Figure~\ref{fig:prompt-action-suffixes-pragmatic}).

\textbf{Object Axis.}
Given that both the item and the receptacle in an ALFWorld goal instruction constitute noun phrases, we treat both slots as targets of object axis paraphrase.
For each entity, we append a type-specific suffix (Figure~\ref{fig:prompt-object-suffixes}) to the object axis prefix (Figure~\ref{fig:prompt-object}) and generate multiple paraphrase candidates.

\textbf{Verification.}
Generated candidates are filtered through an LLM-based verifier.
We apply a separate verifier prompt (Figure~\ref{fig:prompt-verifier}) to the object axis and the action axis, accepting only candidates that satisfy criteria such as semantic preservation, cell compliance, and naturalness.

\textbf{Composition.}
We compose verified action frames with object paraphrases to form the final instructions.
Since the two axes are generated independently, composition can introduce article disagreement (e.g., ``put a apple'') or number disagreement (e.g., ``find two spray bottle'').
Repairing these post hoc with hand-written rules is brittle across irregular forms (a/an, knife$\to$knives, potato$\to$potatoes).
We therefore generate four inflected forms of each object paraphrase at generation time (bare, indefinite, definite, plural), and adopt a slot-filling scheme that inserts verbatim whichever form the action frame requires.

\subsubsection{Statistics of ALFWorld-Para}
\label{app:para_statistics}
The seen split of ALFWorld contains 240 episodes. Since multiple
episodes may share the same goal instruction under different scenes,
these episodes correspond to 226 unique instructions. To establish a
one-to-one correspondence between each original instruction and its
paraphrases, and to prevent instructions with multiple scenes from
being counted more than once, we fix a single scene per unique
instruction (the first episode in the split order), yielding 226
evaluation episodes.
Since there are 226 original instructions, we generate one paraphrase
per instruction for each type, yielding 226 paraphrases per type and
$226 \times 29$ in total.
Table~\ref{tab:para_statistics} presents the breakdown by axis and type.

\subsubsection{Parser Stability}
\label{parser_stability}
Candidate slot spans are extracted from each instruction by a lexicon-free
procedure. Using the part-of-speech tags of the spaCy \texttt{en\_core\_web\_sm}
pipeline~\citep{honnibal2020spacy}, contiguous spans are grouped into regions and trimmed of closed-class words at both edges, following the Universal Dependencies tagset~\citep{nivre-etal-2020-universal}; no entity, verb, or task-specific word list is employed. We evaluate the extraction against the gold substitutions recorded by the paraphrase generator. Across all 6,554 instructions of ALFWorld-Para, 99.8\%
of the 13,108 gold slots are recovered. We further conducted a manual verification on 300 instructions sampled with stratification by paraphrase type, assessing whether each extracted region denotes the same entity as the corresponding gold slot; all 300 instructions were judged correct. Extraction is performed once per episode, requires 2.2\,ms, and is shared by all binding methods; it therefore does not affect their relative cost.

\subsubsection{Human Evaluation}
\label{app:Alfworld_Human_Evaluation}

\paragraph{Annotation setup.}
We sampled 328 paraphrase--original pairs (5\% of the full 6,554-item dataset) via stratified sampling across the 29 paraphrase types.
Three annotators independently judged each pair with a binary label indicating whether the paraphrase instructs a task semantically identical to the original.
Each pair was accompanied by the environment's object reference, and annotators did not consult one another.
Annotators were provided solely with the binary equivalence criterion and were not informed which label would favor the benchmark: the instructions state no expected agreement rate and give no indication that either outcome supports or undermines the dataset.
Annotation was completed before any agent evaluation was conducted; hence, no annotator had seen model results at judgment time.
All annotators are researchers who routinely read and write English-language publications.
The full annotation prompt is shown in Figure~\ref{fig:human-eval-instruction}.

\paragraph{Experimental Results.}
Under majority vote, 324 of 328 pairs (98.8\%) were judged meaning-preserving, and 289 pairs (88.1\%) received unanimous approval.
We adjudicated all 39 split-vote items against the environment object reference: 30 were judged equivalent, 8 referentially ambiguous, and 1 a borderline part-for-whole substitution (\textit{fork} $\rightarrow$ \textit{tine}), which we conservatively counted as a failure.
Combined with the unanimously approved pairs, 319 of 328 (97.3\%) preserve the original task semantics.
All residual errors arise from lexical substitution, such as \textit{cloth} $\rightarrow$ \textit{wipe}, \textit{tomato} $\rightarrow$ \textit{love apple}, \textit{pot} $\rightarrow$ \textit{saucepan}, \textit{dishsponge} $\rightarrow$ \textit{scouring pad}, and \textit{egg} $\rightarrow$ \textit{cackleberry}.
We observed no cases in which the action or goal state itself was altered, indicating that the pragmatic and structural transformations in our generation pipeline preserve task semantics reliably; the residual risk is confined to the lexical axis.

\paragraph{Retention of flagged items.}
The nine flagged samples were retained in the benchmark by design. Terms
such as \textit{love apple} and \textit{cackleberry} are attested
synonyms that refer to the correct object; since they are archaic and
rarely used in contemporary English, they sit at the extreme end of the
lexical axis. We retain them to probe whether instruction understanding
holds even at the low-frequency tail of valid synonymy rather than only
for common alternatives.

\paragraph{Ethics.}
Annotators were unpaid volunteers from our research group who agreed to have their judgments used and reported in this work; given this affiliation, they were blind to which outcome would favor the benchmark (see Annotation setup).
The judged text was public and non-sensitive, and no personally identifiable information was collected.
We judged this not to constitute human subjects research under our institution's guidelines and did not seek IRB review.



\input{appenfigure/codebase}

\subsection{Experimental Results}
\label{appen:ALFWorld-Para:ExperimentalResults}
\subsubsection{Setup}
\label{appen:ALFWorld-Para:Setup}
\paragraph{Codebase-agnostic design.}
Both the benchmark and the method are designed to be independent of any particular agent codebase.
ALFWorld-Para is defined at the level of the environment: each episode is identified by its original ALFWorld task identifier, and the paraphrased instruction replaces the original one upon environment reset.
Because the game, the admissible actions, and the success condition remain untouched, any agent that runs on ALFWorld can be evaluated on ALFWorld-Para without modification.
BINDIntent likewise operates solely on the observations and the instruction, outside the policy and its prompting interface, and is applied unchanged in every codebase.
Neither component therefore depends on how a codebase formats prompts, parses actions, or trains its policy.

\paragraph{Codebases and weights.}
Table~\ref{tab:code_setup} lists the codebase and weights employed for each model.
Each model is evaluated within the codebase in which it was trained, as prompt formats, action parsers, and success checks vary across codebases.
Evaluation inputs are nonetheless held identical across codebases: every run draws on the same task identifiers and the same paraphrased instructions, ensuring that a policy receives exactly the same text irrespective of its codebase.
All results, including those on the original ALFWorld instructions, are obtained by running each model ourselves under this setup.
For a few models, these results deviate slightly from the originally reported figures owing to differences in software versions; since every drop and every method comparison is computed between our own runs of the same model within the same codebase, such deviations leave the comparisons unaffected.
We do not compare absolute SRs across codebases.
Results are averaged over three random seeds, amounting to $3 \times 6{,}554$ episodes per configuration on ALFWorld-Para and $3 \times 2{,}378$ on ScienceWorld-Para.

\paragraph{Pretrained models.}
For the larger pretrained models we use 8-bit (FP8 or INT8 weight-and-activation)
checkpoints, reflecting a practical deployment setting and our hardware budget;
8-bit quantization has been shown to preserve accuracy closely across a broad range
of tasks~\citep{kurtic2025give}. Qwen2.5-7B is used at its original precision. Since
every comparison we report is made between runs of the same model, the choice of
precision does not affect the measured drop or the effect of any binding method.

\paragraph{Fine-tuned models.}
Checkpoints that the original works did not release, namely the 1.5B models on ALFWorld and GiGPO-7B on ScienceWorld, were trained by us. At 1.5B, we train PPO, GRPO, and GiGPO with verl-agent~\cite{feng2026gigpo}, which reports reference results at this scale, and StraTA with its official codebase~\cite{xue2026strata}. Since verl-agent provides no evaluated ScienceWorld setup for GiGPO, we use BEACON~\cite{wang2026milestone}, a codebase built on verl-agent that includes a ScienceWorld implementation of GiGPO, solely to train and evaluate GiGPO-7B. Every model is trained with all settings identical to the default configuration of its source codebase, and the resulting checkpoints reproduce the originally reported performance where reference results exist.

\paragraph{Episode budget and decoding.}
On ALFWorld-Para, every codebase adopts the standard budget of 50 steps per episode, which likewise applies to the pretrained models run in the StraTA codebase.
On ScienceWorld-Para, the budget follows the setting under which each policy was trained: 20 steps for the StraTA codebase and 30 for BEACON.
Although the budgets differ, we prioritize matching the conditions under which each policy was trained, since a budget that deviates from training alters the episode distribution the policy anticipates; as all comparisons are made within a codebase, every compared run shares the same budget.
For the same reason, decoding follows the default settings of each codebase.
In conjunction with the identical instructions, this practice keeps both the inputs and the environment of every RL-finetuned policy as close as possible to those encountered during training, allowing any change in behavior to be attributed to the paraphrase rather than to the evaluation protocol.

\paragraph{Prompting of pretrained models.}
Pretrained models are run as ReAct agents~\citep{yao2023react} with the full interaction history included in the prompt.
These models frequently produce actions that do not conform to the format required by ALFWorld; in such cases, the model is re-queried, up to three times, to provide an action in the required format.
These re-queries neither consume an environment step nor are appended to the interaction history.
This fallback is applied uniformly to every pretrained-model run, including all binding methods and the baseline; consequently, all comparisons among pretrained models are conducted under the same condition.

\paragraph{Hardware and compute.}
RL training is run on NVIDIA A100 GPUs, two per run.
On ALFWorld-Para, models with up to 7B parameters are evaluated on four NVIDIA RTX 4090 GPUs and the larger pretrained models on a single NVIDIA H100 or A100; all ScienceWorld-Para evaluations run on a single A100.
Every model is served without tensor parallelism, with its weights on a single GPU, and all latency measurements are taken on an A100 (Appendix~\ref{appen:Experiment:Latency}).
Table~\ref{tab:compute} reports the compute used: training took 427 A100 hours, and a single evaluation pass took 248.8 RTX 4090 hours and 100.2 A100 hours on ALFWorld-Para and 12.3 A100 hours on ScienceWorld-Para.
\input{tables/gpu_spec}
\paragraph{Consistency across codebases.}
Because the three codebases differ in prompting, action parsing, training
algorithm, and decoding, they provide independent tests of our findings. The
degradation under object paraphrase and the gain of BINDIntent over the baseline
appear in every model, across all three codebases and both environments. The
findings and the effectiveness of BINDIntent are therefore not artifacts of a
particular implementation, but properties of how RL-finetuned agents ground
paraphrased instructions.

\paragraph{Artifacts and licenses.}
ALFWorld (MIT) and ScienceWorld (Apache 2.0) serve as evaluation environments; ConceptNet Numberbatch (CC BY-SA 4.0), \texttt{all-mpnet-base-v2} (Apache 2.0), and spaCy \texttt{en\_core\_web\_sm} (MIT) are used for slot extraction and binding. Agents are trained and evaluated with verl-agent, StraTA, and BEACON (all Apache 2.0). Model weights are used under their respective licenses: Qwen2.5, Qwen3, and Mistral-Small-3 (Apache 2.0), Phi-4 (MIT), and Llama 3.1 (Llama 3.1 Community License); quantized checkpoints follow the license of their base model. Paraphrases were generated with Gemini 3 Flash under the Gemini API Terms of Service, under which Google claims no ownership of generated content. All artifacts are used solely for research evaluation, consistent with their intended use. Our benchmarks, code, and checkpoints will be released under Apache 2.0, which is compatible with the licenses of the environments, codebases, and models they build on.

\paragraph{Release.}
Upon acceptance, we will release ALFWorld-Para, ScienceWorld-Para, the evaluation harness, BINDIntent, the training scripts, and all trained checkpoints in a single repository. Each benchmark is distributed as paraphrased instructions keyed by the original task identifiers, together with their paraphrase-type labels and gold slot annotations; it runs on an unmodified installation of ALFWorld or ScienceWorld and redistributes none of their game files. The generation and verification prompts (Appendix~\ref{app:para_generation}) are included for reproduction, the license notices of the source environments and codebases are retained, and Numberbatch vectors are obtained from their original source rather than redistributed.


\subsubsection{Extended Results}
Figure~\ref{appenfigure:heatmap_no_method} reports Baseline SRs on ALFWorld-Para without any mitigation applied.
Since the object is left as written in the None column, reading down a column isolates the action axis, whereas reading across a row isolates the object axis.
The three columns tend to decline in the order None, Addition, and Same-Polarity, and the drop into Same-Polarity is markedly larger than the gap between the first two.
Row by row, the Same-Polarity cell is almost invariably the lowest of the three, the sole exception being Base-1.5B, which remains near the floor across the entire grid.
As reported in \S~\ref{analysis:analysis1}, the object axis dominates, and this pattern persists within each action type.


%% file: appenfigure/Appen_alfworld_template.tex
\begin{table}[t]
\centering\scriptsize
\renewcommand{\arraystretch}{1.0}
\begin{tabular*}{\columnwidth}{@{\extracolsep{\fill}}p{2.2cm}p{5.2cm}@{}}
\toprule
\textbf{Task Type} & \textbf{Goal Instruction Templates} \\
\midrule
Pick \& Place      & \texttt{put a \{item\} in \{recep\}} \newline \texttt{put some \{item\} on \{recep\}} \\
Examine in Light   & \texttt{look at \{item\} under the \{lamp\}} \newline \texttt{examine the \{item\} with the \{lamp\}} \\
Clean \& Place     & \texttt{put a clean \{item\} in \{recep\}} \newline \texttt{clean some \{item\} and put it in \{recep\}} \\
Heat \& Place      & \texttt{put a hot \{item\} in \{recep\}} \newline \texttt{heat some \{item\} and put it in \{recep\}} \\
Cool \& Place      & \texttt{put a cool \{item\} in \{recep\}} \newline \texttt{cool some \{item\} and put it in \{recep\}} \\
Pick Two \& Place  & \texttt{put two \{item\} in \{recep\}} \newline \texttt{find two \{item\} and put them in \{recep\}} \\
\bottomrule
\end{tabular*}
\caption{The 6 task types in ALFWorld~\cite{shridhar2020alfworld} and their goal instruction templates. 
Each task type has 2 template variants.}
\label{tab:alfworld_templates}
\end{table}

%% file: appenfigure/Appen_alfworld-para_stat.tex
\begin{table}[h]
\centering\footnotesize
\renewcommand{\arraystretch}{1.05}
\begin{tabular*}{\columnwidth}{@{\extracolsep{\fill}}lllr@{}}
\toprule
\textbf{Axis} & \textbf{Category} & \textbf{Type} & \textbf{Count} \\
\midrule
\multirow{2}{*}{Object} & \multirow{2}{*}{Lexical} & Same-Polarity & 226 \\
 & & Addition & 226 \\
\midrule
\multirow{9}{*}{Action} & \multirow{2}{*}{Lexical} & Same-Polarity & 226 \\
 & & Addition & 226 \\
\cmidrule(lr){2-4}
 & \multirow{2}{*}{Structural} & Coordination & 226 \\
 & & Subordination & 226 \\
\cmidrule(lr){2-4}
 & \multirow{5}{*}{Pragmatic} & Need & 226 \\
 & & Embedded & 226 \\
 & & Permission & 226 \\
 & & Question & 226 \\
 & & Hint & 226 \\
\midrule
\multicolumn{3}{@{}l}{Compositional ($2 \times 9$ combinations)} & $226 \times 18$ \\
\midrule
\multicolumn{3}{@{}l}{\textbf{Total}} & \textbf{6,554} \\
\bottomrule
\end{tabular*}
\caption{ALFWorld-Para dataset breakdown by axis, category, and type.
Each of the 226 base instructions is perturbed into 11 single-perturbation
types (2 object + 9 action) and 18 compositional ones (every object $\times$
action pair), yielding $226 \times 29 = 6{,}554$ instructions.}
\label{tab:para_statistics}
\end{table}

%% file: appenfigure/codebase.tex
\newcommand{\hf}[1]{\href{https://huggingface.co/#1}{\texttt{#1}}}
\newcommand{\gh}[1]{\href{https://github.com/#1}{\texttt{#1}}}
\newcommand{\cbStraTA}{\gh{xxyQwQ/StraTA}}
\newcommand{\cbVerl}{\gh{langfengQ/verl-agent}}
\newcommand{\cbBeacon}{\gh{ZJU-REAL/BEACON}}
\newcommand{\oursw}{Ours (to be released after acceptance)}

\begin{table*}[t]
  \centering
  \scriptsize
  \setlength{\tabcolsep}{2pt}
  \renewcommand{\arraystretch}{1.05}
  \begin{tabular*}{\textwidth}{@{\extracolsep{\fill}}llll@{}}
    \toprule
    Model & Codebase & Weights & Fine-tuning code \\
    \midrule
    \multicolumn{4}{l}{\textbf{ALFWorld-Para}} \\
    \rowcolor{groupbg}
    \multicolumn{4}{l}{\textit{Pretrained (Prompting method: ReAct)}} \\
    Mistral-S-24B & \cbStraTA & \hf{RedHatAI/Mistral-Small-24B-Instruct-2501-FP8-dynamic} & -- \\
    Qwen3-14B     & \cbStraTA & \hf{Qwen/Qwen3-14B-FP8} & -- \\
    Phi-4-14B     & \cbStraTA & \hf{RedHatAI/phi-4-quantized.w8a8} & -- \\
    Llama3.1-8B   & \cbStraTA & \hf{RedHatAI/Meta-Llama-3.1-8B-Instruct-FP8} & -- \\
    Qwen2.5-7B    & \cbStraTA & \hf{Qwen/Qwen2.5-7B-Instruct} & -- \\
    \rowcolor{groupbg}
    \multicolumn{4}{l}{\textit{RL-finetuned (Base model: Qwen2.5)}} \\
    Base-7B       & \cbStraTA & \hf{Qwen/Qwen2.5-7B-Instruct} & -- \\
    PPO-7B        & \cbStraTA & \hf{xxyQwQ/train-ppo-alfworld-text-qwen2.5-7b} & -- \\
    GRPO-7B       & \cbStraTA & \hf{xxyQwQ/train-grpo-alfworld-text-qwen2.5-7b} & -- \\
    GiGPO-7B      & \cbVerl   & \hf{langfeng01/GiGPO-Qwen2.5-7B-Instruct-ALFWorld} & -- \\
    StraTA-7B     & \cbStraTA & \hf{xxyQwQ/train-strata-alfworld-text-qwen2.5-7b} & -- \\
    Base-1.5B     & \cbStraTA & \hf{Qwen/Qwen2.5-1.5B-Instruct} & -- \\
    PPO-1.5B      & \cbVerl   & \oursw & \cbVerl \\
    GRPO-1.5B     & \cbVerl   & \oursw & \cbVerl \\
    GiGPO-1.5B    & \cbVerl   & \oursw & \cbVerl \\
    StraTA-1.5B   & \cbStraTA & \oursw & \cbStraTA \\
    \midrule
    \multicolumn{4}{l}{\textbf{ScienceWorld-Para}} \\
    \rowcolor{groupbg}
    \multicolumn{4}{l}{\textit{RL-finetuned (Base model: Qwen2.5)}} \\
    PPO-7B        & \cbStraTA & \hf{xxyQwQ/train-ppo-sciworld-text-qwen2.5-7b} & -- \\
    GRPO-7B       & \cbStraTA & \hf{xxyQwQ/train-grpo-sciworld-text-qwen2.5-7b} & -- \\
    GiGPO-7B      & \cbBeacon & \oursw & \cbBeacon$^{\dagger}$ \\
    StraTA-7B     & \cbStraTA & \hf{xxyQwQ/train-strata-sciworld-text-qwen2.5-7b} & -- \\
    \bottomrule
  \end{tabular*}
  \caption{Codebase and weights per model. Each model is evaluated in the codebase it was trained with; our checkpoints follow that codebase's default training configuration. Codebases: StraTA, verl-agent, and BEACON~\citep{wang2026milestone}. $^{\dagger}$The original GiGPO release does not include a ScienceWorld checkpoint.}
  \label{tab:code_setup}
\end{table*}

%% file: tables/gpu_spec.tex
\begin{table}[t]
\centering
\small
\begin{tabular}{lrr}
\toprule
Model & RTX 4090 h & A100 h \\
\midrule
\multicolumn{3}{l}{\textit{RL training}} \\
PPO-1.5B (ALFWorld) & -- & 94 \\
GRPO-1.5B (ALFWorld) & -- & 50 \\
GiGPO-1.5B (ALFWorld) & -- & 53 \\
StraTA-1.5B (ALFWorld) & -- & 70 \\
GiGPO-7B (ScienceWorld) & -- & 160 \\
\midrule
\multicolumn{3}{l}{\textit{Evaluation on ALFWorld-Para (one pass)}} \\
Mistral-S-24B & -- & 22.7 \\
Qwen3-14B & -- & 37.1 \\
Phi-4-14B & -- & 26.0 \\
Llama3.1-8B & -- & 14.4 \\
Qwen2.5-7B & 73.1 & -- \\
Qwen2.5-1.5B & 30.0 & -- \\
Base-7B & 22.7 & -- \\
PPO-7B & 17.3 & -- \\
GRPO-7B & 20.8 & -- \\
GiGPO-7B & 13.3 & -- \\
StraTA-7B & 14.7 & -- \\
Base-1.5B & 13.8 & -- \\
PPO-1.5B & 11.7 & -- \\
GRPO-1.5B & 8.2 & -- \\
GiGPO-1.5B & 10.2 & -- \\
StraTA-1.5B & 13.0 & -- \\
\midrule
\multicolumn{3}{l}{\textit{Evaluation on ScienceWorld-Para (one pass)}} \\
Qwen2.5-7B & -- & 3.0 \\
Base-7B & -- & 2.9 \\
PPO-7B & -- & 2.3 \\
GRPO-7B & -- & 1.7 \\
GiGPO-7B & -- & 1.7 \\
StraTA-7B & -- & 0.7 \\
\bottomrule
\end{tabular}
\caption{Compute in GPU hours. Models with up to 7B parameters on ALFWorld-Para are reported in RTX 4090 hours and all other rows in A100 hours, with H100 times converted at 2 A100 hours per H100 hour. Evaluation rows report a single pass (one seed) over each benchmark; some times are estimated from measured throughput.}
\label{tab:compute}
\end{table}

%% file: sections/94_appen_analysis.tex
\section{Analysis}
\label{appen:Analysis}

\subsection{Analysis 1: Object Synonym Substitution Is the Dominant Failure Factor}
\label{appen:analysis:analysis1}


Figure~\ref{appenfigure:extended_object_drop} unpacks
Figure~\ref{fig:analysis_object_drop} in two ways. First, Figure~\ref{fig:analysis_object_drop}
pools seven of the nine action types into a single Others group and shows only
Question directive and Hint separately; here all nine are shown individually,
grouped into lexical (Addition, Same-Polarity), structural (Coordination,
Subordination), and pragmatic types (Embedded imperative, Need statement,
Permission directive, Question directive, Hint). Second,
Figure~\ref{fig:analysis_object_drop} reports action$\times$object results only
for SP; here each action type is crossed with both object paraphrases
(second row of each group, Addition left and SP right). The encoding
is unchanged: each marker is one model, marker size indicates parameter count,
and the horizontal bar and diamond denote the median and mean across models.
The expanded view confirms the pattern. Action types differ only modestly, with
Question directive the weakest. Object SP, in contrast,
lowers success far more than any action paraphrase, and the drop is nearly
uniform across action types. Crossing with object Addition stays
close to the action-only level. This holds for both pretrained and RL-finetuned
models, indicating that object synonym substitution is the dominant failure
factor rather than an artifact of any particular action type.

\subsection{Analysis 2: sLLMs Still Know What the Paraphrased Object Refers to}
\label{appen:analysis:analysis2}

\paragraph{Probe set.}
We use the term \emph{object} for both items and receptacles, the two kinds of
entities an ALFWorld instruction can refer to. The probe consists of every
paraphrased object name that appears in ALFWorld-Para, taken verbatim from the
benchmark instructions: 111 names, each paired with the entity type it replaces. Because no item is added, removed, or rewritten, the probe covers exactly the paraphrases the agents encounter during evaluation.

\paragraph{Fully observable lexical space.}
An agent perceives only the entity names it has observed thus far; its failure to bind a paraphrase may therefore stem from incomplete exploration rather than from a lack of knowledge.
The probe eliminates this confound by presenting the complete entity vocabulary of the environment in every prompt.
It thereby measures paraphrase knowledge with the environment's object lexical space rendered fully observable: the model need only recognize which listed entity type the paraphrase denotes.

\paragraph{Prompt templates.}
Because estimates of a language model's knowledge are sensitive to how the query
is worded~\citep{jiang2020can,mizrahi2024stateofwhatart}, each name is inserted into five
templates that pose the same question in different forms: list choice, direct
question, cloze completion, synonym question, and an agent-framed instruction
(Figure~\ref{appenfigure:paraknow_prompt}). 
All templates list the 82 entity types in their environment form and require the answer to be exactly one name from the list, which eliminates free-form matching.

\paragraph{Decoding and scoring.}
We use greedy decoding for probing, with each model returning one deterministic answer per prompt.
An answer is correct only if it is the gold entity type. The Paraphrase Knowledge
score in Figure~\ref{fig:analysis_paraknow} is the accuracy over the 111 names,
reported as the mean and standard deviation across the five templates.

\paragraph{Failure cases.}
To distinguish knowledge a model lacks from sensitivity to prompt wording, we
regard a name as a persistent failure only if the model answers incorrectly under
all five templates. Such failures cover only a small fraction of the 111 names for
models of 7B parameters or more. Every failed name had been judged semantically
equivalent to its target in our human evaluation
(Figure~\ref{fig:human-eval-instruction}); the failures therefore reflect gaps in
model knowledge rather than errors in the benchmark.

Aside from a few names that admit more than one plausible reading, the persistent
failures of the 7B models are rare idiomatic or compound expressions with an
unambiguous referent, such as \textit{range eye} (\texttt{stoveburner}),
\textit{bath tissue spindle} (\texttt{toiletpaperhanger}), and \textit{bathroom
roll} (\texttt{toiletpaper}). Common paraphrases of household objects are thus
largely known to these models. RL fine-tuning does not introduce new kinds of
failure: the names missed by the base model are largely the same names missed by
its RL-finetuned variants, which add at most a few errors of their own. These
failures are too few, and too concentrated on rare expressions, to account for the
performance drop under paraphrase.

\subsection{Analysis 3: Agents Encounter the Target Object but Do Not Take It}
\label{appen:analysis:analysis3}

\paragraph{Definition.}
The taken rate measures whether an agent takes the target once it is in view.
The target is the item slot of the goal instruction, and any instance of its type (e.g., \texttt{book 1} or \texttt{book 2}) counts; receptacles are excluded, since actions on them (e.g., \texttt{go to}, \texttt{open}) can also be part of task execution and are hard to attribute to binding.
A step is \emph{observed} if the target appears in the observation and has not yet been taken, including the step of the take itself; for two-instance tasks, this restarts after the first take.
Let $N^{\text{obs}}_e$ be the number of observed steps in episode $e$, and $N^{\text{take}}_e$ the number of those at which the agent takes the target.
For each object-axis paraphrase type (None, Addition, or Same-Polarity), the taken rate pools these counts over all episodes of that type, $\mathcal{E}_{\text{type}}$:
\begin{equation}
\textsc{TakenRate}(\text{type}) = \frac{\sum_{e \in \mathcal{E}_{\text{type}}} N^{\text{take}}_e}{\sum_{e \in \mathcal{E}_{\text{type}}} N^{\text{obs}}_e}.
\end{equation}

\paragraph{Step-level pooling.}
We pool at the step level because each step corresponds to a single action decision.
The measure thus decomposes behavior into the individual choices of whether to take an observed target, rather than collapsing each episode into a single outcome.

%% file: sections/95_appen_method.tex
\input{sections/95-1_appen_algorithm}

\input{sections/95-2_algorithm_subroutine}

\section{Proposed Method}
\label{appen:Method}
Algorithm~\ref{alg:method} gives the per-step loop of BINDIntent and Algorithm~\ref{alg:subroutines} the three subroutines it calls. \S\ref{appen:BindIntent:ProblemSetup} defines the helper functions through which the loop reads the environment and rewrites the instruction, and \S\ref{appen:Method:Candidate_Scoring}--\S\ref{appen:Method-Persistence-Driven-Binding} describe one subroutine each; within them, \S\ref{app:weibull-derivation} derives the shape posterior and the closed form of the posterior predictive probability $p_t$.

\subsection{Problem Setup and Notation}
\label{appen:BindIntent:ProblemSetup}

The environment is read through $\textsc{Env}$, $\textsc{Objects}$, $\textsc{Units}$, and $\textsc{Explored}$, and the instruction through $\textsc{Match}$ and $\textsc{Replace}$. $\textsc{Agent}$ denotes the frozen policy under evaluation, whose weights and decoding settings are identical across all compared methods. $\textsc{Bound}(s^{(i)}) \in \{0,1\}$ is per-slot state: it starts at $0$, is set by the Bernoulli draw of $\textsc{Bind}$, and is never cleared.

\paragraph{\textsc{Env}.}
$\textsc{Env}(a_{t-1})$ executes the agent's previous action and returns the resulting text observation $o_t$. BINDIntent neither selects nor modifies actions; it reads $o_t$ only to update $V_t$ and $R_t$.

\paragraph{\textsc{Objects} \& \textsc{Units}.}
$\textsc{Objects}(o_t)$ returns the entity types mentioned in $o_t$ with instance identity removed. Two drawers therefore contribute a single entry, \texttt{drawer}, and $|V_t|$ counts distinct labels rather than physical objects. $\textsc{Units}(o_t)$ returns the places whose contents can be inspected, as individual instances, with $R_t$ counting those two drawers twice. The asymmetry is deliberate: an instruction names a label, a unit's own label included, whereas the agent opens one particular unit. Both sets are taken from what the environment exposes rather than from a list fixed in advance.

\paragraph{\textsc{Explored}.}
A unit is \textsc{explored} once its contents have been revealed to the agent, including the case where it turns out to be empty; reaching a unit without seeing inside it does not count. Exploration is never undone, and $k_t$ thus counts the known places whose contents remain unseen, that is, the chances left for a not-yet-observed object to appear. 

\paragraph{\textsc{Replace}.}
$\textsc{Replace}$ substitutes, for each bound slot, its current best match in place of the original mention, within the instruction and on whole words; unbound slots keep their original wording. Because $v^{\star}_t$ is re-evaluated at every step, a bound slot continues to follow the best match as new objects are observed. Binding decides \emph{whether} a mention is rewritten; $\textsc{Match}$ decides \emph{into what}.

\subsection{Candidate Distance}
\label{appen:Method:Candidate_Scoring}

\paragraph{Embedding lookup.}
Multi-word expressions are embedded via their Numberbatch phrase entry when one exists (e.g., \textit{end\_table}) and otherwise as the mean of their word vectors, a simple yet strong composition for short phrases \citep{mitchell2010composition,wieting2016universal}. Concatenated environment labels (e.g., \textit{sidetable}) are first split into their constituent words.

$\textsc{CandidateDistance}$ selects $v^{\star}_t$ and collects into $D_t$ the distances of the candidate referents, the labels lying within the lexical allowance $\tau_t$ (Eq.~\ref{eq:candidates}). Labels observed together in a scene tend to share a common semantic component (e.g., a household context), which inflates their similarity to any query. $\textsc{Match}$ removes this component before ranking. With $\mu_t$ the mean embedding of the labels in $V_t$ and $\tilde{E}(a) = (E(a) - \mu_t)/\lVert E(a) - \mu_t \rVert$, it returns
\begin{equation}
  v^{\star}_t = \operatorname*{arg\,max}_{v \in V_t}\; \tilde{E}(s)^{\top}\tilde{E}(v).
\end{equation}
Centering is used only for this selection; the distances collected in $D_t$ are the uncentered $\delta$ of Eq.~\ref{eq:distance}, which does not depend on $V_t$.

\subsection{Best-Match Persistence under Partial Observability}
\label{appen:Method:Best-Match}

$\textsc{Persistence}$ is called only while the slot is unbound. It turns $D_t$ into $P_t$, the probability that $v^{\star}_t$ remains the best match once the $k_t$ unexplored units have been opened, and it separates three cases. With no candidate referent inside the allowance there is nothing to bind to, and the procedure returns $0$. With fewer than $c$ of them the distance distribution cannot be fit, and $P_t$ is replaced by a count-based estimate: if the referent is equally likely to be any of the $|V_t|$ observed labels or to lie behind one of the $k_t$ unexplored units, it has already been observed with probability $|V_t|/(|V_t| + k_t)$. Otherwise the Weibull of Eq.~\ref{eq:weibull} is fit to $D_t$, giving the posterior predictive probability $p_t$ that an unseen candidate lies farther than $d^{\star}_t$ (Eq.~\ref{eq:predictive}); treating each unexplored unit as one independent chance for such a candidate yields $P_t = p_t^{\,k_t}$. The derivation is in \S\ref{app:weibull-derivation}.

\input{sections/95-0_posterior_and_target_probability}

\subsection{Persistence-Driven Binding}
\label{appen:Method-Persistence-Driven-Binding}

$\textsc{Bind}$ turns $P_t$ into a decision. Since a binding cannot be undone, the probability already committed to it must be preserved: the cumulative binding probability is the running max $P^{\max}_t$ of Eq.~\ref{eq:target} rather than the current $P_t$, and a step on which persistence falls adds nothing. The hazard $h_t$ is the increase $P^{\max}_t - P^{\max}_{t-1}$ expressed as a fraction of the probability still unbound, which is what makes the per-step Bernoulli draws telescope to $\Pr[\textsc{Bound}(s) \text{ by } t] = P^{\max}_t$. Once $\textsc{Bound}(s^{(i)})$ is set, Algorithm~\ref{alg:method} stops calling either subroutine for that slot, and $\textsc{Replace}$ substitutes its best match into the instruction from that step onward.

%% file: sections/95-1_appen_algorithm.tex
\makeatletter
\algnewcommand{\LineComment}[1]{\Statex \hskip\ALG@thistlm \(\triangleright\) #1}
\renewcommand{\dblfloatpagefraction}{0.1}
\makeatother

\begin{algorithm*}[p]
\caption{BINDIntent.}
\label{alg:method}
\begin{algorithmic}[1]
\Require instruction $x$ with slots $s^{(1)}, \dots, s^{(I)}$, horizon $T$, step size $\Delta\tau$, fallback threshold $c$
\State $V_0 \gets \varnothing$,\quad $R_0 \gets \varnothing$,\quad $\tau_0 \gets 0$,\quad $a_0 \gets \texttt{reset}$ \Comment{observed object labels $V$, observed units $R$}
\State $\big\{\, \textsc{Bound}(s^{(i)}) \gets 0,\;\; P^{\max(i)}_0 \gets 0 \,\big\}_{i=1}^{I}$
\For{$t = 1, \dots, T$}
    \State $o_t \gets \textsc{Env}(a_{t-1})$
    \State $V_t \gets V_{t-1} \cup \textsc{Objects}(o_t)$,\quad $R_t \gets R_{t-1} \cup \textsc{Units}(o_t)$
    \State $k_t \gets \big|R_t \setminus \textsc{Explored}(R_t)\big|$ \Comment{unexplored units, the chances left to observe a new object}
    \State $\tau_t \gets t\,\Delta\tau$ \Comment{lexical allowance widens}
    \For{$i = 1, \dots, I$}
        \State $v^{\star(i)}_t,\, D^{(i)}_t \gets \Call{CandidateDistance}{s^{(i)}, V_t, \tau_t}$
        \If{$\lnot\,\textsc{Bound}(s^{(i)})$}
            \State $P^{(i)}_t \gets \Call{Persistence}{s^{(i)}, v^{\star(i)}_t, D^{(i)}_t, V_t, k_t, c}$
            \State $P^{\max(i)}_t \gets \Call{Bind}{s^{(i)}, P^{(i)}_t, P^{\max(i)}_{t-1}}$
        \EndIf
    \EndFor
    \State $x_t \gets \textsc{Replace}\big(x,\; \{\, s^{(i)} \mapsto v^{\star(i)}_t \;:\; \textsc{Bound}(s^{(i)}) \,\}\big)$
    \State $a_t \sim \textsc{Agent}(\cdot \mid x_t)$
\EndFor
\end{algorithmic}
\end{algorithm*}

%% file: sections/95-2_algorithm_subroutine.tex
\begin{algorithm*}[p]
\caption{Subroutines of BINDIntent (described in \S\ref{appen:Method:Candidate_Scoring}--\S\ref{appen:Method-Persistence-Driven-Binding}).}

\label{alg:subroutines}
\begin{algorithmic}[1]
\Procedure{CandidateDistance}{$s$, $V_t$, $\tau_t$}
    \State $v^{\star}_t \gets \textsc{Match}(s, V_t)$ \Comment{best match, tracked even after binding}
    \State $D_t \gets \big(\, \delta(s, v) \;:\; v \in V_t,\; \delta(s, v) \le \tau_t \,\big)$ \Comment{distances of candidate referents}
    \State \Return $v^{\star}_t,\, D_t$
\EndProcedure
\Statex
\Procedure{Persistence}{$s$, $v^{\star}_t$, $D_t$, $V_t$, $k_t$, $c$}
    \State $n_t \gets |D_t|$
    \If{$n_t = 0$}
        \State \Return $0$ \Comment{no candidate referent, so no binding occurs}
    \ElsIf{$n_t < c$}
        \State \Return $\big(|V_t| / (|V_t| + k_t)\big)$ \Comment{too few candidate referents; exchangeability-based fallback}
    \EndIf
    \State $d^{\star}_t \gets \delta(s,\, v^{\star}_t)$ \Comment{best-match distance, raw rather than centered}
    \State $\pi\big(\beta \mid D_t\big) \propto \beta^{\,n_t-2}\, S_t(\beta)^{-n_t} \prod_{d \in D_t} d^{\,\beta-1}$
    \State $p_t \gets \displaystyle\int \pi\big(\beta \mid D_t\big) \left(\frac{S_t(\beta)}{S_t(\beta) + (d^{\star}_t)^{\beta}}\right)^{n_t} d\beta$ \Comment{an unseen candidate is farther than the best match}
    \State \Return $(p_t)^{k_t}$ \Comment{persistence over $k_t$ unexplored units}
\EndProcedure
\Statex
\Procedure{Bind}{$s$, $P_t$, $P^{\max}_{t-1}$}
    \LineComment{$h_t$ is chosen so that $\Pr[\textsc{Bound}(s) \text{ by } t] = P^{\max}_t$}
    \State $P^{\max}_t \gets \max\!\left(P^{\max}_{t-1},\, P_t\right)$ \Comment{running max of persistence}
    \State $h_t \gets \dfrac{P^{\max}_t - P^{\max}_{t-1}}{1 - P^{\max}_{t-1}}$ \Comment{this step's increase, as a fraction of the probability still unbound}
    \State $\textsc{Bound}(s) \sim \operatorname{Bernoulli}(h_t)$
    \State \Return $P^{\max}_t$
\EndProcedure
\end{algorithmic}
\end{algorithm*}

%% file: sections/95-0_posterior_and_target_probability.tex
\subsubsection{Derivation of the Weibull Posterior and Posterior Predictive Probability}
\label{app:weibull-derivation}

We expand the computation behind Eq.~(3)--(4) of the main text.
Dropping the time and slot indices, write
$D=\{d_1,\ldots,d_n\}$ for the observed candidate referent distances,
$d^\star$ for the current best-match distance, and
$S(\beta):=\sum_{i=1}^{n} d_i^\beta$.

\paragraph{Step 1: Likelihood.}
The density corresponding to the Weibull survival function
$\Pr(d>z\mid\beta,\eta)
=\exp[-(z/\eta)^\beta]$ of Eq.~(3) is
\begin{align}
f(d\mid\beta,\eta)
&=
\frac{\beta}{\eta}
\left(\frac{d}{\eta}\right)^{\beta-1}
\exp\!\left[-(d/\eta)^\beta\right]
\notag\\
&=
\beta d^{\beta-1}\eta^{-\beta}
\exp\!\left[-d^\beta\eta^{-\beta}\right].
\label{eq:app-density}
\end{align}

If the elements of $D$ are conditionally i.i.d.\ given $(\beta,\eta)$, then
\begin{align}
p(D\mid\beta,\eta)
&=
\prod_{i=1}^{n}
f(d_i\mid\beta,\eta)
\notag\\
&=
\prod_{i=1}^{n}
\left[
\beta d_i^{\beta-1}
\eta^{-\beta}
e^{-d_i^\beta\eta^{-\beta}}
\right]
\notag\\
&=
\beta^n
\left(
\prod_{i=1}^{n} d_i^{\beta-1}
\right)
\eta^{-n\beta}
\notag\\
&\quad\times
\exp\!\left[
-\eta^{-\beta}
\sum_{i=1}^{n}d_i^\beta
\right]
\notag\\
&=
\beta^n
\left(
\prod_{i=1}^{n} d_i^{\beta-1}
\right)
\eta^{-n\beta}
e^{-S(\beta)\eta^{-\beta}}.
\label{eq:app-like}
\end{align}

\paragraph{Step 2: Joint posterior.}
With the prior
$\pi(\beta,\eta)\propto(\beta\eta)^{-1}$,
Bayes' rule gives
\begin{align}
\pi(\beta,\eta\mid D)
&\propto
p(D\mid\beta,\eta)
\pi(\beta,\eta)
\notag\\
&\propto
\beta^n
\left(
\prod_{i=1}^{n}d_i^{\beta-1}
\right)
\eta^{-n\beta}
\notag\\
&\quad\times
e^{-S(\beta)\eta^{-\beta}}
\frac{1}{\beta\eta}
\notag\\
&\propto
\beta^{n-1}
\left(
\prod_{i=1}^{n}d_i^{\beta-1}
\right)
\eta^{-n\beta-1}
\notag\\
&\quad\times
e^{-S(\beta)\eta^{-\beta}}.
\label{eq:app-jointpost}
\end{align}

Here $\propto$ indicates that factors not depending on
$(\beta,\eta)$ have been dropped.

\paragraph{Step 3: Marginalizing \texorpdfstring{$\eta$}{eta}.}
We integrate $\eta$ out,
\begin{equation}
\pi(\beta\mid D)
=
\int_0^\infty
\pi(\beta,\eta\mid D)\,d\eta.
\end{equation}

Pulling the factors independent of $\eta$ out of
Eq.~\eqref{eq:app-jointpost} gives
\begin{align}
\pi(\beta\mid D)
&\propto
\beta^{n-1}
\left(
\prod_{i=1}^{n}d_i^{\beta-1}
\right)
I(\beta),
\label{eq:app-beta-before}
\end{align}
where
\begin{equation}
I(\beta)
:=
\int_0^\infty
\eta^{-n\beta-1}
e^{-S(\beta)\eta^{-\beta}}
\,d\eta.
\label{eq:app-Ibeta}
\end{equation}

The substitution $u=S(\beta)\eta^{-\beta}$ gives
\begin{align}
\eta
&=
\left(
\frac{S(\beta)}{u}
\right)^{1/\beta},
\notag\\
d\eta
&=
-\frac{1}{\beta}
S(\beta)^{1/\beta}
u^{-1/\beta-1}\,du,
\label{eq:app-substitution}
\end{align}
\begin{align}
\eta^{-n\beta-1}
&=
\left(
\frac{u}{S(\beta)}
\right)^{n+1/\beta},
\end{align}

so that
\begin{align}
\eta^{-n\beta-1}d\eta
&=
-\frac{1}{\beta}
S(\beta)^{-n}
u^{n-1}\,du.
\label{eq:app-transform}
\end{align}

Since $\eta:0\rightarrow\infty$ corresponds to
$u:\infty\rightarrow0$, reversing the limits gives

\begin{align}
I(\beta)
&=
\frac{1}{\beta}
S(\beta)^{-n}
\int_0^\infty
u^{n-1}e^{-u}\,du
\notag\\
&=
\frac{\Gamma(n)}
{\beta S(\beta)^n}.
\label{eq:app-Iclosed}
\end{align}

Substituting into Eq.~\eqref{eq:app-beta-before},
\begin{align}
\pi(\beta\mid D)
&\propto
\beta^{n-1}
\left(
\prod_{i=1}^{n}d_i^{\beta-1}
\right)
\frac{\Gamma(n)}
{\beta S(\beta)^n}
\notag\\
&\propto
\beta^{n-2}
S(\beta)^{-n}
\prod_{i=1}^{n}d_i^{\beta-1}.
\label{eq:app-betapost}
\end{align}

Writing the unnormalized posterior kernel as
\begin{equation}
g(\beta;D)
:=
\beta^{n-2}
S(\beta)^{-n}
\prod_{i=1}^{n}d_i^{\beta-1},
\label{eq:app-gbeta}
\end{equation}
the normalized posterior is
\begin{equation}
\pi(\beta\mid D)
=
\frac{g(\beta;D)}
{\displaystyle
 \int_0^\infty g(b;D)\,db}.
\label{eq:app-normalized-beta}
\end{equation}

\paragraph{Step 4: Posterior predictive probability.}
This step derives Eq.~(4), the posterior predictive probability
\begin{equation}
p
:=
\Pr(d_{\mathrm{new}}>d^\star\mid D).
\end{equation}

Given $(\beta,\eta)$, the Weibull survival function gives
\begin{equation}
\Pr(d_{\mathrm{new}}>d^\star\mid\beta,\eta)
=
e^{-(d^\star)^\beta\eta^{-\beta}}.
\label{eq:app-survival-best}
\end{equation}

Averaging the parameter uncertainty under the posterior,
\begin{align}
p
&=
\int_0^\infty
\int_0^\infty
e^{-(d^\star)^\beta\eta^{-\beta}}
\notag\\
&\qquad\qquad\times
\pi(\beta,\eta\mid D)
\,d\eta\,d\beta.
\label{eq:app-predictive-double}
\end{align}

For fixed $\beta$, the conditional posterior of $\eta$ is
\begin{align}
\pi(\eta\mid\beta,D)
&\propto
\eta^{-n\beta-1}
e^{-S(\beta)\eta^{-\beta}}.
\end{align}

Normalizing with Eq.~\eqref{eq:app-Iclosed},
\begin{align}
\pi(\eta\mid\beta,D)
&=
\frac{\beta S(\beta)^n}{\Gamma(n)}
\eta^{-n\beta-1}
\notag\\
&\quad\times
e^{-S(\beta)\eta^{-\beta}}.
\label{eq:app-eta-cond}
\end{align}

Writing the predictive survival at fixed $\beta$ as
\begin{equation}
q(\beta)
:=
\Pr(d_{\mathrm{new}}>d^\star\mid\beta,D),
\end{equation}
we have
\begin{align}
q(\beta)
&=
\int_0^\infty
e^{-(d^\star)^\beta\eta^{-\beta}}
\pi(\eta\mid\beta,D)
\,d\eta
\notag\\
&=
\frac{\beta S(\beta)^n}{\Gamma(n)}
\int_0^\infty
\eta^{-n\beta-1}
\notag\\
&\quad\times
\exp\!\left[
-\left\{
S(\beta)+(d^\star)^\beta
\right\}\eta^{-\beta}
\right]
d\eta.
\label{eq:app-qbeta-integral}
\end{align}

As before, for any $A>0$ we have
\begin{equation}
\int_0^\infty
\eta^{-n\beta-1}
e^{-A\eta^{-\beta}}
\,d\eta
=
\frac{\Gamma(n)}
{\beta A^n},
\label{eq:app-integral-identity}
\end{equation}

so substituting $A=S(\beta)+(d^\star)^\beta$ gives
\begin{align}
q(\beta)
&=
\frac{\beta S(\beta)^n}{\Gamma(n)}
\frac{\Gamma(n)}
{\beta
 \left[
 S(\beta)+(d^\star)^\beta
 \right]^n}
\notag\\
&=
\left(
\frac{S(\beta)}
{S(\beta)+(d^\star)^\beta}
\right)^n.
\label{eq:app-qbeta}
\end{align}

Finally, averaging the uncertainty in $\beta$ under
$\pi(\beta\mid D)$,
\begin{align}
p
&=
\int_0^\infty
\pi(\beta\mid D)
\left(
\frac{S(\beta)}
{S(\beta)+(d^\star)^\beta}
\right)^n
d\beta.
\label{eq:app-final-p}
\end{align}

\paragraph{Step 5: Restoring the indices.}
With $D=D_t$, $n=n_t$, and $d^\star=d_t^\star$,
\begin{align}
\pi(\beta\mid D_t)
&\propto
\beta^{n_t-2}
S_t(\beta)^{-n_t}
\prod_{d\in D_t}
d^{\beta-1},
\label{eq:app-final-posterior}
\\
p_t
&=
\int_0^\infty
\pi(\beta\mid D_t)
\left(
\frac{S_t(\beta)}
{S_t(\beta)+(d_t^\star)^\beta}
\right)^{n_t}
d\beta.
\label{eq:app-final-predictive}
\end{align}

%% file: sections/96_appen_experiment.tex
\section{Experiments}
\label{appen:Experiment}

\subsection{Main Results}
\label{appen:Experiment:MainResults}

\subsubsection{Setup}
\label{appen:Experiment:Setup}
This section details the methods compared in \S\ref{sec:experiment}.


\paragraph{Shared pipeline.}
All methods share a single substitution layer and differ only in how they determine the referent of each slot.
Before the episode, the parser extracts the object slots from the instruction (Appendix~\ref{parser_stability}).
At every step, the layer accumulates the object labels that have appeared in the observations, determines a label for each slot, and replaces the slot in the instruction before passing it to the frozen agent.

Only candidates that have already been observed can be selected, and the replacement is confined to the instruction text, leaving the system prompt, few-shot examples, observations, and admissible actions unchanged.
Since every method runs at every step under this layer, their task performance and per-step latency are directly comparable.
 
\paragraph{Baseline.}
The paraphrased instruction is passed to the agent unchanged, leaving grounding entirely to the LLM.
 
\paragraph{Greedy.}
Greedy uses the same candidate distance as BINDIntent, including mean-centering over observed labels (\S\ref{sec:scoring}).
At every step, it replaces each slot with the current best match without estimating persistence; an observed label that exactly matches the slot is selected directly.
As new labels are observed, the replacement follows the updated best match.
Greedy therefore isolates the effect of the binding decision in BINDIntent, as both methods select the same best match at every step and differ only in when they commit to it.
 
\paragraph{Self-clarification.}
Self-clarification adapts query-based disambiguation~\citep{ren2023knowno,ramrakhya2025asktoact} to a setting without a user, delegating the binding decision to the agent LLM itself.
At every step, after the agent acts, it constructs a query containing the instruction, the object types observed up to that step, and the referring expressions (Figure~\ref{fig:selfclarify_prompt}).
The agent LLM answers this query in a separate generation call with the same weights, selecting one observed object or \texttt{None} for each expression.
The answer replaces the expressions in the instruction at the next step; an expression answered with \texttt{None} or an unparsable answer remains unchanged.
The query explicitly allows \texttt{None} when no observed object is the referent, and contains no examples or hints, ensuring that the method reflects the agent's own binding ability.
Its format mirrors the list-choice template of the paraphrase-knowledge probe (Appendix~\ref{appen:analysis:analysis2}), on which the same models identify most referents when all objects are listed.
The query receives the same observed candidates as BINDIntent but no exploration progress, since estimating whether an unseen candidate may still be closer is the component that BINDIntent introduces.
 
\paragraph{BINDIntent\textsubscript{SBERT}.}
BINDIntent\textsubscript{SBERT} is identical to BINDIntent except that it computes candidate distances with the sentence encoder \texttt{all-mpnet-base-v2}~\citep{reimers2019sentence} instead of Numberbatch.
Because the encoder is applied to each newly observed label, its latency grows with the number of labels revealed per step (Appendix~\ref{appen:Experiment:Latency}).

\paragraph{BINDIntent hyperparameters.}
BINDIntent uses a lexical allowance step of $\Delta\tau = 0.04$ and a fallback threshold of $c = 3$, the smallest sample size exceeding the two Weibull parameters; results vary little with $c$ (Table~\ref{tab:exp_appen_abl_nprime}).
The same values are used on ScienceWorld-Para without benchmark-specific tuning.
For pretrained models we use $\Delta\tau = 0.025$.
Since $\tau_t$ widens with the step count, the allowance is effective only when it widens in step with exploration.
Unlike RL-finetuned policies, pretrained models are not adapted to the environment and often spend steps on inadmissible or redundant actions that reveal no new objects. A smaller step keeps the allowance aligned with their slower exploration.
All other settings are shared across models.

\tcbset{
  sclbox/.style={
    colback=blue!5, colframe=blue!60!black, boxrule=0.5pt,
    left=6pt, right=6pt, top=4pt, bottom=4pt,
    fontupper=\scriptsize\ttfamily, before upper=\raggedright\frenchspacing,
    coltitle=black, colbacktitle=blue!20},
}
\providecommand{\phb}[1]{\textcolor{blue!70!black}{\texttt{\{#1\}}}}
\begin{figure}[t]
\centering
\begin{tcolorbox}[sclbox, width=\columnwidth, title={\footnotesize\textbf{Self-clarification query}}]
You are an agent in an interactive environment.\\[3pt]
Task instruction: \phb{instruction}\\
Current step: \phb{step}\\
Your current location: \phb{current\_location}\\
Your previous action: \phb{previous\_action}\\
Objects you have observed so far: \phb{observed\_objects}\\[3pt]
In the instruction, which of the observed objects do \phb{phrases} refer to?\\[3pt]
Answer with a single line containing \phb{n} comma-separated value(s), in the same order as the quoted phrase(s).\\
Each value must be exactly one of the observed objects written above, or None if none of them is what the phrase refers to.\\
Write nothing else: no explanation, no quotes, no extra lines.
\end{tcolorbox}
\vspace{-4pt}
\caption{Query used by Self-clarification at every step. \phb{observed\_objects} lists the object types observed so far without instance indices, and \phb{phrases} lists the quoted referring expressions of the instruction.}
\label{fig:selfclarify_prompt}
\end{figure}

Figures~\ref{appenfigure:heatmap_no_method}--\ref{appenfigure:heatmap_ours} decompose Table~\ref{tab:experiment_main} on ALFWorld-Para into all combinations of the nine action-axis types and the three object-axis types. Columns are the object axis and rows the action axis, with each panel's color pattern showing how a method affects the two axes separately.

Under Baseline (Figure~\ref{appenfigure:heatmap_no_method}), the rightmost column, Same-Polarity, turns red in every panel. This is sharpest for RL-finetuned sLLMs, where the left two columns stay green and only the right column collapses, forming a vertical boundary. The dominance of the object axis found in \S~\ref{analysis:analysis1} is therefore consistent across action types.

Greedy (Figure~\ref{appenfigure:heatmap_greedy}) erases this boundary but darkens the panel as a whole. Same-Polarity brightens while None and Addition fall below Baseline, since Greedy substitutes another object before the true referent is observed.
StraTA-1.5B constitutes a notable exception: Greedy surpasses BINDIntent on SP (63.7 vs.\ 56.1) while scarcely degrading None (84.4 $\rightarrow$ 81.9), in contrast to StraTA-7B (89.6 $\rightarrow$ 67.2; Table~\ref{tab:experiment_main}).
StraTA devises a strategy at the outset of each episode and acts upon it thereafter.
Presumably owing to its limited capacity, StraTA-1.5B tends to compose vague initial strategies that prescribe indiscriminate search, whereas StraTA-7B formulates concrete strategies anchored to environment elements such as object and receptacle names.
Under a premature substitution, the 1.5B agent consequently continues to explore, whereas the 7B agent commits to a strategy built around the substituted object.

Self-clarification (Figure~\ref{appenfigure:heatmap_self_clar}) shows a pattern that varies with model size. For pretrained LLMs and 7B RL-finetuned sLLMs, Same-Polarity tends to brighten enough to offset the loss in None, whereas for 1.5B models it changes little while None and Addition drop. Because it also adds one LLM call per step even when the object is unchanged, the number of calls grows with every trajectory (Figure~\ref{fig:experiment_latency}).

BINDIntent\textsubscript{SBERT} (Figure~\ref{appenfigure:heatmap_sbert}) raises Same-Polarity by a limited margin, blurring the vertical boundary without removing it.

Under BINDIntent (Figure~\ref{appenfigure:heatmap_ours}) the boundary disappears in most panels, but how it closes differs by model. In GiGPO-7B, Same-Polarity spreads from 38.1 to 64.2 across action types, following the spread of its own None column (56.2--95.1). In StraTA-1.5B, Same-Polarity flattens to 50.9--59.7, more uniform than its None column, which still spans 72.1--90.3. Addition rises most in the rows lowest under Baseline, Question (58.0 $\rightarrow$ 73.9) and Hint (55.3 $\rightarrow$ 70.8), narrowing the spread across rows from 24.8 to 16.4. In both panels, Question and Hint are also the lowest rows in the None column, hence their darkness under Same-Polarity follows the difficulty of these action types rather than a residual failure of binding. Across the five heatmaps, the other methods pay for a gain on one axis with a loss on the other, while only BINDIntent raises Same-Polarity without giving up either.

The magnitude of this gain, however, differs between model groups: BINDIntent improves SP considerably more for RL-finetuned sLLMs (+17.5--49.1\,pp) than for pretrained LLMs (+9.6--21.1\,pp). RL-finetuned agents possess environment-specific priors over exploration and action, and their failures under SP originate predominantly from binding, as evidenced by a wide None--SP gap (42.8--77.4\,pp); once binding is resolved, these priors translate more directly into task success. Pretrained LLMs lack such priors, and their narrower gap (27.4--45.8\,pp) is attributable to two sources. Models with at least 14B parameters already resolve synonyms in part, retaining 41--58\% of their None SR under SP, and the 7--8B models remain constrained by low SR even on unchanged objects. In both cases, determining how to proceed with the task persists as a bottleneck even after binding.

\input{tables/exp_appen_abl_tau}

\input{tables/exp_appen_abl_kappa}
\subsubsection{Ablation Study}
\label{appen:ablation}
We ablate BINDIntent on GiGPO-7B and StraTA-1.5B, which differ in both scale and codebase, varying one component at a time with all others fixed.
Results are averaged over three seeds, and shaded rows denote the setting used in all main experiments. 
BINDIntent accounts for unseen candidates through two mechanisms: the lexical allowance defers non-exact bindings early in an episode, and persistence defers them while unexplored units may still hold a closer candidate.
The ablations show that both mechanisms contribute and that their default settings lie in a stable range.

\paragraph{Lexical allowance.}
With $\Delta\tau=0$, only exact matches are permitted. No paraphrase is ever bound and the results coincide with Baseline (Table~\ref{tab:exp_appen_abl_tau}; cf.\ Table~\ref{tab:experiment_main}).
Increasing $\Delta\tau$ admits synonyms and recovers SP, which remains stable for $\Delta\tau$ between 0.02 and 0.05.
A larger step admits distant candidates before exploration can rule out an unseen referent: at $\Delta\tau=0.20$, SP falls to 50.2 and 49.8, and None and Addition also decline, as expressions whose referents have not yet been observed are bound to other objects.
The allowance is thus effective only when it widens in step with exploration.

\paragraph{Unexplored units.}
$\kappa$ scales the number of chances for an unseen candidate to be closer than the current best match (Table~\ref{tab:exp_appen_abl_kappa}).
With $\kappa=0$, persistence ignores unexplored units, and a slot is bound as soon as a candidate enters the allowance.
This yields the highest SP but lowers None and Addition by 3.1--5.3\,pp relative to $\kappa=1$, since expressions are bound before their true referents are observed.
As $\kappa$ grows, binding is deferred more strongly, preserving None and Addition at the cost of SP.
Accounting for unexplored units thus prevents premature binding on instructions that require no substitution.
Although $\kappa=0.25$ yields the highest All, calibrating $\kappa$ amounts to estimating the density of objects per unit, which requires environment-specific knowledge.
We therefore adopt $\kappa=1$, which counts each unexplored unit as a single chance without tuning.

\input{tables/exp_appen_abl_nprime}
\input{tables/exp_appen_abl_component}

\paragraph{Persistence estimate.}
\emph{Fallback-only persistence} replaces the Weibull model with the exchangeability-based estimate, which ignores candidate distances (Table~\ref{tab:exp_appen_abl_component}).
SP drops to 38.3 and 36.7, as it cannot distinguish a distinctive best match from an ambiguous one.
Modeling the distance distribution is thus essential for judging whether an unseen candidate may be closer.
The fallback threshold $c$ has little effect, with All varying within 0.6\,pp for $c=3$--$7$ (Table~\ref{tab:exp_appen_abl_nprime}).

\paragraph{Binding rule and centering.}
\emph{Deterministic binding} commits every slot once $P^{\max}_t$ exceeds 0.5, slightly raising SP but lowering Addition by 2.7 and 2.8\,pp and yielding a lower All.
The hazard instead binds a slot by step $t$ with probability equal to $P^{\max}_t$, calibrating commitment to persistence without a fixed threshold.
\emph{No centering} lowers SP by 2.7 and 3.0\,pp, since components shared across a scene dominate the raw similarities and obscure the signal that distinguishes candidates.

\subsection{Latency}
\label{appen:Experiment:Latency}
\input{appenfigure/latency_breakdown_table}
\paragraph{Principle.}
Latency must be measured under identical conditions for every method. The
compared methods, however, rewrite the instruction, and measuring them on live rollouts would let each method alter the trajectory it is timed on. A rewrite
can turn a failure into a success or vice versa and thereby change the episode
length; since later steps contain many more observed objects, a different mix of
short and long episodes would bias the per-step statistics. Attaching a method
to a live policy would likewise alter the policy's inputs, leaving the methods timed on different distributions. We therefore measure all methods by
\emph{open-loop replay} of a fixed set of trajectories: each logged step supplies
the observation, the admissible actions, and the executed action, and every
method is run on exactly the same sequence of steps.

\paragraph{Trajectory set.}
The trajectories are collected from Qwen2.5-7B-Instruct running as a ReAct agent on ALFWorld-Para without any binding method, to keep the observed scenes free of influence from the compared methods. To cover all task types and a range of episode lengths, we sample one success stratum and one failure stratum for each of the six ALFWorld task types, with five episodes per stratum, giving 60 episodes (30 successes and 30 failures).
Failed episodes run to the 50-step limit, and half of the set accordingly consists of long-horizon trajectories in which many objects have been observed. Every method replays every step of these episodes in order, since its
internal state accumulates over an episode.

\paragraph{Environment.}
All measurements are taken on a single NVIDIA A100 80GB PCIe GPU with batch size
one, on a server with two Intel Xeon Gold 6338 CPUs (2.0\,GHz, 64 cores in total)
and 384\,GB of RAM, running Ubuntu 22.04 with CUDA 12.2; the second GPU on the
server is left idle. We use vLLM~\citep{kwon2023efficient} as an in-process engine, which keeps network and serialization overhead out of every measurement. All measurements
start from a warm state: loading the language model and the methods and
initializing the ALFWorld environment are excluded, and ten warm-up generations are
discarded before timing begins. Each measurement starts when the observation of a
step becomes available and ends when the corresponding computation completes, with
GPU synchronization at both ends of every span that involves the GPU.

\paragraph{What is timed.}
Every method runs at every step. For BINDIntent and the greedy baseline, the
measured span covers the entire computation of the method: updating the observed
objects and units, obtaining embeddings, computing distances, estimating
persistence, deciding the binding, and rewriting the instruction. No language-model
call is involved. For BINDIntent\textsubscript{SBERT}, the span additionally includes
the encoder forward pass. Self-clarification, by definition, delegates reference
resolution to the language model itself; at every step it poses a single question
covering all slots, and its span consists mainly of that generation. This cost is
inherent to any approach that delegates clarification to an LLM: it is dominated by
decoding rather than prefill, recurs at every step, and grows with the length of the
answer, that is, with the number of slots in the instruction.

\paragraph{Caching.}
Since an embedding depends only on the word, a name embedded earlier in the same episode is reused, and the cache is cleared at the start of each episode. For BINDIntent\textsubscript{SBERT}, only the names newly observed at a step are encoded, in a single batched forward pass. Distances, centering, and the persistence estimate are recomputed at every step. The prefix cache of the language model is enabled within an episode and reset between episodes, as in the evaluation runs.

\paragraph{Reporting.}
We report the mean per-step time of each method over all replayed steps. Because the pretrained models are served at different precisions, methods should be compared within a model rather than by absolute times across models.

\paragraph{Results.}
Table~\ref{tab:latency_breakdown} decomposes the per-step time of each method.
Greedy, BINDIntent, and BINDIntent\textsubscript{SBERT} do not depend on the agent model, and their totals differ by less than 0.03\,ms across models. BINDIntent takes 0.38\,ms per step, 0.13\,ms more than Greedy, which corresponds to the persistence
estimate and the binding decision. Replacing Numberbatch with SBERT
(all-mpnet-base-v2) raises the total to 2.86\,ms, almost entirely due to the encoder forward pass (2.41\,ms). Because only newly observed names are encoded, this cost is concentrated in the steps where new objects appear, and the median step takes only 0.53\,ms. The subsequent stages are also slightly slower with SBERT (e.g., 0.20 vs.\ 0.15\,ms for distance computation), as its embeddings have 768 dimensions
compared with 300 for Numberbatch.

For Self-clarification, the additional LLM call accounts for more than 98\% of the step time, and the cost grows with model size, from about 30\,ms for 1.5B models to 107\,ms for Mistral-Small-3-24B. Its binding stage, which builds the clarification query with the chat template of the model, varies with the tokenizer rather than with model size or training: the ten models sharing the Qwen2.5 tokenizer, from 1.5B to 7B and across all four RL recipes, cluster at 0.29--0.33\,ms, whereas Qwen3-14B and Mistral-Small-3-24B take 0.47 and 0.57\,ms.

\paragraph{Precision.}
Llama3.1-8B, Qwen3-14B, and Mistral-Small-3-24B are served as FP8 checkpoints. Since the A100 does not natively support FP8, vLLM runs them with weight-only FP8 kernels, which may be slower than on FP8-capable hardware. This affects only the LLM call of Self-clarification, not the LLM-free methods, and the gap between the two spans two orders of magnitude; the comparison is therefore unaffected.

%% file: tables/exp_appen_abl_tau.tex
\begin{table}[t]
  \centering
  \scriptsize
  \setlength{\tabcolsep}{3pt}
  \renewcommand{\arraystretch}{0.95}
  \resizebox{\columnwidth}{!}{%
  \begin{tabular}{c!{\color{black}\vrule}cccc!{\color{black}\vrule}cccc}
    \toprule
    & \multicolumn{4}{c}{GiGPO-7B} & \multicolumn{4}{c}{StraTA-1.5B} \\
    \cmidrule(lr){2-5} \cmidrule(lr){6-9}
    $\Delta\tau$ & None & Add & Same-P & All & None & Add & Same-P & All \\
    \midrule
    0.20          & 87.4 & 82.1 & 50.2 & 72.7 & 81.4 & 74.2 & 49.8 & 68.0 \\
    0.10          & 84.6 & 85.6 & 54.6 & 74.6 & 86.1 & 80.8 & 54.3 & 73.3 \\
    0.05          & 88.5 & 87.0 & 58.1 & 77.5 & 84.5 & 82.1 & 55.8 & 73.8 \\
    \rowcolor{oursbg}
    0.04          & 88.9 & 86.8 & 57.6 & 77.4 & 85.2 & 81.2 & 56.1 & 73.8 \\
    0.02          & 89.0 & 86.7 & 56.4 & 77.0 & 84.7 & 77.4 & 55.7 & 72.2 \\
    0.00          & 89.2 & 86.8 & 12.1 & 61.8 & 84.4 & 71.3 & \phantom{0}7.0 & 53.2 \\
    \bottomrule
  \end{tabular}%
  }
  \vspace{-5pt}
  \caption{Ablation of the lexical allowance step $\Delta\tau$ ($\tau_t = t\,\Delta\tau$); $\Delta\tau = 0$ permits only exact matches throughout the episode. Success rate (\%) on ALFWorld-Para by object-axis paraphrase type; the shaded row is the setting used in all main experiments.}
  \vspace{-5pt}
  \label{tab:exp_appen_abl_tau}
\end{table}

%% file: tables/exp_appen_abl_kappa.tex
\begin{table}[t]
  \centering
  \scriptsize
  \setlength{\tabcolsep}{3pt}
  \renewcommand{\arraystretch}{0.95}
  \resizebox{\columnwidth}{!}{%
  \begin{tabular}{c!{\color{black}\vrule}cccc!{\color{black}\vrule}cccc}
    \toprule
    & \multicolumn{4}{c}{GiGPO-7B} & \multicolumn{4}{c}{StraTA-1.5B} \\
    \cmidrule(lr){2-5} \cmidrule(lr){6-9}
    $\kappa$ & None & Add & Same-P & All & None & Add & Same-P & All \\
    \midrule
    0          & 85.5 & 81.6 & 65.4 & 77.2 & 82.1 & 75.9 & 63.4 & 73.5 \\
    0.25          & 88.0 & 86.7 & 63.1 & 78.9 & 84.9 & 81.0 & 60.8 & 75.2 \\
    0.5           & 89.4 & 86.2 & 61.7 & 78.7 & 85.4 & 81.6 & 58.1 & 74.7 \\
    \rowcolor{oursbg}
    1             & 88.9 & 86.8 & 57.6 & 77.4 & 85.2 & 81.2 & 56.1 & 73.8 \\
    2             & 88.6 & 87.3 & 52.4 & 75.7 & 85.4 & 81.7 & 51.9 & 72.6 \\
    4             & 89.0 & 88.2 & 47.6 & 74.4 & 85.6 & 82.3 & 45.6 & 70.7 \\
    8             & 88.9 & 88.1 & 41.5 & 72.3 & 85.2 & 81.6 & 40.3 & 68.5 \\
    \bottomrule
  \end{tabular}%
  }
  \vspace{-5pt}
  \caption{Ablation of the scale $\kappa$ on the number of remaining chances, $k_t \to \kappa\, k_t$; $\kappa = 1$ counts each unexplored unit once. Success rate (\%) on ALFWorld-Para by object-axis paraphrase type; the shaded row is the setting used in all main experiments.}
  \vspace{-5pt}
  \label{tab:exp_appen_abl_kappa}
\end{table}

%% file: tables/exp_appen_abl_nprime.tex
\begin{table}[t]
  \centering
  \scriptsize
  \setlength{\tabcolsep}{3pt}
  \renewcommand{\arraystretch}{0.95}
  \resizebox{\columnwidth}{!}{%
  \begin{tabular}{c!{\color{black}\vrule}cccc!{\color{black}\vrule}cccc}
    \toprule
    & \multicolumn{4}{c}{GiGPO-7B} & \multicolumn{4}{c}{StraTA-1.5B} \\
    \cmidrule(lr){2-5} \cmidrule(lr){6-9}
    $c$ & None & Add & Same-P & All & None & Add & Same-P & All \\
    \midrule
    \rowcolor{oursbg}
    3             & 88.9 & 86.8 & 57.6 & 77.4 & 85.2 & 81.2 & 56.1 & 73.8 \\
    4             & 88.6 & 86.8 & 57.3 & 77.2 & 85.3 & 80.8 & 56.3 & 73.8 \\
    5             & 89.2 & 86.8 & 58.3 & 77.7 & 85.5 & 81.4 & 55.5 & 73.7 \\
    6             & 88.6 & 86.9 & 59.0 & 77.8 & 85.1 & 82.1 & 55.9 & 74.0 \\
    7             & 89.3 & 86.9 & 57.8 & 77.6 & 85.7 & 81.5 & 55.9 & 74.0 \\
    \bottomrule
  \end{tabular}%
  }
  \vspace{-5pt}
  \caption{Ablation of the fallback threshold $c$ below which \textsc{Persistence} falls back to the exchangeability-based estimate. Success rate (\%) on ALFWorld-Para by object-axis paraphrase type; the shaded row is the setting used in all main experiments.}
  \vspace{-5pt}
  \label{tab:exp_appen_abl_nprime}
\end{table}

%% file: tables/exp_appen_abl_component.tex
\begin{table}[t]
  \centering
  \scriptsize
  \setlength{\tabcolsep}{3pt}
  \renewcommand{\arraystretch}{0.95}
  \resizebox{\columnwidth}{!}{%
  \begin{tabular}{l!{\color{black}\vrule}cccc!{\color{black}\vrule}cccc}
    \toprule
    & \multicolumn{4}{c}{GiGPO-7B} & \multicolumn{4}{c}{StraTA-1.5B} \\
    \cmidrule(lr){2-5} \cmidrule(lr){6-9}
    Variant & None & Add & Same-P & All & None & Add & Same-P & All \\
    \midrule
    \rowcolor{oursbg}
    BINDIntent                       & 88.9 & 86.8 & 57.6 & 77.4 & 85.2 & 81.2 & 56.1 & 73.8 \\
    \quad Deterministic binding      & 88.2 & 84.1 & 59.0 & 76.7 & 84.5 & 78.4 & 58.1 & 73.2 \\
    \quad Fallback-only persistence  & 89.3 & 82.1 & 38.3 & 69.2 & 85.6 & 76.4 & 36.7 & 65.6 \\
    \quad No centering               & 88.4 & 87.1 & 54.9 & 76.4 & 85.1 & 81.0 & 53.1 & 72.7 \\
    \bottomrule
  \end{tabular}%
  }
  \vspace{-5pt}
  \caption{Component ablation (SR, \%) on ALFWorld-Para. \emph{Deterministic binding}: bind once $P^{\max}_t > 0.5$; \emph{Fallback-only persistence}: $|V_t|/(|V_t|+k_t)$ in place of the Weibull model; \emph{No centering}: \textsc{Match} on raw embeddings. Shaded: full method.}
  \vspace{-5pt}
  \label{tab:exp_appen_abl_component}
\end{table}

%% file: appenfigure/latency_breakdown_table.tex
\newcolumntype{R}{>{\raggedleft\arraybackslash}X}
\definecolor{ptbg}{HTML}{EAF0FD}
\definecolor{rlbg}{HTML}{FDEFE8}
\begin{table}[t]
\centering
\footnotesize
\setlength{\tabcolsep}{1.6pt}
\begin{tabular}{@{}lrrrrr@{\hspace{2pt}}|@{\hspace{2pt}}r@{}}
\toprule
 & \multicolumn{5}{c}{Per stage} &  \\
\cmidrule(lr){2-6}
Method & Embed & Dist. & Weibull & LLM & Bind. & Total \\
\midrule
Greedy & 0.02 & 0.14 & -- & -- & 0.03 & 0.25 \\
BINDIntent$_{\text{SBERT}}$ & 2.41 & 0.20 & 0.10 & -- & 0.07 & 2.86 \\
BINDIntent & 0.02 & 0.15 & 0.09 & -- & 0.06 & 0.38 \\
\midrule
\rowcolor{ptbg}\multicolumn{7}{@{}l}{\textit{Self-clarification, Pretrained}} \\
\hspace{0.6em}Qwen2.5-1.5B & -- & -- & -- & 29.92 & 0.32 & 30.43 \\
\hspace{0.6em}Qwen2.5-7B & -- & -- & -- & 60.84 & 0.29 & 61.30 \\
\hspace{0.6em}Llama3.1-8B & -- & -- & -- & 46.76 & 0.36 & 47.33 \\
\hspace{0.6em}Phi-4-14B & -- & -- & -- & 71.04 & 0.30 & 71.55 \\
\hspace{0.6em}Qwen3-14B & -- & -- & -- & 68.89 & 0.47 & 69.58 \\
\hspace{0.6em}Mistral-S-24B & -- & -- & -- & 106.10 & 0.57 & 106.88 \\
\rowcolor{rlbg}\multicolumn{7}{@{}l}{\textit{Self-clarification, RL-finetuned}} \\
\hspace{0.6em}PPO-1.5B & -- & -- & -- & 30.56 & 0.33 & 31.08 \\
\hspace{0.6em}GRPO-1.5B & -- & -- & -- & 30.33 & 0.30 & 30.81 \\
\hspace{0.6em}GiGPO-1.5B & -- & -- & -- & 30.74 & 0.33 & 31.26 \\
\hspace{0.6em}StraTA-1.5B & -- & -- & -- & 31.34 & 0.33 & 31.87 \\
\hspace{0.6em}PPO-7B & -- & -- & -- & 58.25 & 0.32 & 58.77 \\
\hspace{0.6em}GRPO-7B & -- & -- & -- & 59.42 & 0.33 & 59.93 \\
\hspace{0.6em}GiGPO-7B & -- & -- & -- & 56.67 & 0.33 & 57.20 \\
\hspace{0.6em}StraTA-7B & -- & -- & -- & 61.83 & 0.33 & 62.36 \\
\bottomrule
\end{tabular}
\caption{Per-step latency breakdown (ms). Embed: embedding of the referring expression and observed object names; Dist.: candidate distance computation; Weibull: persistence estimation; LLM: additional call to the agent LLM; Bind.: binding decision. Greedy, BINDIntent$_{\text{SBERT}}$, and BINDIntent do not call the agent LLM and are averaged over all agent models; Self-clarification is reported per agent model. Total includes observation update and instruction rewriting.}
\label{tab:latency_breakdown}
\end{table}

%% file: sections/97_appen_scipara.tex
\begin{table*}[t]
  \centering
  \scriptsize
  \renewcommand{\arraystretch}{1.15}
  \begin{tabularx}{\textwidth}{@{}>{\raggedright\arraybackslash}p{1.7cm}X@{}}
    \toprule
    Task & Action frame \\
    \midrule
    Conductivity &
    Your task is to determine if \texttt{[obj]} is electrically conductive. The \texttt{[obj]} is located around the \texttt{[obj]}. First, focus on the \texttt{[obj]}. If it is electrically conductive, place it in the red \texttt{[obj]}. If it is electrically nonconductive, place it in the green \texttt{[obj]}. \\
    Melting point &
    Your task is to measure the melting point of \texttt{[obj]}, which is located around the \texttt{[obj]}. First, focus on the \texttt{[obj]}. Next, focus on the \texttt{[obj]}. If the melting point of \texttt{[obj]} is above 150.0 degrees celsius, focus on the red \texttt{[obj]}. If the melting point of \texttt{[obj]} is below 150.0 degrees celsius, focus on the green \texttt{[obj]}. The \texttt{[obj]} are located around the \texttt{[obj]}. \\
    Find &
    Your task is to find a(n) \texttt{[obj]}. First, focus on the thing. Then, move it to the green \texttt{[obj]} in the \texttt{[obj]}. \\
    Life span &
    Your task is to find the \texttt{[obj]} with the longest life span. The \texttt{[obj]} are in the `outside' location. Focus on the \texttt{[obj]} with the longest life span. \\
    Chemistry &
    Your task is to use chemistry to create the substance `\texttt{[obj]}'. A \texttt{[obj]} and some of the ingredients might be found near the \texttt{[obj]}. When you are done, focus on the \texttt{[obj]}. \\
    Paint mixing &
    Your task is to use chemistry to create green \texttt{[obj]}. When you are done, focus on the green \texttt{[obj]}. \\
    \bottomrule
  \end{tabularx}
  \caption{Action frames of ScienceWorld-Para by task. \texttt{[obj]} denotes an object slot paraphrased on the object axis; all other text is kept fixed.}
  \label{tab:sciworld_example}
\end{table*}

\section{ScienceWorld-Para}
\label{appen:ScienceWorld-Para}
ScienceWorld-Para follows the construction and evaluation protocol of ALFWorld-Para (Appendix~\ref{appen:ALFWorld-Para}); this section describes only the components that differ.

\subsection{Benchmark Generation}
\label{appen:ScienceWorld-Para:BenchmarkGeneration}
 
\paragraph{Source episodes.}
We start from the ScienceWorld~\citep{wang2022scienceworld} test split used by StraTA~\citep{xue2026strata} (175 episodes, 11 task types), which covers the task types on which the RL policies were trained.
Test episodes differ from training episodes only in task variation, and their instructions follow the same templates as those seen during training.
As in ALFWorld-Para, paraphrasing is thus the sole source of linguistic variation relative to training.
We exclude task types whose targets are distinguished only by alphanumeric labels (e.g., \textit{unknown substance B}), since such names admit no synonym that preserves the referent.
From the remaining episodes, we remove those with duplicate goal instructions and sample 82 instructions preserving the task-type distribution, yielding $82 \times 29 = 2{,}378$ instructions.

\textbf{Paraphrase Generation.}
ScienceWorld-Para is produced with the factored pipeline of Appendix~\ref{app:para_generation}, using the same generation model (Gemini 3 Flash~\cite{gemini3}), the same type-specific suffixes (Figures~\ref{fig:prompt-action-suffixes-lexical}--\ref{fig:prompt-action-suffixes-pragmatic} and~\ref{fig:prompt-object-suffixes}), and the same verification and composition procedure; only the inputs and the environment-specific parts of the prompts differ.

\textbf{Action Axis.}
We replace every object slot in the task instructions with a placeholder and define the remainder as the action frame (Table~\ref{tab:sciworld_example}).
Frames are paraphrased with the action axis prefix (Figure~\ref{fig:prompt-action}) adapted to ScienceWorld placeholders.
Since a ScienceWorld instruction spans several sentences, each action axis type is applied once per instruction, and the remaining steps are kept verbatim.

\textbf{Object Axis.}
Every object slot, including targets, containers, rooms, and category names, is a target of object axis paraphrase, and candidates are generated with the object axis prefix (Figure~\ref{fig:prompt-object}).

\textbf{Verification and Composition.}
Candidates are filtered with the verifier prompts of Figure~\ref{fig:prompt-verifier}, and verified frames are composed with object paraphrases through the same slot-filling scheme with four inflected forms.
As in ALFWorld-Para (Table~\ref{tab:para_statistics}), the 2 object types, 9 action types, and their 18 compositions yield 29 paraphrase types per instruction.

\subsubsection{Parser Stability}
The same extraction procedure is applied without modification to ScienceWorld-Para,
whose instructions span multiple sentences and contain 4.56 slots on average.
Evaluated against the gold substitutions of all 2,378 instructions, the extraction
recovers 99.96\% of the 10,844 gold slots. Manual verification of 300 instructions,
sampled with stratification by paraphrase type and assessed under the same
criterion, found 296 to be correct (98.7\%). Extraction is performed once per
episode, requires 3.2\,ms, and is likewise shared by all binding methods, leaving
their relative cost unchanged.


\subsubsection{Human Evaluation}
\label{app:Sciworld_Human_Evaluation}

\paragraph{Annotation setup.}
We sampled 145 paraphrase--original pairs (6.1\% of the full 2,378-item dataset) via stratified sampling across the 29 paraphrase types, five per type.
Three annotators, blind to which label would favor the benchmark, independently judged each pair with a binary label (Figure~\ref{fig:sciworld-human-eval-instruction}): whether the two commands are consistent with the agent performing the same task.
A pair was marked equivalent only if the target object, destination, handling (cleaning, heating, cooling), and count all remained the same despite differences in wording, tone, sentence structure, or directive form; it was marked non-equivalent if any of these changed or if the intent was too vague to determine.
Annotators were given a reference list of the distinct entities in the environment---targets, rooms, and the animal, plant, living-thing, and non-living-thing categories---so that a renamed object could be checked against entities it might collide with.
Since ScienceWorld tasks hinge on measurements, the same list fixed the properties that must survive unchanged: melting point, electrical conductivity, life span, and the temperature threshold with its unit.
The three annotators were disjoint from those who annotated ALFWorld-Para; recruitment and the blinding protocol follow Appendix~\ref{app:Alfworld_Human_Evaluation}.

\paragraph{Experimental Results.}
Under majority vote, 142 of 145 pairs (97.9\%) were judged meaning-preserving, and 137 (94.5\%) unanimously so; none was unanimously rejected.
Adjudicating the 8 split-vote items against the entity reference, 7 were equivalent and 1 a failure: an object paraphrase that reduced the target to its head noun (\textit{jam sandwich} $\rightarrow$ \textit{sandwich}), leaving the substance the task asks the agent to synthesize under-specified.
In total, 144 of 145 (99.3\%) preserve the original task semantics.
Unlike those in ALFWorld-Para, the disagreements are not confined to the lexical axis: three involve no object substitution and arise purely from pragmatic reframing, where one annotator read a question or hint form (e.g., \textit{Your task is to determine if paper clip is electrically conductive} $\rightarrow$ \textit{It remains a mystery whether the paper clip is electrically conductive}) as not explicitly instructing the action.
The other five pair an object substitution with such a reframing, and the substituted terms, such as \textit{animal} $\rightarrow$ \textit{beast}, \textit{living room} $\rightarrow$ \textit{front room}, and \textit{green paint} $\rightarrow$ \textit{green colorant}, name no other distinct entity.
No case altered the measured property, the temperature threshold, or the goal state.

\input{tables/appen_scipara_sr_drop}
\paragraph{Retention of flagged items.}
All 145 sampled pairs, including the eight that drew split votes, were retained in the benchmark by design.
Terms such as \textit{beast} and \textit{colorant} are attested synonyms of the intended target, and the question and hint forms are exactly the pragmatic reframings the benchmark is built to measure; excluding the pairs that human annotators found hardest to judge would remove the upper end of the difficulty range rather than correct an error.
We retain them to probe whether instruction understanding holds at the margins of valid paraphrase rather than only for clear-cut cases.
The single adjudicated failure was also retained: although under-specified in this task, referring to an object by its head noun alone is an uncommon but natural user expression, and one pair in 145 cannot meaningfully shift the aggregate scores we report.

\paragraph{Ethics.}
Annotators were unpaid volunteers from our research group who agreed to have their judgments used and reported in this work; given this affiliation, they were blind to which outcome would favor the benchmark (Appendix~\ref{app:Alfworld_Human_Evaluation}).
The judged text was public and non-sensitive, and no personally identifiable information was collected.
We judged this not to constitute human subjects research under our institution's guidelines and did not seek IRB review.

\input{tables/appen_scipara_object_drop}
\subsection{Experimental Results}
We evaluate RL-finetuned sLLMs trained with PPO, GRPO, GiGPO, and StraTA on the shared Qwen2.5-7B backbone, on both ScienceWorld and ScienceWorld-Para.
Each agent runs with the step budget of its training codebase (20 steps for StraTA and 30 for BEACON; Appendix~\ref{appen:ALFWorld-Para:Setup}), and following StraTA, we report the task score (0--100) provided by ScienceWorld.
All models exhibit consistent degradation on ScienceWorld-Para, with relative score drops ranging from 10.1\% to 20.3\% (Table~\ref{tab:sciworldpara_score_drop}).
The extent of degradation varies across RL algorithms despite the shared backbone.
GiGPO-7B performs on par with StraTA-7B on ScienceWorld (63.7 vs.\ 63.9) but suffers the largest drop (20.3\%), trailing it by 4.5\,pp on ScienceWorld-Para (50.8 vs.\ 55.3).
GRPO-7B shows the smallest relative drop (10.1\%), although its low original score (30.7) makes it difficult to disentangle paraphrase sensitivity from baseline instability.
These results indicate that, as on ALFWorld-Para, advantages on template-based instructions do not necessarily transfer to paraphrased ones.
The following analyses examine action and object paraphrases separately to identify the transformations driving these losses.


\input{figures/appen_scipara_paraknow}

\subsubsection{Analysis 1: Object Synonym Substitution Is the Dominant Failure Factor}
\label{appen:sciworld:analysis1}
We evaluate the four RL-finetuned sLLMs without any method and report the average score per paraphrase type, grouped as in Figure~\ref{fig:analysis_object_drop}: Others averages the seven action types other than Question and Hint, and the action$\times$object columns pair each action group with Same-Polarity (Table~\ref{tab:sciworld_para_breakdown}).
Paraphrased instructions reduce scores by 10.1--20.3\% relative to the original instructions (Table~\ref{tab:sciworldpara_score_drop}), and as on ALFWorld-Para, the drop is driven by the object axis.
Action paraphrases leave the mean score nearly unchanged (50.4--52.2 vs.\ 50.9 under None), and Addition causes only a minor decline (47.7).
Unlike on ALFWorld-Para, even Question and Hint cause little degradation: ScienceWorld instructions consist of sequential steps (Table~\ref{tab:sciworld_example}), and since each action type is applied once per instruction, the remaining steps still state the required actions explicitly.
Same-Polarity, in contrast, lowers the mean score to 29.9, and crossing it with any action group yields a similar level (29.4--30.6).
This pattern holds for every policy, confirming that object synonym substitution is the dominant failure factor.
\input{figures/appen_scipara_seennottaken}
\input{tables/appen_scipara_result}
\subsubsection{Analysis 2: sLLMs Still Know What the Paraphrased Object Refers to}
\label{appen:sciworld:analysis2}
The probe follows Appendix~\ref{appen:analysis:analysis2}: it consists of every paraphrased object name in ScienceWorld-Para, and each prompt lists all entity types of ScienceWorld as answer choices.
We probe the Qwen2.5-7B backbone and its four RL-finetuned variants with the same five templates and greedy decoding.
The backbone achieves 71.2\% paraphrase knowledge, and its RL-finetuned variants remain at 69.8--73.5\% (Figure~\ref{fig:appen_scipara_paraknow}).
As on ALFWorld-Para, RL fine-tuning largely preserves paraphrase knowledge, whereas the same policies lose most of their score under Same-Polarity. Knowledge loss during fine-tuning therefore does not account for the degradation.

\subsubsection{Analysis 3: Agents Encounter the Target Object but Do Not Take It}
\label{appen:sciworld:analysis3}
The taken rate follows the definition in Appendix~\ref{appen:analysis:analysis3}.
Since a ScienceWorld instruction refers to several objects, the target at each step is the object that the agent must pick up at the current stage of the task, and it is observed when it appears in the observation at that step.
The backbone rarely takes the target even when it is observed (16.4--19.7\%), consistent with its low score (11.5 overall; Table~\ref{tab:experiment_sciworld}).
RL fine-tuning raises taken rates under None (40.3--87.7\%) and Addition (31.5--84.5\%) for every policy (Figure~\ref{fig:appen_scipara_seennottaken}).
Under Same-Polarity, however, taken rates fall below None for every policy (4.2--65.9\%), with GRPO dropping even below the backbone.
Thus, as on ALFWorld-Para, RL-finetuned agents take observed targets reliably under original labels but far less consistently under synonyms.

\input{tables/appen_scipara_latency}

\subsubsection{Main Results}

Table~\ref{tab:experiment_sciworld} compares all methods on ScienceWorld-Para under the setup of \S\ref{sec:experiment}.
On RL-finetuned sLLMs, BINDIntent improves SP by 7.0--10.2\,pp and All by 2.1--3.4\,pp over Baseline, while None and Addition remain within 1.6\,pp.
In contrast, Base-7B gains little (+0.3\,pp in All).
Its score is low even when the object is unchanged (13.8 under None), and it rarely takes the target even when observed (16.4--19.7\%; Appendix~\ref{appen:sciworld:analysis3}). The bottleneck is task-solving rather than binding, as on ALFWorld-Para (\S\ref{sec:MainResults}).
Greedy substitutes other objects before the true referent is observed, sharply lowering None (e.g., GiGPO-7B 62.6 $\rightarrow$ 24.7), and its All falls below Baseline on every RL-finetuned sLLM.
Self-clarification raises SP on most policies, surpassing BINDIntent on PPO-7B and StraTA-7B, but lowers None and All on GiGPO-7B (50.8 $\rightarrow$ 46.8 in All) and requires an additional agent LLM call at every step, at over 160$\times$ the latency of BINDIntent.
BINDIntent\textsubscript{SBERT} also improves SP, by a smaller margin than Numberbatch (+5.3--7.8\,pp).
Overall, BINDIntent consistently improves over Baseline across all RL-finetuned sLLMs while preserving performance on unchanged objects, reproducing the trends observed on ALFWorld-Para.

We measure per-step latency with the protocol of Appendix~\ref{appen:Experiment:Latency} (Table~\ref{tab:latency-scienceworld}).
Greedy, BINDIntent, and BINDIntent\textsubscript{SBERT} never call the agent LLM, leaving their latency independent of the backbone.
BINDIntent takes 0.78\,ms per step, higher than on ALFWorld-Para (0.38\,ms) because a ScienceWorld instruction contains more object slots to bind.
BINDIntent\textsubscript{SBERT} takes 12.4\,ms on average because it runs an encoder forward pass for each newly observed name, and a ScienceWorld observation lists all objects in the current room, including container contents; its cost is thus concentrated in the steps that reveal new names, and its median latency is 0.92\,ms.
Self-clarification takes 129.5--160.3\,ms on 7B backbones, exceeding even its 24B cost on ALFWorld-Para (107\,ms), since its single query per step must decode one referent per object slot and generation accounts for over 99\% of its latency.
\label{appen:MainResults:Heatmap}

%% file: tables/appen_scipara_sr_drop.tex
\begin{table}[t]
  \centering
  \scriptsize
  \setlength{\tabcolsep}{5.5pt}
  \renewcommand{\arraystretch}{0.95}
  \resizebox{\columnwidth}{!}{%
  \begin{tabular}{lcc>{\color{dropred}}c}
    \toprule
    \multicolumn{1}{c}{Model} & ScienceWorld & ScienceWorld-Para & Score Drop (\%) \\
    \midrule
    PPO-7B        & 45.2 & 38.1 & $-$15.7 \\
    GRPO-7B       & 30.7 & 27.6 & $-$10.1 \\
    GiGPO-7B      & 63.7 & 50.8 & $-$20.3 \\
    StraTA-7B     & 63.9 & 55.3 & $-$13.5 \\
    \bottomrule
  \end{tabular}%
  }
  \vspace{-5pt}
  \caption{Average score on ScienceWorld and ScienceWorld-Para. Score Drop: relative change.}
  \vspace{-5pt}
  \label{tab:sciworldpara_score_drop}
\end{table}

%% file: tables/appen_scipara_object_drop.tex
\begin{table}[t]
  \centering
  \scriptsize
  \setlength{\tabcolsep}{2.6pt}
  \renewcommand{\arraystretch}{0.95}
  \resizebox{\columnwidth}{!}{%
  \begin{tabular}{l!{\color{black}\vrule}c!{\color{black}\vrule}ccc!{\color{black}\vrule}cc!{\color{black}\vrule}ccc}
    \toprule
    & \multicolumn{1}{c}{ScienceWorld} & \multicolumn{8}{c}{ScienceWorld-Para} \\
    \cmidrule(lr){2-2} \cmidrule(lr){3-10}
    & & \multicolumn{3}{c}{Action} & \multicolumn{2}{c}{Object} & \multicolumn{3}{c}{Action $\times$ Same-P} \\
    \cmidrule(lr){3-5} \cmidrule(lr){6-7} \cmidrule(lr){8-10}
    \multicolumn{1}{c}{Model} & None & Others & Quest. & Hint & Add & Same-P & Others & Quest. & Hint \\
    \midrule
    PPO-7B      & 45.2 & 46.7 & 46.7 & 42.3 & 43.7 & 26.5 & 25.1 & 27.8 & 23.2 \\
    GRPO-7B     & 30.7 & 36.2 & 36.2 & 33.6 & 32.1 & 14.7 & 16.3 & 13.0 & 17.1 \\
    GiGPO-7B    & 63.7 & 63.2 & 55.3 & 63.8 & 54.9 & 36.0 & 35.7 & 35.4 & 36.5 \\
    StraTA-7B   & 63.9 & 62.8 & 63.3 & 61.8 & 60.3 & 42.5 & 43.4 & 46.1 & 41.0 \\
    \midrule
    Mean        & 50.9 & 52.2 & 50.4 & 50.4 & 47.7 & 29.9 & 30.1 & 30.6 & 29.4 \\
    \bottomrule
  \end{tabular}%
  }
  \vspace{-5pt}
  \caption{Average score of RL-finetuned Qwen2.5-7B (Baseline) on ScienceWorld-Para by action and object paraphrase type. Others: mean over the seven action types other than Question and Hint.}
  \vspace{-5pt}
  \label{tab:sciworld_para_breakdown}
\end{table}

%% file: figures/appen_scipara_paraknow.tex
\begin{figure}[t]
  \centering
  \includegraphics[width=\columnwidth]{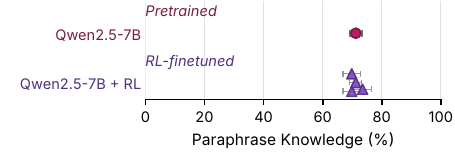}
  \caption{Paraphrase knowledge (\%) on ScienceWorld-Para of the pretrained Qwen2.5-7B backbone (top) and its RL-finetuned variants (bottom). The RL row shows four RL algorithms; error bars denote the standard deviation across the five probe templates.}
  \label{fig:appen_scipara_paraknow}
\end{figure}

%% file: figures/appen_scipara_seennottaken.tex
\begin{figure}[t]
  \centering
  \includegraphics[width=\columnwidth]{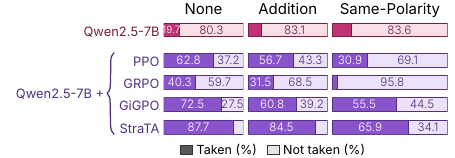}
  \caption{
  Taken rate (\%): among steps in which the target object is observed, up to and including its first take, the proportion of those at which the agent takes it. The top row denotes the shared Qwen2.5-7B backbone.
  }
  \label{fig:appen_scipara_seennottaken}
\end{figure}

%% file: tables/appen_scipara_result.tex
\newcolumntype{Q}{w{c}{16pt}}

\begin{table*}[t]
  \centering
  \scriptsize
  \setlength{\tabcolsep}{1.8pt}
  \renewcommand{\arraystretch}{0.95}
  \begin{tabular}{l!{\color{black}\vrule}QQQQ!{\color{black}\vrule}QQQQ!{\color{black}\vrule}QQQQ!{\color{black}\vrule}QQQQ!{\color{black}\vrule}>{\columncolor{oursbg}}Q>{\columncolor{oursbg}}Q>{\columncolor{oursbg}}Q>{\columncolor{oursbg}}Q}
    \toprule
    \multicolumn{1}{c}{Model}
      & \multicolumn{4}{c}{Baseline}
      & \multicolumn{4}{c}{Greedy}
      & \multicolumn{4}{c}{Self-clarification}
      & \multicolumn{4}{c}{BINDIntent\textsubscript{SBERT}}
      & \multicolumn{4}{c}{\cellcolor{oursbg}BINDIntent (ours)} \\
    \cmidrule(lr){2-5} \cmidrule(lr){6-9} \cmidrule(lr){10-13} \cmidrule(lr){14-17} \cmidrule(lr){18-21}
    \multicolumn{1}{r}{\tiny Split =} & \multicolumn{1}{c}{\tiny None} & \multicolumn{1}{c}{\tiny Add} & \multicolumn{1}{c}{\tiny Same-P} & \multicolumn{1}{c}{\tiny All} & \multicolumn{1}{c}{\tiny None} & \multicolumn{1}{c}{\tiny Add} & \multicolumn{1}{c}{\tiny Same-P} & \multicolumn{1}{c}{\tiny All} & \multicolumn{1}{c}{\tiny None} & \multicolumn{1}{c}{\tiny Add} & \multicolumn{1}{c}{\tiny Same-P} & \multicolumn{1}{c}{\tiny All} & \multicolumn{1}{c}{\tiny None} & \multicolumn{1}{c}{\tiny Add} & \multicolumn{1}{c}{\tiny Same-P} & \multicolumn{1}{c}{\tiny All} & \multicolumn{1}{c}{\cellcolor{oursbg}\tiny None} & \multicolumn{1}{c}{\cellcolor{oursbg}\tiny Add} & \multicolumn{1}{c}{\cellcolor{oursbg}\tiny Same-P} & \multicolumn{1}{c}{\cellcolor{oursbg}\tiny All} \\      \rowcolor{basebg}
    Base-7B       & 13.8 & 11.6 & \phantom{0}9.0 & 11.5 & 14.0 & 14.2 & 11.6 & \textbf{13.3} & 14.2 & 13.5 & \textbf{12.0} & 13.2 & 14.0 & 12.2 & \phantom{0}9.5 & 11.9 & \cellcolor{oursbg}13.5 & \cellcolor{oursbg}12.6 & \cellcolor{oursbg}\phantom{0}9.4 & \cellcolor{oursbg}11.8 \\
    \midrule
    PPO-7B        & 46.1 & 42.8 & 25.3 & 38.1 & 22.1 & 18.8 & 18.2 & 19.7 & 43.3 & 43.8 & \textbf{37.9} & \textbf{41.7} & 44.3 & 41.7 & 32.6 & 39.5 & 45.8 & 43.0 & 34.3 & 41.0 \\
    GRPO-7B       & 35.4 & 31.5 & 15.9 & 27.6 & 22.3 & 19.9 & 18.2 & 20.1 & 35.3 & 32.8 & 23.6 & 30.6 & 32.9 & 33.5 & 23.7 & 30.0 & 36.5 & 30.3 & \textbf{26.1} & \textbf{31.0} \\
    GiGPO-7B      & 62.6 & 54.1 & 35.8 & 50.8 & 24.7 & 22.2 & 21.6 & 22.9 & 56.0 & 50.1 & 34.2 & 46.8 & 62.2 & 53.7 & 41.1 & 52.3 & 62.7 & 53.2 & \textbf{42.8} & \textbf{52.9} \\
    StraTA-7B     & 62.9 & 59.8 & 43.3 & 55.3 & 45.7 & 40.7 & 36.4 & 41.0 & 62.9 & 59.9 & \textbf{52.1} & 58.3 & 61.3 & 60.6 & 50.3 & 57.4 & 63.0 & 61.4 & 51.7 & \textbf{58.7} \\
    \bottomrule
  \end{tabular}
  \caption{Average score (0--100) on ScienceWorld-Para by object-axis paraphrase type: None (object unchanged), Add (Addition), Same-P (Same-Polarity), and All. BINDIntent\textsubscript{SBERT} replaces Numberbatch embeddings with SBERT. Best per row in \textbf{bold} for Same-P and All.}
  \label{tab:experiment_sciworld}
\end{table*}

%% file: tables/appen_scipara_latency.tex
\begin{table}[t]
\centering
\footnotesize
\setlength{\tabcolsep}{3.5pt}
\begin{tabular}{@{}lrrrrr@{}}
\toprule
 & \multicolumn{5}{c}{Qwen2.5-7B} \\
\cmidrule(l){2-6}
Method & Base & PPO & GRPO & GiGPO & StraTA \\
\midrule
Greedy                      & 0.51  & 0.51  & 0.51  & 0.51  & 0.51  \\
Self-clarification          & 140.5 & 129.5 & 160.3 & 139.9 & 132.9 \\
BINDIntent$_{\text{SBERT}}$ & 12.4  & 12.4  & 12.4  & 12.4  & 12.4  \\
BINDIntent                  & 0.78  & 0.78  & 0.78  & 0.78  & 0.78  \\
\bottomrule
\end{tabular}
\caption{Per-step method latency (ms) on ScienceWorld-Para.}
\label{tab:latency-scienceworld}
\end{table}


%% file: sections/98_ai_assistants.tex
\section{AI Assistants}

Gemini 3 Flash~\cite{gemini3} is a component of the benchmark construction pipeline rather than a writing aid: it generates the paraphrases of ALFWorld-Para and ScienceWorld-Para under the factored procedure described in Appendices~\ref{appen:ALFWorld-Para:BenchmarkGeneration} and~\ref{app:para_generation}, and its outputs were validated by the human evaluation reported in Appendices~\ref{app:Alfworld_Human_Evaluation} and~\ref{app:Sciworld_Human_Evaluation}. Other AI assistants, namely Claude~\cite{anthropic2026opus} and ChatGPT~\cite{openai2026gpt}, were used only for writing polishing and code auto-completion. They were not used to develop research ideas, design experiments, analyze results, or write the scientific content of this paper. All experimental design, analysis, and conclusions are the authors' own.


\clearpage

\input{appenfigure/Appen_alfworld_task_example}

\input{appenfigure/Appen_prompt_action_prefix}
\input{appenfigure/Appen_prompt_action_type}

\input{appenfigure/Appen_prompt_object_prefix}

\input{appenfigure/Appen_prompt_object_type}
\input{appenfigure/Appen_prompt_verifier}

\input{figures/appen_alfworldpara_humaneval_guide}

\input{appenfigure/extended_object_drop}

\input{appenfigure/appen_paraknow_prompt}

\input{figures/appen_sciworldpara_humaneval_guide}


\input{appenfigure/heatmap_entire}

%% file: appenfigure/Appen_alfworld_task_example.tex
\begin{figure*}[t]
\centering

\begin{tcolorbox}[
  colback=white, colframe=black, boxrule=1pt,
  left=6pt, right=6pt, top=4pt, bottom=4pt,
  fontupper=\small\bfseries
]
Task: put some book on sidetable
\end{tcolorbox}

\vspace{4pt}

\begin{flushleft}
\begin{tcolorbox}[
  colback=blue!5, colframe=blue!40, boxrule=0.5pt,
  left=4pt, right=4pt, top=4pt, bottom=4pt,
  fontupper=\scriptsize\ttfamily,
  width=0.62\textwidth,
  title={\scriptsize\textbf{Environment Observation: Step 0}},
  coltitle=black, colbacktitle=blue!15
]
You are in the middle of a room. Looking quickly around you, you see
a bed 1, a diningtable 1, a drawer 1, a drawer 2, a drawer 3,
a drawer 4, a garbagecan 1, a sidetable 1, and a sidetable 2.\\[4pt]
\textnormal{\textbf{Admissible actions:}}\\
go to bed 1, go to diningtable 1, go to drawer 1, \ldots,
go to sidetable 1, go to sidetable 2, look, inventory
\textnormal{(12 actions)}
\end{tcolorbox}
\end{flushleft}

\vspace{-4pt}

\begin{flushright}
\begin{tcolorbox}[
  colback=orange!5, colframe=orange!50, boxrule=0.5pt,
  left=4pt, right=4pt, top=4pt, bottom=4pt,
  fontupper=\scriptsize\ttfamily,
  width=0.62\textwidth,
  title={\scriptsize\textbf{Strategy (StraTA only, generated once at Step 0)}},
  coltitle=black, colbacktitle=orange!15
]
\textnormal{\textit{Prepended to every subsequent action turn as context.}}\\[3pt]
First, I will look for a book and a sidetable.
If a book target is found, I will look for an action to perform
involving that book target. If a valid action involving this book
target leads to the target being put on a sidetable my job is done.
If a valid action involving this book target fails due to the
presence of another object, I will look for an alternative target
for this book action target which has the final objective of putting
a book target on a sidetable.
\end{tcolorbox}
\end{flushright}

\vspace{-4pt}

\begin{flushright}
\begin{tcolorbox}[
  colback=green!5, colframe=green!40, boxrule=0.5pt,
  left=4pt, right=4pt, top=4pt, bottom=4pt,
  fontupper=\scriptsize\ttfamily,
  width=0.62\textwidth,
  title={\scriptsize\textbf{Agent: Action 0}},
  coltitle=black, colbacktitle=green!15
]
\textnormal{\textit{(reasoning omitted)}}\\[3pt]
go to diningtable 1
\end{tcolorbox}
\end{flushright}

\vspace{-4pt}

\begin{flushleft}
\begin{tcolorbox}[
  colback=blue!5, colframe=blue!40, boxrule=0.5pt,
  left=4pt, right=4pt, top=4pt, bottom=4pt,
  fontupper=\scriptsize\ttfamily,
  width=0.62\textwidth,
  title={\scriptsize\textbf{Environment Observation: Step 1}},
  coltitle=black, colbacktitle=blue!15
]
You arrive at diningtable 1. On the diningtable 1, you see
a alarmclock 1, a book 3, a cd 1, a cd 3, a creditcard 2,
a keychain 1, a keychain 3, a laptop 1, a mug 1, a mug 2,
a mug 3, a pen 1, and a pencil 1.\\[4pt]
\textnormal{\textbf{Admissible actions:}}\\
examine diningtable 1, go to sidetable 1, \ldots,
take book 3 from diningtable 1, take laptop 1 from diningtable 1,
\ldots\ \textnormal{(25 actions)}
\end{tcolorbox}
\end{flushleft}

\vspace{-4pt}

\begin{flushright}
\begin{tcolorbox}[
  colback=green!5, colframe=green!40, boxrule=0.5pt,
  left=4pt, right=4pt, top=4pt, bottom=4pt,
  fontupper=\scriptsize\ttfamily,
  width=0.62\textwidth,
  title={\scriptsize\textbf{Agent: Action 1}},
  coltitle=black, colbacktitle=green!15
]
\textnormal{\textit{(reasoning omitted)}}\\[3pt]
take book 3 from diningtable 1
\end{tcolorbox}
\end{flushright}

\caption{Example interaction trajectory on ALFWorld. 
Environment observations (left, blue) provide text observations and 
admissible actions. The agent (right, green) selects actions based on 
the goal instruction. StraTA (orange) additionally generates a global 
strategy at Step 0, which is prepended to every subsequent action turn.}
\label{fig:alfworld_example}
\end{figure*}

%% file: appenfigure/Appen_prompt_action_prefix.tex
\begin{figure*}[t]
\centering
\begin{tcolorbox}[
  colback=red!5,
  colframe=red!50,
  boxrule=0.5pt,
  left=6pt, right=6pt, top=6pt, bottom=6pt,
  fontupper=\scriptsize\ttfamily,
  title={\small\textbf{Action Axis Paraphrase Generation Prompt}},
  coltitle=black,
  colbacktitle=red!15
]
Your task is to paraphrase a household agent instruction frame while maintaining
its semantic meaning. Generate diverse paraphrases that vary in expression but
preserve the core intent.\\[4pt]
The frame contains placeholders. Reproduce them EXACTLY:\\
~~\{obj:FORM\}~~~~the target object\\
~~\{recep:FORM\}~~the destination receptacle\\
~~\{toggle:FORM\}~the lamp, in look tasks\\[2pt]
FORM is one of: indefinite, definite, plural, bare. Choose the one your sentence
needs --- "put \{obj:indefinite\} on \{recep:definite\}" becomes
"put an apple on the desk".
Never write an article next to a placeholder; the form carries it.\\[4pt]
\textnormal{\textbf{CRITICAL RULES:}}\\
- NEVER change plurality (singular/plural)\\
- NEVER ADD NEW visual attributes (color, size, shape, material) where none existed\\
- NEVER add spatial attributes (position, location, direction)\\
- Preserve the exact sentence structure and action verb unless specified in the task\\
- Never add or drop a step or a NEW goal. ("put a hot X in Y" implies heating first;
  do not add washing, opening, etc.) Exception: the structural cells MAY make an
  already-implied preparation step explicit as a clause --- see their guidance.\\
- Preserve the GOAL STATE, never a METHOD or MECHANISM. Do not name an appliance or
  technique (microwave, oven, fridge) and do not swap in a verb that implies a
  different mechanism or object type: heating is "warm"/"heat" --- NOT "cook", "bake",
  "microwave", "reheat"; cooling is "cool"/"chill" --- NOT "refrigerate"/"freeze";
  cleaning is "clean"/"wash"/"rinse"/"scrub" --- NOT "launder" (fabric-only).\\
- NEVER change the object COUNT. A two-object task must keep "two" (or "both"/
  "a pair of"); a single-object task must stay singular.\\
- Keep the destination relation EXACTLY ("in"/"on" the receptacle). Do not weaken it
  to "near", "beside", "by", or "where the receptacle is".\\
- The look/examine task ("... under/with the lamp") means inspecting the HELD object
  while the lamp is on --- it is NOT a placement task. Keep a perception verb
  (look at / examine / inspect / view); never phrase it as putting, positioning,
  holding, or setting the object under/near the lamp.\\
- Only modify what is EXPLICITLY specified in the task-specific guidelines below\\[4pt]
\textnormal{\textbf{PARAPHRASE SCOPE:}}\\
- Change ONLY action-related elements; the placeholders carry the objects\\[4pt]
\textnormal{\textbf{OUTPUT REQUIREMENTS:}}\\
- Output ONLY the paraphrased frames, one per line\\
- Do NOT include explanations, meta-commentary, or multiple alternatives in one line\\
- Each frame should be a single, complete, natural instruction
\end{tcolorbox}
\caption{Prompt used for generating action axis paraphrases.
Placeholders \texttt{\{obj:FORM\}}, \texttt{\{recep:FORM\}}, and
\texttt{\{toggle:FORM\}} are preserved in the output and filled with
entity names at assembly time.}
\label{fig:prompt-action}
\end{figure*}

%% file: appenfigure/Appen_prompt_action_type.tex
\begin{figure*}[t]
\centering
\begin{tcolorbox}[
  colback=red!5, colframe=red!50, boxrule=0.5pt,
  left=6pt, right=6pt, top=6pt, bottom=6pt,
  fontupper=\scriptsize\ttfamily,
  title={\small\textbf{Action $\cdot$ Lexical $\cdot$ Addition}},
  coltitle=black, colbacktitle=red!15
]
\textnormal{\textbf{Task:}} Add or remove manner adverbs or action modifiers
(single words only).\\[2pt]
\textnormal{\textbf{Guidelines:}}\\
- Add a manner adverb ("carefully", "gently") or drop one that is present.\\
- The adverb must not imply a new sub-goal or constraint.\\[2pt]
\textnormal{\textbf{Examples:}}\\
- "put \{obj:indef\} in \{recep:def\}" -> "carefully put \{obj:indef\} in \{recep:def\}"\\
- "cool some \{obj:bare\} and put it in \{recep:def\}" -> "gently cool \{obj:indef\} and put it in \{recep:def\}"
\end{tcolorbox}
\vspace{3pt}
\begin{tcolorbox}[
  colback=red!5, colframe=red!50, boxrule=0.5pt,
  left=6pt, right=6pt, top=6pt, bottom=6pt,
  fontupper=\scriptsize\ttfamily,
  title={\small\textbf{Action $\cdot$ Lexical $\cdot$ Same-Polarity}},
  coltitle=black, colbacktitle=red!15
]
\textnormal{\textbf{Task:}} Replace action verbs with synonyms.\\[2pt]
\textnormal{\textbf{Guidelines:}}\\
- Replace the verb, keep structure and placeholders unchanged.\\
- Keep the same aspect and the same number of steps.\\[2pt]
\textnormal{\textbf{Examples:}}\\
- "put \{obj:indef\} in \{recep:def\}" -> "place \{obj:indef\} in \{recep:def\}"\\
- "heat some \{obj:bare\} and put it in \{recep:def\}" -> "warm up \{obj:indef\} and set it in \{recep:def\}"\\
- "examine the \{obj:bare\} with the \{toggle:bare\}" -> "inspect \{obj:def\} using \{toggle:def\}"
\end{tcolorbox}
\caption{Type-specific suffixes for the action axis (lexical types),
appended to the action axis prefix (Figure~\ref{fig:prompt-action}).}
\label{fig:prompt-action-suffixes-lexical}
\end{figure*}

\begin{figure*}[t]
\centering
\begin{tcolorbox}[
  colback=red!5, colframe=red!50, boxrule=0.5pt,
  left=6pt, right=6pt, top=6pt, bottom=6pt,
  fontupper=\scriptsize\ttfamily,
  title={\small\textbf{Action $\cdot$ Structural $\cdot$ Coordination}},
  coltitle=black, colbacktitle=red!15
]
\textnormal{\textbf{Task:}} Change the coordination structure of the action.\\[2pt]
\textnormal{\textbf{Guidelines:}}\\
- SINGLE-CLAUSE frame: split into coordinated clauses.
  Either coordinate pick-up + placement, or make the implied preparation
  explicit as its own clause.\\
- ALREADY-COORDINATED frame: re-coordinate the two clauses --- add explicit
  ordering ("First ..., then ...") or split into two sentences.\\[2pt]
\textnormal{\textbf{Examples:}}\\
- "put \{obj:indef\} in \{recep:def\}" -> "pick up \{obj:indef\} and put it in \{recep:def\}"\\
- "put a hot \{obj:bare\} in \{recep:def\}" -> "heat \{obj:bare\} and put it in \{recep:def\}"\\
- "clean some \{obj:bare\} and put it in \{recep:def\}" -> "First clean \{obj:bare\}, then put it in \{recep:def\}"
\end{tcolorbox}

\vspace{3pt}

\begin{tcolorbox}[
  colback=red!5, colframe=red!50, boxrule=0.5pt,
  left=6pt, right=6pt, top=6pt, bottom=6pt,
  fontupper=\scriptsize\ttfamily,
  title={\small\textbf{Action $\cdot$ Structural $\cdot$ Subordination}},
  coltitle=black, colbacktitle=red!15
]
\textnormal{\textbf{Task:}} Change subordination and nesting structure.\\[2pt]
\textnormal{\textbf{Guidelines:}}\\
- Use a subordinate clause: purpose, relative, or temporal.\\
- SINGLE-CLAUSE frame: may keep modifier or express preparation in subordinate
  clause ("once \{obj\} is hot, put it in \{recep\}").\\
- ALREADY-COORDINATED frame: turn first conjunct into subordinate clause.\\[2pt]
\textnormal{\textbf{Examples:}}\\
- "put \{obj:indef\} in \{recep:def\}" -> "once you have \{obj:indef\}, put it in \{recep:def\}"\\
- "clean some \{obj:bare\} and put it in \{recep:def\}" -> "once \{obj:def\} is clean, put it in \{recep:def\}"\\
- "heat some \{obj:bare\} and put it in \{recep:def\}" -> "after heating \{obj:bare\}, put it in \{recep:def\}"
\end{tcolorbox}
\caption{Type-specific suffixes for the action axis (structural types),
appended to the action axis prefix (Figure~\ref{fig:prompt-action}).}
\label{fig:prompt-action-suffixes-structural}
\end{figure*}

\begin{figure*}[t]
\centering
\begin{tcolorbox}[
  colback=red!5, colframe=red!50, boxrule=0.5pt,
  left=6pt, right=6pt, top=4pt, bottom=4pt,
  fontupper=\scriptsize\ttfamily,
  title={\small\textbf{Action $\cdot$ Pragmatic $\cdot$ Need Statement}},
  coltitle=black, colbacktitle=red!15
]
\textnormal{\textbf{Task:}} Express command as speaker's need or requirement.\\[2pt]
\textnormal{\textbf{Guidelines:}}\\
- Convert imperative to first-person need statement ("I need ...", "I want ...").\\
- State the desired END STATE, not an instruction to the listener.\\
- FORBIDDEN: "I need you to <verb> ..." --- that directs the listener.\\[2pt]
\textnormal{\textbf{Examples:}}\\
- "put \{obj:indef\} in \{recep:def\}" -> "I need \{obj:indef\} in \{recep:def\}."\\
- "clean some \{obj:bare\} and put it in \{recep:def\}" -> "I need \{obj:bare\} cleaned and in \{recep:def\}."
\end{tcolorbox}

\vspace{3pt}

\begin{tcolorbox}[
  colback=red!5, colframe=red!50, boxrule=0.5pt,
  left=6pt, right=6pt, top=4pt, bottom=4pt,
  fontupper=\scriptsize\ttfamily,
  title={\small\textbf{Action $\cdot$ Pragmatic $\cdot$ Embedded Imperative}},
  coltitle=black, colbacktitle=red!15
]
\textnormal{\textbf{Task:}} Embed command in a question frame with modal verbs.\\[2pt]
\textnormal{\textbf{Guidelines:}}\\
- Embed the action in a question frame ("Could you ...?", "Would you mind ...?").\\
- Interrogative; the action stays explicit.\\[2pt]
\textnormal{\textbf{Examples:}}\\
- "put \{obj:indef\} in \{recep:def\}" -> "Could you put \{obj:indef\} in \{recep:def\}?"\\
- "put two \{obj:plural\} in \{recep:def\}" -> "Would you mind putting two \{obj:plural\} in \{recep:def\}?"
\end{tcolorbox}

\vspace{3pt}

\begin{tcolorbox}[
  colback=red!5, colframe=red!50, boxrule=0.5pt,
  left=6pt, right=6pt, top=4pt, bottom=4pt,
  fontupper=\scriptsize\ttfamily,
  title={\small\textbf{Action $\cdot$ Pragmatic $\cdot$ Permission Directive}},
  coltitle=black, colbacktitle=red!15
]
\textnormal{\textbf{Task:}} Request the action as PERMISSION, but the LISTENER
still performs it.\\[2pt]
\textnormal{\textbf{Guidelines:}}\\
- Frame as permission request ("May I ...?", "Could I ...?"), using causative/possessive:
  "May I have \{obj\} in \{recep\}?".\\
- FORBIDDEN: "May I put/place/set \{obj\} ...?" --- speaker becomes the actor.\\[2pt]
\textnormal{\textbf{Examples:}}\\
- "put \{obj:indef\} in \{recep:def\}" -> "May I have \{obj:indef\} in \{recep:def\}?"\\
- "clean some \{obj:bare\} and put it in \{recep:def\}" -> "Could I have \{obj:bare\} cleaned and put in \{recep:def\}?"
\end{tcolorbox}

\vspace{3pt}

\begin{tcolorbox}[
  colback=red!5, colframe=red!50, boxrule=0.5pt,
  left=6pt, right=6pt, top=4pt, bottom=4pt,
  fontupper=\scriptsize\ttfamily,
  title={\small\textbf{Action $\cdot$ Pragmatic $\cdot$ Question Directive}},
  coltitle=black, colbacktitle=red!15
]
\textnormal{\textbf{Task:}} Pose an INTERROGATIVE that implies the desired action.\\[2pt]
\textnormal{\textbf{Guidelines:}}\\
- MUST end with "?".\\
- Ask about the state of the world, not the action; do not name the verb.\\
- Distinct from embedded imperative, which names the verb explicitly.\\[2pt]
\textnormal{\textbf{Examples:}}\\
- "put \{obj:indef\} in \{recep:def\}" -> "Is \{obj:def\} in \{recep:def\} yet?"\\
- "put a hot \{obj:bare\} in \{recep:def\}" -> "Is \{obj:def\} warm and in \{recep:def\} yet?"
\end{tcolorbox}

\vspace{3pt}

\begin{tcolorbox}[
  colback=red!5, colframe=red!50, boxrule=0.5pt,
  left=6pt, right=6pt, top=4pt, bottom=4pt,
  fontupper=\scriptsize\ttfamily,
  title={\small\textbf{Action $\cdot$ Pragmatic $\cdot$ Hint}},
  coltitle=black, colbacktitle=red!15
]
\textnormal{\textbf{Task:}} Make a DECLARATIVE statement that implies action
through inference.\\[2pt]
\textnormal{\textbf{Guidelines:}}\\
- MUST be declarative --- do NOT end with "?".\\
- State current condition, precondition, or desired state of the world.\\
- NEVER name a placement verb (put, place, set, store, move).\\[2pt]
\textnormal{\textbf{Examples:}}\\
- "put \{obj:indef\} in \{recep:def\}" -> "\{recep:def\} is still missing \{obj:indef\}."\\
- "put a clean \{obj:bare\} in \{recep:def\}" -> "\{obj:def\} is still dirty and not yet in \{recep:def\}."\\
- "put two \{obj:plural\} in \{recep:def\}" -> "\{recep:def\} is still short of two \{obj:plural\}."
\end{tcolorbox}
\caption{Type-specific suffixes for the action axis (pragmatic types),
appended to the action axis prefix (Figure~\ref{fig:prompt-action}).}
\label{fig:prompt-action-suffixes-pragmatic}
\end{figure*}

%% file: appenfigure/Appen_prompt_object_prefix.tex
\begin{figure*}[t]
\centering

\begin{tcolorbox}[
  colback=gray!5, colframe=gray!50, boxrule=0.5pt,
  left=6pt, right=6pt, top=6pt, bottom=6pt,
  fontupper=\scriptsize\ttfamily,
  title={\small\textbf{Object $\cdot$ Lexical $\cdot$ Addition}},
  coltitle=black, colbacktitle=gray!15
]
\textnormal{\textbf{Task:}} Keep the object's own noun EXACTLY and add ONE neutral
category or function word next to it.\\[2pt]
\textnormal{\textbf{Guidelines:}}\\
- The object's own noun must survive verbatim.\\
- Added word must be NON-SENSORY (no color, size, shape, material).\\
- The result must be a NATURAL phrase a person would actually say.\\[2pt]
\textnormal{\textbf{Examples:}}\\
- Apple -> apple fruit~~|~~SoapBar -> hand soap bar~~|~~Knife -> kitchen knife\\
- Mug -> drinking mug~~|~~Box -> storage box\\[2pt]
\textnormal{\textbf{Forbidden (sensory):}}
"red apple", "raw egg", "shiny cd", "metal fork", "ceramic mug"
\end{tcolorbox}
\vspace{3pt}
\begin{tcolorbox}[
  colback=gray!5, colframe=gray!50, boxrule=0.5pt,
  left=6pt, right=6pt, top=6pt, bottom=6pt,
  fontupper=\scriptsize\ttfamily,
  title={\small\textbf{Object $\cdot$ Lexical $\cdot$ Same-Polarity}},
  coltitle=black, colbacktitle=gray!15
]
\textnormal{\textbf{Task:}} Replace the object noun with a SYNONYM or a HYPONYM
(a more specific sub-kind or variety). Never a HYPERNYM (a broader category).\\[2pt]
\textnormal{\textbf{Guidelines:}}\\
- Replace the noun with a natural alternative that still picks out the SAME object.\\
- A synonym (Mug -> tumbler) OR a sub-kind / variety (Apple -> gala,
  Bread -> sourdough) is fine. Hyponyms are welcome.\\
- The object's own noun must NOT appear in the result.\\
- Keep the same grammatical number.\\
- Do NOT over-generalize to a broad category (a HYPERNYM).\\[2pt]
\textnormal{\textbf{Examples:}}\\
- SprayBottle -> spritzer / squirt bottle~~|~~GarbageCan -> trash can / waste bin\\
- Mug -> tumbler~~|~~Apple -> gala / granny smith~~|~~WineBottle -> merlot\\[2pt]
\textnormal{\textbf{Forbidden (HYPERNYM):}}
Apple -> "fruit";~~Fork -> "cutlery item";~~Pencil -> "writing utensil"
\end{tcolorbox}
\caption{Type-specific suffixes for the object axis, appended to the object axis
prefix (Figure~\ref{fig:prompt-object}).}
\label{fig:prompt-object-suffixes}
\end{figure*}

%% file: appenfigure/Appen_prompt_object_type.tex
\begin{figure*}[t]
\centering
\begin{tcolorbox}[
  colback=gray!5,
  colframe=gray!50,
  boxrule=0.5pt,
  left=6pt, right=6pt, top=6pt, bottom=6pt,
  fontupper=\scriptsize\ttfamily,
  title={\small\textbf{Object Axis Paraphrase Generation Prompt}},
  coltitle=black,
  colbacktitle=gray!15
]
Your task is to rename an object in a household environment while keeping the same referent.
Generate diverse alternative names that vary in expression but still pick out that object.\\[4pt]
\textnormal{\textbf{CRITICAL RULES:}}\\
- The name must not denote another entity in the scene (see the list below)\\
- NEVER add spatial attributes (position, location, direction)\\
- NEVER ADD a sensory attribute as a MODIFIER (color, size, shape, material,
  ripeness/state): the scene never states them, so "red apple", "raw egg",
  "shiny cd", "metal fork", "small statue" are all forbidden.
  (This bans an ADDED descriptor; it does not ban a synonym whose own fixed name
  happens to contain such a word, e.g.\ "graphite stick" for Pencil or
  "wine carafe" for WineBottle.)\\
- Natural English, lowercase --- write "spray bottle", not ALFWorld's "spraybottle"\\
- No proper nouns or brand names\\[4pt]
\textnormal{\textbf{ALFWORLD ENVIRONMENT ENTITIES:}}\\
\{ALFWORLD\_ENV\_OBJECTS\}\\[4pt]
\textnormal{\textbf{COMPOUND-WORD HANDLING:}}\\
- Compound NOUNS like "spray bottle", "alarm clock", "coffee table", "toilet paper"
  refer to a single environment object. Substitute the WHOLE compound at once,
  not word-by-word.\\
- Good: "SprayBottle" -> "spritzer" or "squirt bottle".\\
- Bad: "SprayBottle" -> "misting flask" (over-segmented, unnatural).\\[4pt]
\textnormal{\textbf{DISTINCTNESS BETWEEN THE TWO OBJ CELLS:}}\\
- same\_polarity replaces the object's own noun with a SYNONYM or a HYPONYM
  (sub-kind / variety): "Apple" -> "gala" or "SprayBottle" -> "spritzer"
  (the noun is gone). Never a HYPERNYM that over-generalizes
  ("Apple" -> "fruit" is WRONG).\\
- addition\_deletion keeps that noun EXACTLY and adds one category/function word
  or drops down to the bare noun: "Apple" -> "apple fruit" or bare "apple".
  Never a sensory word ("red apple" is WRONG).\\
- A name that keeps the noun does not belong in same\_polarity, and one that
  drops it does not belong in addition\_deletion.\\[4pt]
\textnormal{\textbf{OUTPUT REQUIREMENTS:}}\\
- Output ONLY JSON objects, one per line, in this exact shape:\\
~~\{{"bare": "spray bottle", "indefinite": "a spray bottle",
  "definite": "the spray bottle", "plural": "spray bottles"\}}\\
- All four forms filled and consistent. "indefinite" carries the right article
  ("an apple", not "a apple"); "plural" is properly inflected ("knives").\\
- Do NOT include explanations, meta-commentary, or multiple alternatives in one line
\end{tcolorbox}
\caption{Prompt used for generating object axis paraphrases.
The \texttt{\{ALFWORLD\_ENV\_OBJECTS\}} placeholder is replaced with the
full entity list at generation time.}
\label{fig:prompt-object}
\end{figure*}

%% file: appenfigure/Appen_prompt_verifier.tex
\begin{figure*}[t]
\centering
\begin{tcolorbox}[
  colback=gray!5, colframe=gray!50, boxrule=0.5pt,
  left=6pt, right=6pt, top=6pt, bottom=6pt,
  fontupper=\scriptsize\ttfamily,
  title={\small\textbf{Object Axis Verifier Prompt}},
  coltitle=black, colbacktitle=gray!15
]
You are verifying alternative names generated for an object in a household
environment.\\[4pt]
\textnormal{\textbf{EVALUATION CRITERIA:}}\\
1. \textnormal{\textbf{Same Referent:}} Does the name still pick out the original
   object?\\
2. \textnormal{\textbf{No Lexical Confusion:}} Does it avoid confusion with existing
   scene entities? REJECT if confusable. ("tumbler" for Mug: OK; "cup": NOT OK)\\
3. \textnormal{\textbf{Cell Compliance:}} same\_polarity: noun must be GONE, replaced
   by synonym/sub-kind. REJECT HYPERNYM. addition\_deletion: noun must be KEPT.
   REJECT sensory attributes.\\
4. \textnormal{\textbf{Naturalness:}} Natural English a person would say.\\
5. \textnormal{\textbf{Inflected Forms:}} Four forms right and consistent.
   REJECT wrong article or plural.\\
6. \textnormal{\textbf{No LLM Artifacts:}} REJECT meta-commentary or multiple options
   on one line.\\[4pt]
\textnormal{\textbf{OUTPUT:}} ONLY the ACCEPTED JSON lines, unchanged, one per line.
\end{tcolorbox}

\vspace{4pt}

\begin{tcolorbox}[
  colback=red!5, colframe=red!50, boxrule=0.5pt,
  left=6pt, right=6pt, top=6pt, bottom=6pt,
  fontupper=\scriptsize\ttfamily,
  title={\small\textbf{Action Axis Verifier Prompt}},
  coltitle=black, colbacktitle=red!15
]
You are verifying paraphrased instruction frames for a household agent.\\[4pt]
\textnormal{\textbf{EVALUATION CRITERIA:}}\\
1. \textnormal{\textbf{Placeholders Intact:}} Every placeholder must be
   \{obj:FORM\}, \{recep:FORM\} or \{toggle:FORM\}. REJECT if missing or malformed.
   REJECT hand-written article next to placeholder.\\
2. \textnormal{\textbf{Semantic Preservation:}} Same object, destination, count and
   action. REJECT if a step was added or dropped.\\
3. \textnormal{\textbf{Cell Compliance:}} "hint" ending in "?" is a failure.
   "question\_directive" naming the verb is a failure.
   "permission\_directive" where SPEAKER is the actor is a failure.
   "need\_statement" phrased as "I need you to ..." is a failure.\\
4. \textnormal{\textbf{Naturalness:}} Grammatically correct and natural-sounding.\\
5. \textnormal{\textbf{No LLM Artifacts:}} REJECT meta-commentary or multiple options
   on one line.\\[4pt]
\textnormal{\textbf{OUTPUT:}} ONLY the ACCEPTED frames, one per line.
\end{tcolorbox}
\caption{Verifier prompts for object axis (top) and action axis (bottom).
Generated paraphrases are filtered through these verifiers before
inclusion in ALFWorld-Para.}
\label{fig:prompt-verifier}
\end{figure*}

%% file: figures/appen_alfworldpara_humaneval_guide.tex
\begin{figure*}[t]
\centering
\begin{tcolorbox}[
  colback=blue!3,
  colframe=blue!40,
  boxrule=0.5pt,
  left=6pt, right=6pt, top=6pt, bottom=6pt,
  fontupper=\small\ttfamily,
  title={\small\textbf{Human Evaluation Instruction}},
  coltitle=black,
  colbacktitle=blue!15
]
ALFWorld is a text-based household environment. An agent reads the room as text and
does chores: pick up an item and put it somewhere (optionally cleaning, heating, or
cooling it first), or turn on a lamp to look at an item.\\[4pt]
Each row has an \textbf{original} command and a \textbf{paraphrased} version of it.\\[4pt]
\textbf{Question:} Given these two commands in this environment, \textbf{is it consistent
with the intent for the agent to do the same thing?}\\[4pt]
Put \textbf{O} (yes) or \textbf{X} (no) in the answer column.\\[2pt]
\quad\textbullet~\textbf{O} if it aims for the same object, same place, same handling
(clean/heat/cool), and same count --- even when the wording, tone, sentence structure,
or question form differs.\\
\quad\textbullet~\textbf{X} if the target object, destination, count, or handling
changes, or if it is too vague to tell.\\[4pt]
If unsure whether a renamed object still refers to the same thing, consult the object
reference below.
\end{tcolorbox}

\vspace{4pt}

\begin{tcolorbox}[
  colback=blue!5,
  colframe=blue!60!black,
  boxrule=0.5pt,
  left=6pt, right=6pt, top=6pt, bottom=6pt,
  fontupper=\small\ttfamily,
  title={\small\textbf{Object Reference (provided alongside)}},
  coltitle=black,
  colbacktitle=blue!20
]
\textbf{Items:} alarm clock, apple, book, bowl, box, bread, butter knife, cd, candle,
cell phone, cloth, credit card, cup, dish sponge, egg, fork, glassbottle, key chain,
knife, ladle, laptop, lettuce, mug, newspaper, pan, pencil, pillow, plate, pot, potato,
remote control, soap bar, soap bottle, spatula, spray bottle, statue, tissue box,
toilet paper, tomato, vase, watch, wine bottle\\[4pt]
\textbf{Receptacles:} arm chair, bed, cabinet, cart, coffee machine, coffee table,
counter top, desk, dining table, drawer, dresser, fridge, garbage can, microwave,
ottoman, safe, shelf, side table, sofa, stove burner, toilet, toilet paper hanger\\[4pt]
\textbf{Lamp} (for look/examine tasks): desk lamp
\end{tcolorbox}

\caption{Instruction (top) and object reference (bottom) provided to human 
evaluators for verifying semantic equivalence of paraphrased instructions.}
\label{fig:human-eval-instruction}
\end{figure*}

%% file: appenfigure/extended_object_drop.tex
\begin{figure*}[t]
  \centering
  \includegraphics[width=\textwidth]{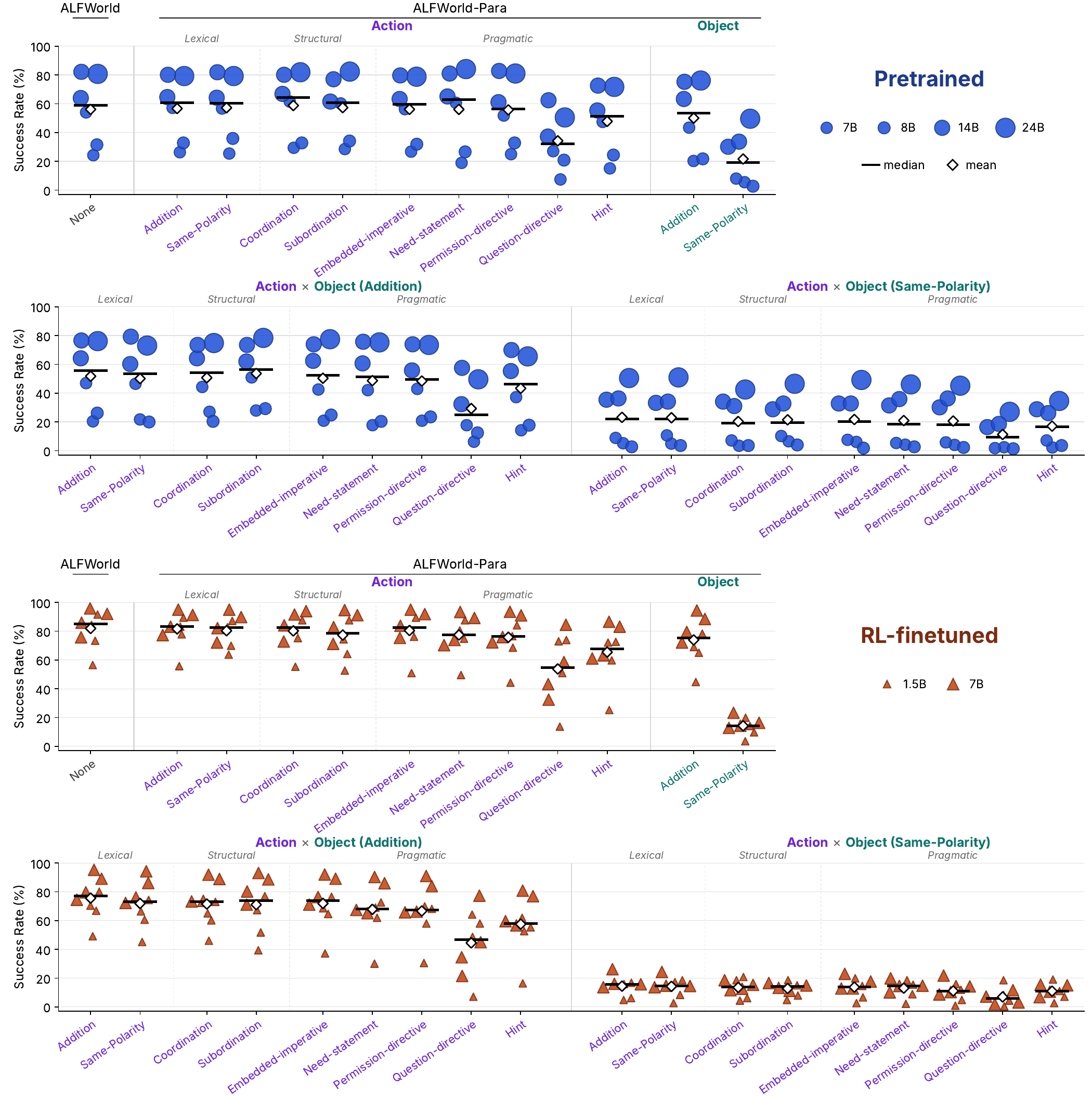}
  \caption{Per-type breakdown of Figure~\ref{fig:analysis_object_drop}, showing every action paraphrase type individually. The top two rows show pretrained LLMs and the bottom two rows show RL-finetuned LLMs. In the first row of each group, success rate (\%) is shown on original ALFWorld (None), on each of the nine action paraphrase types (grouped into lexical, structural, and pragmatic), and on the two object paraphrase types. The second row combines each action type with an object paraphrase, Addition (left) or Same-Polarity (right). As in Figure~\ref{fig:analysis_object_drop}, each marker is one model, with marker size indicating parameter count; horizontal bars and diamonds denote the median and mean across models.}
  \label{appenfigure:extended_object_drop}
\end{figure*}

%% file: appenfigure/appen_paraknow_prompt.tex
\tcbset{
  probebox/.style={
    colback=blue!5, colframe=blue!60!black, boxrule=0.5pt,
    left=6pt, right=6pt, top=4pt, bottom=4pt,
    fontupper=\small\ttfamily, before upper=\raggedright\frenchspacing,
    coltitle=black, colbacktitle=blue!20},
}
\newcommand{\ph}[1]{\textcolor{red!60!black}{\texttt{\{#1\}}}}

\begin{figure*}[t]
\centering
\begin{tcolorbox}[probebox, title={\small\textbf{\ph{CHOICES\_BLOCK}} (82 entity types, shared by all templates)}]
alarmclock, apple, book, bowl, box, bread, butterknife, cd, candle, cellphone, cloth,
creditcard, cup, dishsponge, egg, fork, glassbottle, keychain, knife, ladle, laptop,
lettuce, mug, newspaper, pan, pencil, pillow, plate, pot, potato, remotecontrol,
soapbar, soapbottle, spatula, spraybottle, statue, tissuebox, toiletpaper, tomato,
vase, watch, winebottle, armchair, bed, cabinet, cart, coffeemachine, coffeetable,
countertop, desk, diningtable, drawer, dresser, fridge, garbagecan, microwave,
ottoman, safe, shelf, sidetable, sofa, stoveburner, toilet, toiletpaperhanger,
desklamp, bathtubbasin, handtowelholder, houseplant, kettle, mirror, papertowelroll,
pen, peppershaker, pineapple, saltshaker, sink, sinkbasin, spoon, television, toaster,
towelholder, window
\end{tcolorbox}
\vspace{-2pt}
\begin{tcolorbox}[probebox, title={\small\textbf{\ph{TAIL}}}]
Answer with exactly one name from the list, nothing else.
\end{tcolorbox}
\vspace{-2pt}
\begin{tcolorbox}[probebox, title={\small\textbf{S1: List choice}}]
In a household environment there are these entity types (objects and locations):\\
\ph{CHOICES\_BLOCK}\\[4pt]
An instruction referred to something called a "\ph{p}". Which single entity type from the
list above is it most likely referring to? \ph{TAIL}
\end{tcolorbox}
\vspace{-2pt}
\begin{tcolorbox}[probebox, title={\small\textbf{S2: Direct QA}}]
These are all the entity types in a household environment:\\
\ph{CHOICES\_BLOCK}\\[4pt]
What is a "\ph{p}" called in this environment? \ph{TAIL}
\end{tcolorbox}
\vspace{-2pt}
\begin{tcolorbox}[probebox, title={\small\textbf{S3: Cloze}}]
Entity types in a household environment:\\
\ph{CHOICES\_BLOCK}\\[4pt]
Complete the sentence with one entity type from the list: "A \ph{p} is a kind of \_\_\_". \ph{TAIL}
\end{tcolorbox}
\vspace{-2pt}
\begin{tcolorbox}[probebox, title={\small\textbf{S4: Synonym}}]
Entity types in a household environment:\\
\ph{CHOICES\_BLOCK}\\[4pt]
Which entity type in the list is a synonym of "\ph{p}"? \ph{TAIL}
\end{tcolorbox}
\vspace{-2pt}
\begin{tcolorbox}[probebox, title={\small\textbf{S5: Agent-framed}}]
You are a household robot. The entity types you can interact with are:\\
\ph{CHOICES\_BLOCK}\\[4pt]
You are told to pick up the "\ph{p}". Which entity type do you pick up? \ph{TAIL}
\end{tcolorbox}
\caption{The five prompt templates of the ALFWorld paraphrase-knowledge probe. \ph{p} is the paraphrased noun; \ph{CHOICES\_BLOCK} lists the 82 entity types in environment (lowercase) form, ordered as objects (42), receptacles (22), toggleable (1) and scenery (17); \ph{TAIL} is identical to the ScienceWorld probe.}
\label{appenfigure:paraknow_prompt}
\end{figure*}

%% file: figures/appen_sciworldpara_humaneval_guide.tex
\begin{figure*}[t]
\centering
\begin{tcolorbox}[
  colback=blue!3,
  colframe=blue!40,
  boxrule=0.5pt,
  left=6pt, right=6pt, top=6pt, bottom=6pt,
  fontupper=\small\ttfamily,
  title={\small\textbf{Human Evaluation Instruction}},
  coltitle=black,
  colbacktitle=blue!15
]
ScienceWorld is a simulation of the world abstracted through a complex interactive
text environment in English with many objects, actions, and simulation engines.\\[4pt]
Each row has an \textbf{original} command and a \textbf{paraphrased} version of it.\\[4pt]
\textbf{Question:} Given these two commands in this environment, \textbf{is it consistent
with the intent for the agent to do the same thing?}\\[4pt]
Put \textbf{O} (yes) or \textbf{X} (no) in the answer column.\\[2pt]
\quad\textbullet~\textbf{O} if it aims for the same object, same place, same handling
(clean/heat/cool), and same count --- even when the wording, tone, sentence structure,
or question form differs.\\
\quad\textbullet~\textbf{X} if the target object, destination, count, or handling
changes, or if it is too vague to tell.\\[4pt]
If unsure whether a renamed object still refers to the same thing (i.e., object name
uniqueness), consult the object reference below.
\end{tcolorbox}

\vspace{4pt}

\begin{tcolorbox}[
  colback=blue!5,
  colframe=blue!60!black,
  boxrule=0.5pt,
  left=6pt, right=6pt, top=6pt, bottom=6pt,
  fontupper=\small\ttfamily,
  title={\small\textbf{Object Reference (provided alongside)}},
  coltitle=black,
  colbacktitle=blue!20
]
Consult only if a synonym is unclear. Distinct entities here (a synonym that names a
DIFFERENT one of these is NOT the same task):\\[4pt]
\textbf{Targets} (the thing the task is about): tin, water, metal pot, paper clip,
plastic fork, jam sandwich\\[4pt]
\textbf{Rooms:} bathroom, kitchen, living room, outside, workshop\\[4pt]
\textbf{Categories:}\\
\quad\textit{animal:} ant, beaver, blue jay, brown bear, butterfly, chameleon, chipmunk,
common toad, crocodile, dove, dragonfly, elephant, frog, giant tortoise, hedgehog, moth,
mouse, parrot, rabbit, turtle, wolf\\
\quad\textit{plant:} apple tree, avocado tree, banana tree, cherry tree, lemon tree,
orange tree, pea plant, peach tree\\
\quad\textit{living thing:} every animal and every plant above\\
\quad\textit{non-living thing:} anything movable that is not an animal or a plant\\[4pt]
\textbf{Measurements and properties} (these must survive unchanged): melting point,
electrical conductivity (conductive / nonconductive), life span, the temperature
threshold and its unit (degrees celsius)
\end{tcolorbox}

\caption{Instruction (top) and object reference (bottom) provided to human
evaluators for verifying semantic equivalence of paraphrased instructions in
ScienceWorld-Para.}
\label{fig:sciworld-human-eval-instruction}
\end{figure*}

%% file: appenfigure/heatmap_entire.tex
\begin{figure*}[t]
  \centering
  \includegraphics[width=\textwidth]{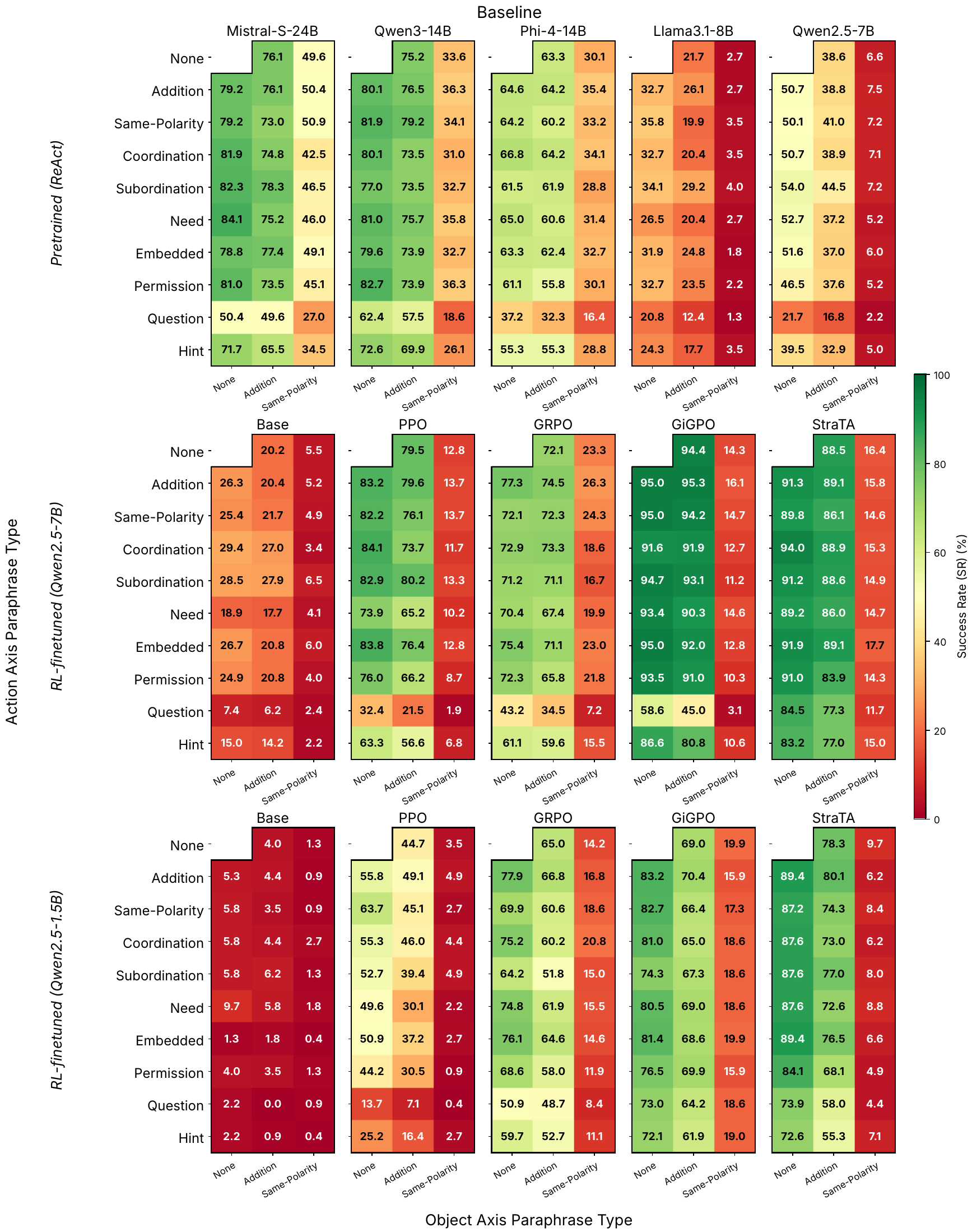}
  \caption{Success rate (SR) (\%) on ALFWorld-Para for every action-axis (rows) and object-axis (columns) paraphrase type, without any mitigation (Baseline). Each panel is one model, grouped into ReAct-prompted pretrained LLMs (top) and RL-finetuned sLLMs on the Qwen2.5-7B (middle) and 1.5B (bottom) backbones.}
  \label{appenfigure:heatmap_no_method}
\end{figure*}

\begin{figure*}[t]
  \centering
  \includegraphics[width=\textwidth]{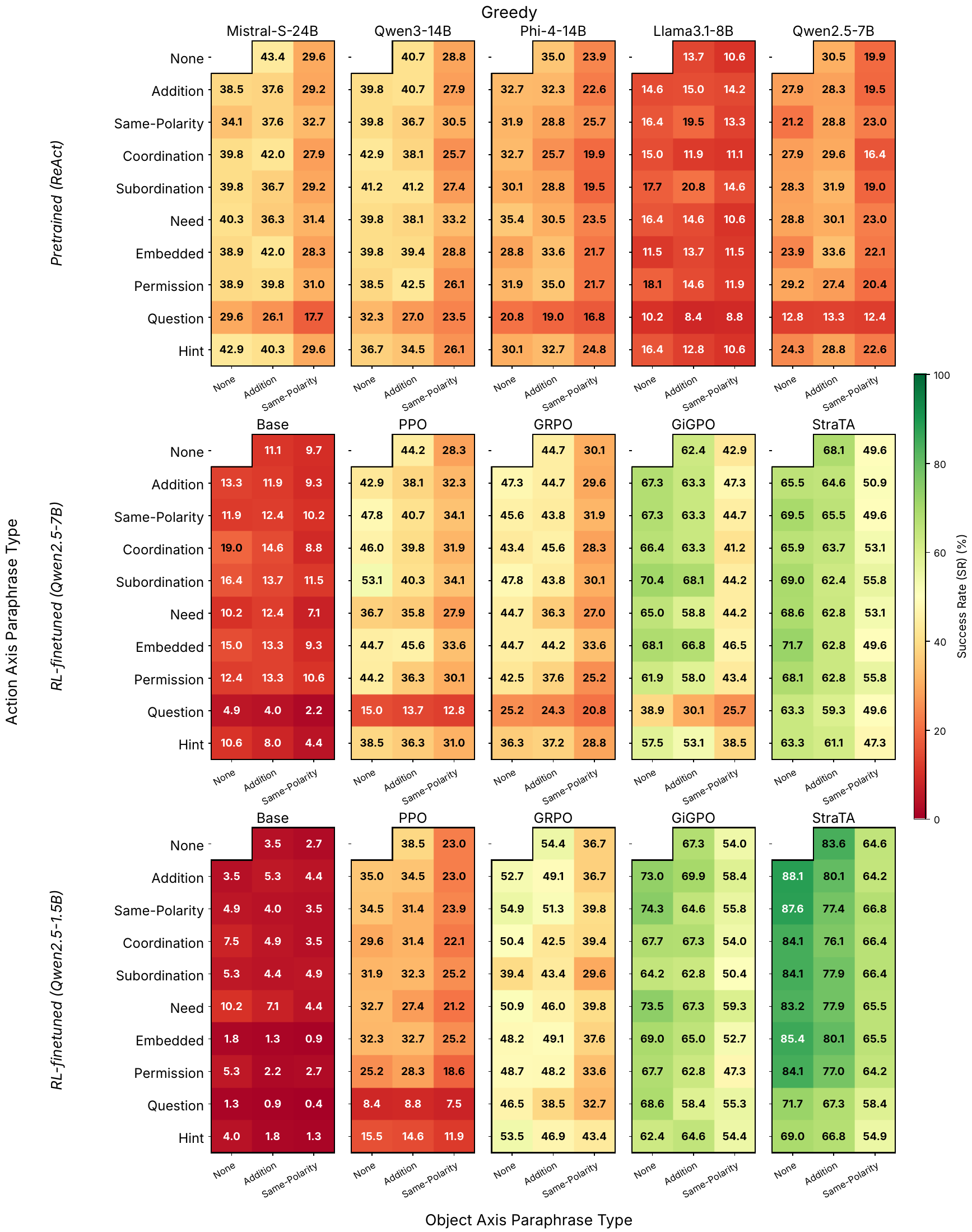}
  \caption{Success rate (SR) (\%) on ALFWorld-Para by paraphrase type under Greedy.}
  \label{appenfigure:heatmap_greedy}
\end{figure*}

\begin{figure*}[t]
  \centering
  \includegraphics[width=\textwidth]{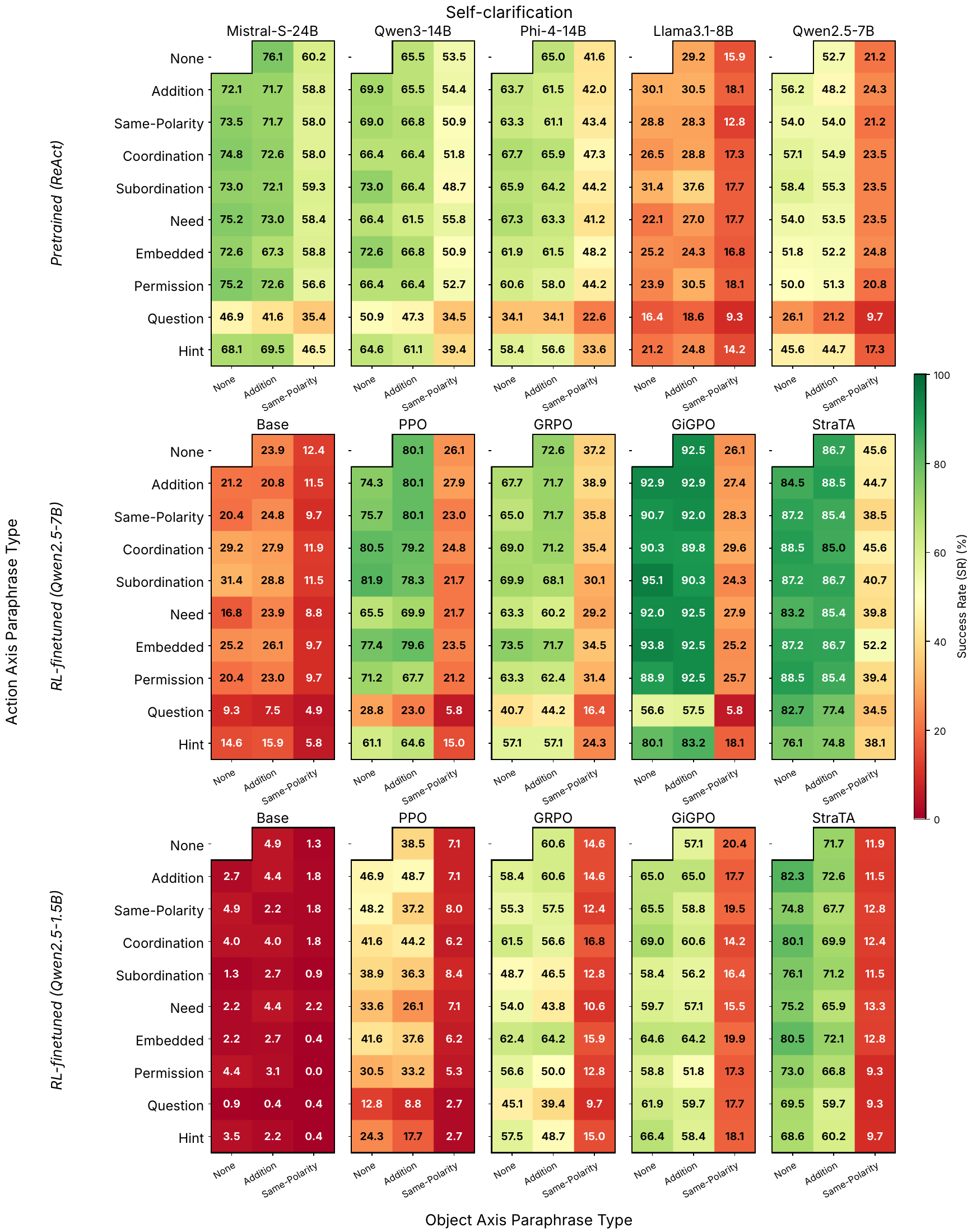}
  \caption{Success rate (SR) (\%) on ALFWorld-Para by paraphrase type under Self-clarification.}
  \label{appenfigure:heatmap_self_clar}
\end{figure*}

\begin{figure*}[t]
  \centering
  \includegraphics[width=\textwidth]{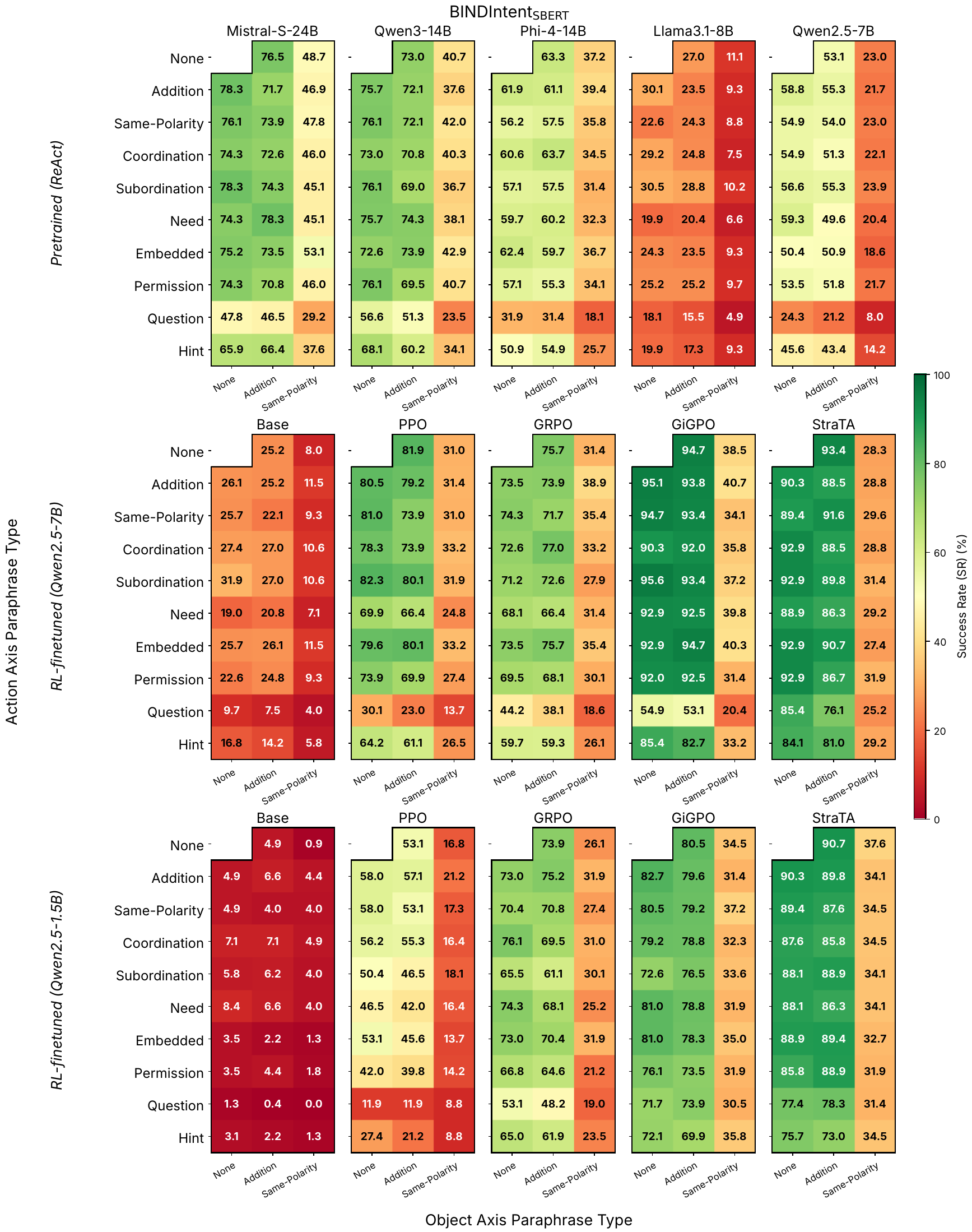}
  \caption{Success rate (SR) (\%) on ALFWorld-Para by paraphrase type under BINDIntent\textsubscript{SBERT}.}
  \label{appenfigure:heatmap_sbert}
\end{figure*}

\begin{figure*}[t]
  \centering
  \includegraphics[width=\textwidth]{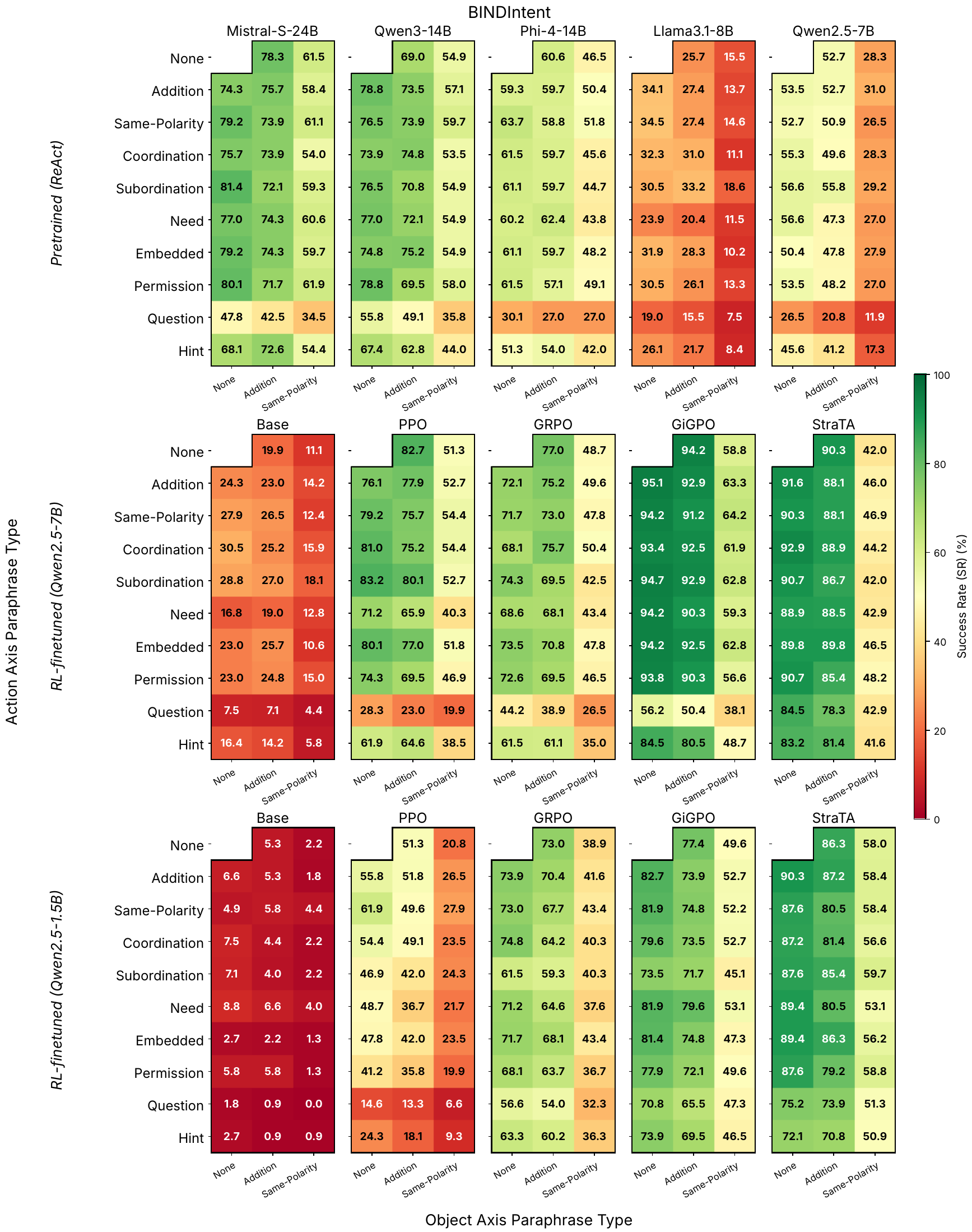}
  \caption{Success rate (SR) (\%) on ALFWorld-Para by paraphrase type under BINDIntent (ours).}
  \label{appenfigure:heatmap_ours}
\end{figure*}